\documentclass[11pt,a4paper]{article}

\usepackage[a4paper,margin=1in]{geometry}
\usepackage{amsmath,amssymb,amsfonts}
\usepackage{array, adjustbox}
\usepackage{float}
\usepackage{booktabs}
\usepackage{tcolorbox}
\tcbuselibrary{skins,breakable}
\usepackage{tabularx}
\usepackage{multirow}
\usepackage{afterpage}
\usepackage{pifont}
\usepackage{physics}
\usepackage{enumitem}
\usepackage{changepage}
\usepackage{tikz}
\usetikzlibrary{arrows.meta,positioning,matrix,calc,shapes.geometric,decorations.pathreplacing}
\tikzset{>=Stealth}
\usepackage{algorithm}
\usepackage{algorithmic}
\usepackage[hidelinks]{hyperref}

\newcommand{\cmark}{\checkmark}
\newcommand{\xmark}{\ding{55}}

\newcolumntype{L}{>{\raggedright\arraybackslash}X}
\newcolumntype{C}{>{\centering\arraybackslash}X}
\newcolumntype{R}{>{\raggedleft\arraybackslash}X}

\usepackage[letterspace=150]{microtype}
\usepackage{caption}
\makeatletter
\newcommand{\keywords}[1]{%
  \vspace{0.75\baselineskip}%
  \noindent\textbf{Keywords:} #1%
}
\makeatother

\newlength{\extralength}
\newlength{\fulllength}
\newcommand{\authorcontributions}[1]{\par\medskip\noindent\textbf{Author Contributions:} #1\par\medskip}
\newcommand{\funding}[1]{\par\medskip\noindent\textbf{Funding:} #1\par\medskip}
\newcommand{\dataavailability}[1]{\par\medskip\noindent\textbf{Data Availability Statement:} #1\par\medskip}
\newcommand{\conflictsofinterest}[1]{\par\medskip\noindent\textbf{Conflicts of Interest:} #1\par\medskip}
\newcommand{\acknowledgments}[1]{\par\medskip\noindent\textbf{Acknowledgments:} #1\par\medskip}

\newcommand{\appendixtitles}[1]{}
\newcommand{\appendixstart}{}

\title{Can Semantic Methods Enhance Team Sports Tactics?\\A Methodology for Football with Broader Applications}

\author{
Alessio Di Rubbo$^{1}$\thanks{Stake Lab, University of Molise, Italy. ORCID: 0009-0001-2004-3791.} 
\and Mattia Neri$^{2}$\thanks{Bioretics, Bologna, Italy. ORCID: 0009-0000-1024-7064.} 
\and Remo Pareschi$^{1,*}$\thanks{Stake Lab, University of Molise, Italy. ORCID: 0000-0002-4912-582X. Corresponding author: \texttt{remo.pareschi@unimol.it}} 
\and Marco Pedroni$^{3}$\thanks{Institute for Generative Strategy, Ferrara, Italy. ORCID: 0009-0005-9516-4079.} 
\and Roberto Valtancoli$^{4}$\thanks{Cesena Femminile Football Club, Cesena, Italy. ORCID: 0009-0008-5946-6303.}  
\and Paolino Zica$^{5}$\thanks{Zica Sport, Benevento, Italy. ORCID: 0009-0006-0556-5651.}
}

\date{}

\begin{document}

\maketitle

\begin{center}
\begin{tcolorbox}[colback=white, colframe=black!40, boxrule=0.4pt, arc=1pt,
                  left=8pt, right=8pt, top=6pt, bottom=6pt, width=0.92\textwidth]
\noindent\textbf{arXiv Note:} This version corresponds to the final published article
in \emph{Sci} 2026, Vol.~8, Issue~3, Article~63.
DOI: \href{https://doi.org/10.3390/sci8030063}{10.3390/sci8030063}.
Formatted for arXiv compliance by removing publisher-specific formatting while
preserving all scholarly content, corrections, figures, and references.
\end{tcolorbox}
\end{center}

\begin{abstract}
This paper explores how semantic-space reasoning, traditionally used in computational linguistics, can be extended to tactical decision-making in team sports. Building on the analogy between \emph{texts
} and \emph{teams}---where players act as words and collective play conveys meaning---the proposed methodology models tactical configurations as compositional semantic structures. Each player is represented as a multidimensional vector integrating technical, physical, and psychological attributes; team profiles are aggregated through contextual weighting into a higher-level semantic representation.
Within this shared vector space, tactical templates such as \emph{high press}, \emph{counterattack}, or \emph{possession build-up} are encoded analogously to linguistic concepts. Their alignment with team profiles is evaluated using vector-distance metrics, enabling the computation of tactical ``fit'' and opponent-exploitation potential. A Python-based prototype demonstrates how these methods can generate interpretable, dynamically adaptive strategy recommendations, accompanied by fine-grained diagnostic insights at the attribute level. Evaluation through synthetic scenarios and a pilot study with real match data establishes internal consistency and feasibility of the approach; operational validity in live coaching contexts remains an open question for future prospective validation.
Beyond football, the framework offers a potentially generalizable approach for collective decision-making in team-based domains---ranging from basketball and hockey to cooperative robotics and human--AI coordination systems. The paper concludes by outlining future directions toward real-world data integration, predictive simulation, and the validation work required before operational deployment
\end{abstract}




\section{Introduction}
Modern football has undergone a radical transformation, evolving from a discipline grounded mainly in coaches' intuition and experience into one profoundly shaped by objective data analysis. The~widespread adoption of advanced analytics systems, proprietary metrics such as expected goals (xG) and expected assists (xA), and~the availability of detailed information on players’ physical, technical, and~tactical performance have enabled a quantitative understanding of phenomena once accessible only through human judgment~\cite{pollard1997measuring}.

In this data-driven landscape, \textbf{tactical optimization
}---the ability to select and dynamically adjust playing strategies according to the team’s internal characteristics and the contingent match conditions---has become a decisive competitive factor. At~elite levels, marginal advantages can determine the outcome of an entire season. Tactical effectiveness no longer depends solely on individual talent or preparation quality but also on the ability to interpret complex contexts, anticipate opponents’ actions, and~adapt strategies in real time. However, traditional decision models based primarily on qualitative heuristics and experience reach their limits when faced with the high dimensionality and dynamism of modern play~\cite{mackenzie2013performance,sarmento2014match}.

Despite significant progress in match analysis, a~fundamental disconnect persists between \textbf{quantitative tools} (performance indicators, spatial distributions, xG models) and \textbf{qualitative factors} that critically influence performance: group cohesion, psychological resilience, team morale, and~residual energy~\cite{weinberg2023foundations,mclean2017systems}. Current decision support systems emphasize easily measurable variables while neglecting intangible dimensions that often prove decisive under~pressure.

\textbf{Research gap:} This disconnect causes (i) loss of strategically relevant information, \mbox{(ii) limited} adaptability of recommendations, and~(iii) persistent reliance on subjective intuition in crucial match phases~\cite{rein2016bigdata}. No existing framework provides a unified representation integrating quantitative performance data with qualitative contextual factors into a single, computationally tractable space for automated tactical~recommendation.

To address this gap, the~present study introduces a \textbf{Decision Support System for Tactical Optimization} using a novel \textbf{semantic-distance methodology}. The~core innovation is representing both team states and tactical strategies as vectors in a shared 14-dimensional attribute space, enabling direct geometric comparison. The~system recommends strategies by minimizing weighted Euclidean distance between a team's profile and each strategy's requirements; the rationale for this metric choice is detailed in Section~\ref{sec:distance_matching}.

This approach extends the semantic-distance methodology from \emph{Recommending Actionable Strategies
} \cite{ghisellini2025recommending}---originally developed for bridging analytical frameworks with decision heuristics---to operational football tactics. The~key adaptations~are:
\begin{itemize}[]
    \item replacing general decision categories with 14 concrete \textbf{macro-attributes} capturing technical, psychological, and~organizational dimensions; and
    \item replacing general heuristics with 20 \textbf{canonical football strategies} (e.g., high pressing, counterattack, positional defense).
\end{itemize}

\textbf{What distinguishes this work:} (i) unified representation integrating quantitative and qualitative factors; (ii) dynamic attribute weights adjusting in real time based on match context; (iii) transparent diagnostics showing \emph{why} a strategy is~recommended.

\subsection{Objectives and Contributions
}
This study aims to: (1) formalize a semantic model encoding team states and strategies in a shared attribute space; (2) develop a prototype DSS with context-aware recommendations; (3) evaluate internal consistency and real-data~feasibility.

The main contributions~are:
\begin{enumerate}[]
    \item \textbf{Semantic Model:} 14 macro-attributes synthesizing team complexity, with~20 canonical strategies as ideal-profile vectors.
    \item \textbf{Adaptive Engine:} Python prototype with dynamic weighting adjusting recommendations based on energy, time pressure, and~opponent characteristics.
    \item \textbf{Systematic Validation:} Evaluation through synthetic scenarios and German youth football data, including ablation and robustness analyses.
\end{enumerate}

\subsection{Paper Organization}
Section~\ref{sec:background} reviews related work; Section~\ref{sec:methodology} presents methodology; Section~\ref{sec:implementation} describes implementation; Section~\ref{sec:experiments} reports evaluation; Section~\ref{sec:pilot_validation} presents the pilot study; Section~\ref{sec:discussion} discusses limitations; Section~\ref{sec:conclusion} concludes.

\section{Background and Related~Work}
\label{sec:background}

The introductory section highlighted the need to bridge the gap between quantitative analytics and heuristic decision-making in football. To~formalize the proposed solution, it is first necessary to establish a solid conceptual foundation that clarifies the distinction between \emph{strategy} and \emph{tactics}, and~then to situate this distinction within the broader context of semantic modeling and decision-support~research.

\subsection{Strategic and Tactical Analysis in~Football}

In everyday football discourse, the~terms \emph{strategy} and \emph{tactics} are often used interchangeably. However, in~the academic and analytical literature, they refer to distinct levels of decision-making that are crucial to our~methodology.

\textbf{Strategy} (or playing identity) defines the overall approach or long-term plan through which a team intends to compete. It depends on structural and contextual factors such as squad quality, key players’ technical and physical profiles, seasonal goals, the~coach’s philosophy, and~the team’s physical and psychological resources~\cite{grehaigne1995tactical,mclean2017systems}. Strategy answers the question: \emph{What do we want to achieve?}—for example, controlling the game through ball~possession.

\textbf{Tactics}, in~contrast, represent the operational choices and on-field configurations that translate strategy into concrete actions, often in response to real-time match dynamics. They include formation choices, player assignments, coordinated movements (e.g., defensive shifts), and~in-game adaptations such as introducing an additional forward when chasing a result. Tactics answer the question: \emph{How do we achieve it?}

This distinction is central to the proposed \textbf{Decision Support System (DSS)}. The~system operates at the tactical level—optimizing action choices based on a multidimensional strategic representation of the team. The~semantic-distance model quantifies the alignment~between:
\begin{enumerate}[]
    \item the \emph{strategic vector} of the team (its current state, defined by 14 macro-attributes), and~    \item the \emph{ideal tactical vector} (the target profile of a given strategy, such as counterattack or high pressing).
\end{enumerate}

A correct balance between strategic identity and tactical flexibility ensures internal coherence. Teams with strong strategic identity but low adaptability become predictable and fragile, while excessive tactical improvisation undermines structural stability and collective performance~\cite{he2022flexibility}.

\subsection{Canonical Tactical Strategies in Modern~Football}

The following tactical archetypes comprise the conceptual foundation of our vector modeling framework. For~each, the~team attributes required for effective implementation are~indicated.

\textbf{High Pressing.} A proactive approach aimed at regaining possession in the opponent’s half by applying intense, coordinated pressure. It reduces opponents’ time and space, forcing errors and enabling rapid goal opportunities~\cite{andrienko2017pressing,bauer2021counterpressing,low2021defending}. It requires exceptional physical conditioning, coordination, and~risk~tolerance.

\textbf{Counterattack (Rapid Transition).} Based on defending in a compact mid-low block to lure the opponent forward, then striking rapidly upon regaining possession. It exploits spaces behind the defense and requires speed, verticality, and~sharp~decision-making.

\textbf{Positional Defense.} A space-oriented approach emphasizing spatial control over immediate 
pressure. Spatio-temporal analysis methods have been developed to 
quantify team coordination and territorial control~\cite{gudmundsson2017spatiotemporal}. Positional defense prioritizes 
equilibrium, communication, and~tactical discipline while conserving 
energy~\cite{forcher2024compactness}.

\textbf{Gegenpressing (Pressing After Loss).} An aggressive evolution of pressing, aiming to recover the ball within 3–5 s after losing it by exploiting the opponent’s temporary disorganization. Extremely demanding, it requires maximal energy, readiness, and~synchronization.

\textbf{Build-up Play.} A possession-based approach initiating offensive buildup from the back through short passes and gradual progression, designed to control tempo and overcome pressure via numerical superiority~\cite{turney2010frequency}. It requires technically skilled players across all lines, especially defenders and goalkeepers, who can distribute the~ball.

These archetypes serve as idealized templates within our system, allowing the computational comparison of a team’s actual state with prototypical tactical~profiles.

\subsection{Semantic Distance~Models}
\label{sec:semantic_distance_background}

\textbf{Semantic distance} provides a quantitative measure of how far two informational entities—concepts, documents, or~representations—differ in meaning when embedded in a shared vector space. In~natural language processing (NLP), such models rest on the principle that numerical representations of linguistic units capture latent semantic relations, enabling mathematical comparison across heterogeneous content~\cite{turney2010frequency,mikolov2013efficient}.

Classical approaches~include:
\begin{itemize}[]
    \item \textbf{Cosine similarity}, which measures the angle between normalized vectors, robust to scale differences;
    \item \textbf{Euclidean distance}, which quantifies geometric deviation in continuous space;
    \item \textbf{Probabilistic metrics}, such as Kullback–Leibler~\cite{kullback1951information} or Jensen–Shannon~\cite{lin1991divergence} divergences, used when entities are modeled as probability distributions.
\end{itemize}

``With the advent of Transformer architectures (e.g., BERT, RoBERTa, 
Sentence-\mbox{BERT) \cite{devlin2019bert, reimers2019sentencebert}}, 
Contextual embeddings have dramatically improved representation quality, dynamically capturing meaning and outperforming static models such as Word2Vec and GloVe. These techniques have been widely adopted in information retrieval, question answering, text classification, and~recommender systems~\cite{turban2011decision}.

In the reference paper \emph{Recommending Actionable Strategies} \cite{ghisellini2025recommending}, semantic distance was used to integrate two historically distinct traditions in strategy~theory:
\begin{enumerate}[]
    \item structured analytical frameworks (e.g., SWOT, 6C), and~    \item decision heuristics (e.g., the~Thirty-Six Stratagems).
\end{enumerate}

\noindent Both
 were projected into a shared semantic space, enabling the computation of similarity matrices that link structured analysis to heuristic insight. This pipeline demonstrated how semantic methods can act as an interpretive bridge between abstract models and actionable~guidance.

The present research adapts that paradigm to the football domain, replacing general analytical categories with 14 football-specific macro-attributes (e.g., Offensive Strength, Tactical Cohesion, Psychological Resilience) and general heuristics with canonical tactical strategies. The~optimal tactical choice $S^*$ is thus defined as the strategy minimizing the semantic distance $d(V_\text{team}, V_\text{strategy}(S))$ between the team’s current vector representation and the target tactical profile:
\[
S^* = \arg\min_{S} d(V_\text{team}, V_\text{strategy}(S)).
\]

\subsection{Decision Support Systems in~Sports}

\textbf{Decision Support Systems (DSS)} are computational tools designed to assist coaches, analysts, and~managers in complex decision-making by integrating quantitative data, expert knowledge, and~predictive modeling capabilities. The~increasing availability of high-resolution data—from GPS tracking, wearable sensors, and~video-analysis platforms—has fostered the development of DSS capable of transforming information into operational insight~\cite{rein2016bigdata,sanders2018fatigue}.

Across sports, DSS applications range from performance optimization to injury prevention and tactical~planning:
\begin{itemize}[]
    \item \textbf{Athletics and individual sports}—systems such as Catapult AMS or Kitman Labs monitor fatigue and workload by combining physiological and subjective data;
    \item \textbf{Basketball and team sports}—platforms like Synergy Sports and Second Spectrum merge positional tracking with video analytics to identify offensive and defensive patterns~\cite{goldsberry2012courtvision};
    \item \textbf{Cycling and endurance disciplines}—predictive tools such as Performance Management Charts use power and heart-rate data to optimize training loads.
\end{itemize}

In football, systems like \textbf{Wyscout} and \textbf{InStat} provide video-based statistical analytics; \textbf{StatsBomb IQ} integrates positional and event data into advanced metrics (e.g., xG, passing networks); \textbf{SciSports Insight} uses AI-based indices for player recruitment and compatibility analysis; and \textbf{SkillCorner} applies computer vision to extract player trajectories in real time~\cite{pappalardo2019public}.

While these systems have expanded analytical capabilities, most focus on quantitative or spatial data, overlooking qualitative and psychological aspects such as morale, cohesion, and~resilience. Moreover, strategic recommendations often rely on expert interpretation rather than automated reasoning. The~present work addresses this methodological gap by introducing a semantic-distance-based DSS that integrates multidimensional, context-aware modeling—combining quantitative metrics and tacit knowledge into a unified, interpretable~framework.

\section{Methodology}
\label{sec:methodology}
\unskip

\subsection{Theoretical~Framework}
We adapt the methodology of \emph{Recommending Actionable Strategies} \cite{ghisellini2025recommending} to the football domain, aiming to build a tactical recommender that integrates a team’s technical, organizational, and~psychological dimensions within a shared semantic space. The~core idea is to encode both (i) the contextual state of a team and (ii) the ideal profiles of canonical tactical strategies in the same vector space, and~then to select the tactic whose profile is closest (in a semantic–geometric sense) to the team’s current state. Recommendations can be updated dynamically as match conditions evolve (e.g., residual energy, technical/physical gaps, time pressure).

Three pillars characterize this~approach:
\begin{enumerate}[]
    \item \textbf{Multidimensional integration} of quantitative (individual and collective performance) and qualitative (morale, cohesion, psychological resilience) factors.
    \item \textbf{Semantic formalization} via normalized vectors in a common space, enabling consistent comparisons between teams and tactics.
    \item \textbf{Dynamic adaptability} through real-time reweighting of distances using match \mbox{conditions}.
\end{enumerate}


\subsection{Context Tree and~Aggregation}
\label{sec:context_tree}

We represent team context with a hierarchical \emph{context tree} that aggregates heterogeneous data sources into a unified vector representation. The~tree has three levels:

\begin{enumerate}
    \item \textbf{Leaf level:} Raw observables from match analytics—player-level metrics from event data (passes, shots, tackles), tracking data (sprint distance, positioning), and~physiological monitoring (heart rate, estimated fatigue).
    
    \item \textls[-20]{\textbf{Intermediate level:} Role-aggregated attributes computed by combining leaf-level data within positional groups (e.g., ``forward line offensive output,'' ``\mbox{midfield ball retention}'').}
    
    \item \textbf{Root level:} The 14 macro-attributes ($A_1, \ldots, A_{14}$) that define the shared semantic space, computed by a weighted combination of intermediate-level signals.
\end{enumerate}

\begin{figure}[H]

\begin{tikzpicture}[
    >=Stealth,
    level 1/.style={sibling distance=4.2cm, level distance=1.6cm},
    level 2/.style={sibling distance=2.0cm, level distance=1.6cm},
    every node/.style={align=center, font=\small},
    root/.style={draw, rounded corners, fill=blue!15, minimum width=2.8cm, minimum height=0.7cm},
    intermediate/.style={draw, rounded corners, fill=green!15, minimum width=2.2cm, minimum height=0.6cm, font=\scriptsize},
    leaf/.style={draw, rounded corners, fill=gray!15, minimum width=1.6cm, minimum height=0.5cm, font=\scriptsize},
    edge from parent/.style={draw, ->, thick}
]

\node[root] {$A_1$: Offensive Strength}
    child {
        node[intermediate] {Forward Line\\Output}
        child { node[leaf] {xG} }
        child { node[leaf] {Shot Acc.} }
    }
    child {
        node[intermediate] {Midfield\\Creativity}
        child { node[leaf] {xA} }
        child { node[leaf] {Key Passes} }
    }
    child {
        node[intermediate] {Wide\\Contribution}
        child { node[leaf] {Crosses} }
        child { node[leaf] {Dribbles} }
    };

\end{tikzpicture}

\vspace{0.8cm}

\begin{tikzpicture}[
    >=Stealth,
    level 1/.style={sibling distance=4.2cm, level distance=1.6cm},
    level 2/.style={sibling distance=2.0cm, level distance=1.6cm},
    every node/.style={align=center, font=\small},
    root/.style={draw, rounded corners, fill=blue!15, minimum width=2.8cm, minimum height=0.7cm},
    intermediate/.style={draw, rounded corners, fill=green!15, minimum width=2.2cm, minimum height=0.6cm, font=\scriptsize},
    leaf/.style={draw, rounded corners, fill=gray!15, minimum width=1.6cm, minimum height=0.5cm, font=\scriptsize},
    edge from parent/.style={draw, ->, thick}
]

\node[root] {$A_8$: Residual Energy}
    child {
        node[intermediate] {Outfield\\Stamina}
        child { node[leaf] {Sprint Dist.} }
        child { node[leaf] {High-Int. Runs} }
    }
    child {
        node[intermediate] {Recovery\\State}
        child { node[leaf] {Rest Days} }
        child { node[leaf] {Match Load} }
    }
    child {
        node[intermediate] {In-Match\\Fatigue}
        child { node[leaf] {Min. Played} }
        child { node[leaf] {Intensity Decay} }
    };

\end{tikzpicture}

\caption{Context
 tree structure for two representative macro-attributes. Leaf nodes contain raw observables from match data; intermediate nodes aggregate by functional role; root nodes are the macro-attributes used in semantic distance computation. Edges represent weighted \mbox{aggregation~functions}.}
\label{fig:context_tree}
\end{figure}
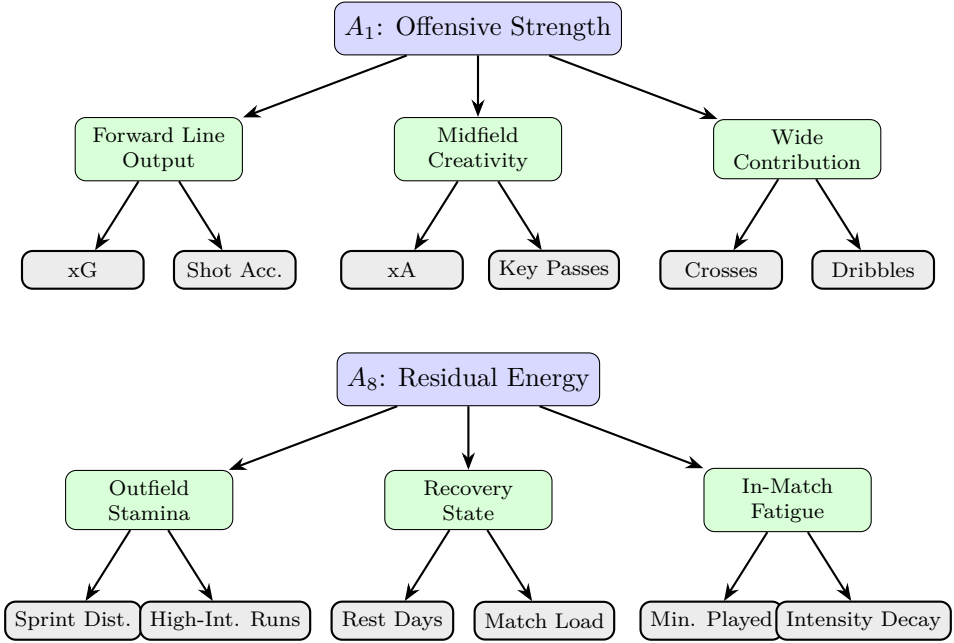

Figure~\ref{fig:context_tree} illustrates this hierarchical structure for a subset of~attributes.

\subsubsection{Aggregation Example%
}
To illustrate the aggregation process concretely, consider how $A_1$ (Offensive Strength) is computed for a team fielding a 4-3-3 formation:

\begin{enumerate}
    \item \textbf{Leaf level:} Extract per-player metrics—e.g., Striker A: xG $= 0.82$, shot accuracy $= 0.71$; Winger B: xA $= 0.65$, successful dribbles $= 0.78$.
    
    \item \textbf{Intermediate level:} Aggregate within positional groups using role-based weights:
    \begin{align*}
        \text{Forward Output} &= 0.5 \times \text{xG}_{\text{ST}} + 0.3 \times \text{ShotAcc}_{\text{ST}} + 0.2 \times \text{xG}_{\text{wings}} \\
        \text{Midfield Creativity} &= 0.6 \times \text{xA}_{\text{CAM}} + 0.4 \times \text{KeyPasses}_{\text{CM}}
    \end{align*}

    \item \textbf{Root level:} Combine intermediate values into the macro-attribute:
    \[
        A_1 = 0.50 \times \text{\textls[-15]{Forward Output}} + 0.30 \times \text{\textls[-15]{Midfield Creativity}} + 0.20 \times \text{\textls[-15]{Wide Contribution}}
    \]
\end{enumerate}

All intermediate and final values are normalized to $[0,1]$ via min-max scaling against league or historical benchmarks, ensuring cross-team comparability. The~normalization procedure is specified in detail~below.

\subsubsection{Normalization Procedure}
\label{sec:normalization}

To ensure consistent scaling across teams and time periods, each raw attribute value $x$ is transformed to a normalized value $\tilde{x} \in [0,1]$ using the following protocol:

\textbf{Min-Max Scaling} \textbf{Formula.}

\begin{equation}
\tilde{x} = \frac{x - x_{\min}}{x_{\max} - x_{\min}}
\label{eq:minmax}
\end{equation}

\noindent where $x_{\min}$ and $x_{\max}$ are benchmark bounds derived from reference populations as \mbox{specified~below}.

{\textbf{Benchmark Population Definition.}}
Normalization benchmarks are computed from a \emph{reference population} defined as follows:

\begin{itemize}
    \item \textbf{League-level benchmarks (default):} For professional deployments, benchmarks are derived from all players in the same league and division (e.g., Bundesliga, Serie A) over the reference window. This ensures that a normalized value of 0.5 represents league-average~performance.
    
    \item \textbf{Competition-level benchmarks:} For tournament contexts (e.g., Champions League, World Cup), benchmarks may be computed across all participating teams to reflect the elevated~baseline.
    
    \item \textbf{Historical team benchmarks:} For longitudinal tracking of a single team, benchmarks may be derived from that team's own historical range, enabling detection of relative improvement or decline.
\end{itemize}

\noindent {The} current prototype uses \emph{synthetic benchmarks} derived from the role-specific distributions in Table~\ref{tab:role_distributions} (Appendix~\ref{app:specification}), with~$x_{\min} = \mu - 2\sigma$ and $x_{\max} = \mu + 2\sigma$ for each attribute--role~combination.

{\textbf{Reference Window (Temporal Scope).}}
To prevent instability from outlier matches, benchmarks are computed over a rolling window:

\begin{itemize}
    \item \textbf{Season window (default):} The most recent complete season (e.g., 34--38 matches for major European leagues). This captures stable population characteristics while remaining~current.
    
    \item \textbf{Rolling window:} For mid-season deployment, a~rolling window of the most recent $N = 10$ league matches provides more responsive benchmarks, updated~weekly.
    
    \item \textbf{Fixed historical window:} For retrospective analysis or cross-season comparison, a~fixed reference period (e.g., 2022--23 season) ensures consistent scaling.
\end{itemize}

{\textbf{Clipping and Robustness.}}
To handle outliers and ensure bounded outputs:

\begin{enumerate}
    \item \textbf{Pre-normalization clipping:} Raw values outside $[x_{\min}, x_{\max}]$ are clipped to the boundary values before applying Equation~(\ref{eq:minmax}). This prevents exceptional performances (positive or negative) from distorting the~scale.
    
    \item \textbf{Robust benchmark estimation:} Benchmarks may optionally use the 5th and 95th percentiles rather than true min/max to reduce sensitivity to extreme outliers:
    \[
    x_{\min} = P_5(\mathcal{D}), \quad x_{\max} = P_{95}(\mathcal{D})
    \]
    where $\mathcal{D}$ is the reference population~distribution.
    
    \item \textbf{Floor for near-zero ranges:} If $x_{\max} - x_{\min} < \epsilon$ (indicating near-constant values), the~attribute is assigned the default value 0.5 to avoid division instability.
\end{enumerate}

{\textbf{Leakage Prevention.}}
In retrospective evaluation and real-time deployment, normalization must use only information available at decision time:

\begin{enumerate}
    \item \textbf{Temporal ordering:} Benchmarks for match $t$ are computed from data up to match $t-1$ only. Future match data are never included in benchmark~computation.
    
    \item \textbf{Held-out validation:} When evaluating DSS performance over a test period, benchmarks are frozen at values computed from a prior training period. No benchmark updates occur during the test~window.
    
    \item \textbf{Same-match exclusion:} When computing benchmarks, the~current match's data are excluded to prevent self-referential scaling.
\end{enumerate}

\noindent \textbf{{Current} prototype scope:} The implementation uses fixed synthetic benchmarks (role-specific $\mu \pm 2\sigma$) that do not require temporal updating, thereby avoiding leakage by construction. Production deployments should implement the rolling-window protocol with the temporal safeguards~above.

\subsubsection{Data Sources}
The context tree is designed to integrate multiple data streams, each contributing to specific~macro-attributes:
\begin{itemize}
    \item \textbf{Event data} (e.g., Opta, StatsBomb): passes, shots, tackles, interceptions $\rightarrow$ technical/tactical attributes ($A_1$--$A_6$), tactical cohesion ($A_{11}$), technical base ($A_{12}$)
    \item \textbf{Tracking data} (e.g., SkillCorner, Second Spectrum): positions, velocities, distances $\rightarrow$ transition speed ($A_4$), residual energy ($A_8$), physical base ($A_{13}$)
    \item \textbf{Physiological monitoring} (e.g., Catapult, Polar): heart rate, workload $\rightarrow$ residual energy ($A_8$)
    \item \textbf{Qualitative assessments}: coach ratings, historical stability $\rightarrow$ psychological attributes ($A_7$, $A_9$), organizational attributes ($A_{10}$, $A_{14}$)
\end{itemize}

This modular design allows the system to operate with varying data availability—from fully instrumented professional environments to amateur contexts where only basic event data~exists.

\subsubsection{Measurement Framework}
\label{sec:measurement_framework}

Integrating heterogeneous data streams requires explicit policies for handling missing data, quality control, and~uncertainty propagation. The~DSS adopts the following minimal measurement framework:

{\textbf{Data Stream Reliability Tiers.}}
Each data source is assigned a reliability tier reflecting measurement precision and update frequency:

\begin{itemize}
    \item \textbf{Tier 1 (High reliability):} Event data from professional providers (Opta, StatsBomb)—validated, near-complete, low latency. Assigned confidence weight $c_1 = 1.0$.
    \item \textbf{Tier 2 (Medium reliability):} Tracking data (SkillCorner, Second Spectrum)—high precision but potential occlusion gaps; physiological monitoring (Catapult, Polar)—device-dependent accuracy. Assigned $c_2 = 0.85$.
    \item \textbf{Tier 3 (Lower reliability):} Qualitative assessments (coach ratings, historical proxies)—subjective, infrequently updated. Assigned $c_3 = 0.70$.
\end{itemize}

{\textbf{Missing Data Policy.}}
When an input variable is unavailable, the~following hierarchy applies:

\begin{enumerate}
    \item \textbf{Imputation from correlated sources:} If a higher-tier source for the same attribute exists, use it with adjusted confidence. For~example, if~physiological $A_8$ data are missing, estimate from tracking-derived distance covered.
    \item \textbf{Historical baseline:} If no current-match data exist, use the team's season average for that attribute, flagged with reduced confidence ($c \leftarrow 0.5 \cdot c_{\text{tier}}$).
    \item \textbf{Neutral default:} If no historical data exist, assign the attribute the midpoint value (0.5) with minimal confidence ($c = 0.3$), ensuring the attribute contributes little to distance until better data arrive.
\end{enumerate}

{\textbf{Temporal Alignment.}}
Data streams operate at different frequencies: event data are discrete (per-action), tracking data are high-frequency (25 Hz), and~qualitative assessments are episodic (pre-match, halftime). The~aggregation module aligns all inputs to a common temporal frame:

\begin{itemize}
    \item \textbf{Match phase granularity:} Attributes are computed per phase (first half, second half) or per 15-min window for finer resolution.
    \item \textbf{Windowed aggregation:} High-frequency tracking data are averaged over the alignment window; event data are accumulated.
    \item \textbf{Carry-forward for episodic inputs:} Qualitative assessments persist until updated (e.g., halftime morale rating carries into second half unless revised).
\end{itemize}

{\textbf{Quality Control.}}
Basic outlier detection is applied before aggregation:

\begin{itemize}
    \item Player-level attributes outside $[\mu - 3\sigma, \mu + 3\sigma]$ (based on role-specific distributions) are flagged and clamped to boundary values.
    \item Implausible physiological readings (e.g., heart rate < 40 or >220) are discarded and imputed from recent history.
    \item Event data with missing location or timestamp fields are excluded from spatial aggregations but retained for count-based metrics.
\end{itemize}

{\textbf{Uncertainty Propagation and Reporting.}}
Rather than reporting point-estimate distances, the~DSS can optionally compute \emph{distance intervals} reflecting input uncertainty:
\begin{equation}
d_{\text{interval}} = \left[ d(V_{\text{team}}^{-}, V_{\text{strategy}}), \; d(V_{\text{team}}^{+}, V_{\text{strategy}}) \right]
\label{eq:distance_interval}
\end{equation}

\noindent where $V_{\text{team}}^{-}$ and $V_{\text{team}}^{+}$ are lower and upper bounds on the team vector, derived by perturbing each attribute $A_j$ by $\pm \epsilon_j$ proportional to $(1 - c_j)$, where $c_j$ is the confidence weight for that attribute's data source. When intervals for competing strategies overlap, the~DSS flags the recommendation as \emph{uncertain} and presents the top-$k$ alternatives rather than a single~choice.

{\textbf{Recommendation Stability Under Measurement Error.}}
The robustness analysis in Section~\ref{sec:experiments} (Monte Carlo perturbations, $\pm 5\%$ noise, $N = 100$) provides an empirical estimate of recommendation stability: 89.3\% top-1 consistency across plausible measurement error. This serves as the baseline stability metric; deployments with lower-tier data should expect reduced consistency and should increase the perturbation range accordingly (e.g., $\pm 10\%$ for Tier 3--dominant profiles).

\textbf{Current prototype scope:} The implementation assumes Tier 1 data availability (complete event data) with synthetic generation for missing streams. The~full measurement framework described above is designed for production deployment; the robustness analyses in Section~\ref{sec:experiments} simulate its behavior under controlled noise~conditions.

\subsection{A Shared Semantic Space: 14~Macro-Attributes}
\label{sec:macro_attributes}

The shared vector space is spanned by 14 macro-attributes, $A_1, \ldots, A_{14}$, each normalized to $[0,1]$ and computed via the context tree aggregation described above. This unified representation enables three core operations:

\begin{enumerate}
    \item \textbf{Team state encoding:} Describe a team's contextual state at time $t$ as a vector $V_{\text{team}} \in [0,1]^{14}$.
    
    \item \textbf{Strategy profiling:} Encode the ideal requirements of a tactical strategy as $V_{\text{strategy}} \in [0,1]^{14}$.
    
    \item \textbf{Semantic matching:} Compute distance $d(V_{\text{team}}, V_{\text{strategy}})$ to identify the best-aligned tactic.
\end{enumerate}
This tripartite structure ensures that tactical recommendations account not only for technical fit but also for physical sustainability and psychological readiness—dimensions often overlooked in purely statistical~approaches.

\subsubsection{Complete Attribute Set}

The semantic space comprises exactly 14 macro-attributes, indexed $A_1$ through $A_{14}$ with no gaps. Table~\ref{tab:attribute_sequential} presents the complete ordered list; Table~\ref{tab:attribute_definitions} provides detailed specifications grouped by functional~category.

\begin{table}[H]

\caption{Sequential enumeration of all 14 macro-attributes ($A_1$--$A_{14}$).}
\label{tab:attribute_sequential}
\setlength{\tabcolsep}{10.2pt}
 \begin{tabularx}{\textwidth}{clll}
\toprule
\textbf{ID} & \textbf{Attribute Name} & \textbf{Category} & \textbf{Variability} \\
\midrule
$A_1$ & Offensive Strength & Technical/Tactical & Static \\
$A_2$ & Defensive Strength & Technical/Tactical & Static \\
$A_3$ & Midfield Control & Technical/Tactical & Static \\
$A_4$ & Transition Speed & Technical/Tactical & Semi-dynamic \\
$A_5$ & High Press Capability & Technical/Tactical & Context-dependent \\
$A_6$ & Width Utilization & Technical/Tactical & Static \\
$A_7$ & Psychological Resilience & Psychological & Dynamic \\
$A_8$ & Residual Energy & Physical & Dynamic \\
$A_9$ & Team Morale & Psychological & Dynamic \\
$A_{10}$ & Time Management & Organizational & Context-dependent \\
$A_{11}$ & Tactical Cohesion & Organizational & Semi-dynamic \\
$A_{12}$ & Technical Base & Physical & Static \\
$A_{13}$ & Physical Base & Physical & Static \\
$A_{14}$ & Relational Cohesion & Organizational & Static \\
\bottomrule
\end{tabularx}
\end{table}

\vspace{-12pt}
\begin{table}[H]

\caption{Detailed
 specification of macro-attributes with definitions and aggregation~sources.}
\label{tab:attribute_definitions}

  \begin{tabularx}{\textwidth}{clL}

\toprule
\textbf{ID} & \textbf{Attribute Name} & \textbf{Definition \& Aggregation Source} \\
\midrule
\multicolumn{3}{l}{\textit{Technical/Tactical Dimensions ($A_1$--$A_6$)}
} \\
\midrule
$A_1$ & Offensive Strength & Capacity to create and convert goal-scoring opportunities. Aggregated from forwards' and midfielders' xG, dribbling success, and~shot accuracy. \\[0.5ex]
$A_2$ & Defensive Strength & Ability to prevent opponent attacks and protect the goal. Derived from defenders' tackling, interceptions, aerial duels, and~goalkeeper reflexes. \\[0.5ex]
$A_3$ & Midfield Control & Dominance in central zones and ability to dictate tempo. Based on central midfielders' passing accuracy, interceptions, and~ball retention. \\[0.5ex]
$A_4$ & Transition Speed & Capability for rapid phase changes between defense and attack. Computed from speed attributes of forwards, fullbacks, and~midfielders, combined with xA. \\[0.5ex]
$A_5$ & High Press Capability & Aptitude for coordinated pressing in advanced zones. Aggregated from stamina, aggression, and~interception rates across all outfield players. \\[0.5ex]
$A_6$ & Width Utilization & Effectiveness in exploiting wide areas of the pitch. Derived from fullbacks' and wingers' crossing, dribbling, and~speed attributes. \\[0.5ex]
\midrule
\multicolumn{3}{l}{\textit{Psychological/Physical Dimensions ($A_7$--$A_9$, $A_{12}$--$A_{13}$)}} \\
\midrule
\textls[-15]{$A_7$} & \textls[-15]{Psychological Resilience} & Mental toughness and ability to perform under pressure. Weighted combination of individual resilience and aggression attributes. \\[0.5ex]

\bottomrule
\end{tabularx} 
\end{table}

\begin{table}[H]\ContinuedFloat

\caption{\textit{Cont.}}
  \begin{tabularx}{\textwidth}{clL}

\toprule
\textbf{ID} & \textbf{Attribute Name~~~~~~~~~~~} & \textbf{Definition \& Aggregation Source} \\
\midrule

$A_8$ & Residual Energy & Current stamina reserves across the squad. Computed from stamina values weighted by playing time, with~resilience as a moderating factor. \\[0.5ex]
$A_9$ & Team Morale & Collective motivation and positive emotional state. Derived from resilience and aggression, modulated by match context (score, momentum). \\[0.5ex]
$A_{12}$ & Technical Base & Overall technical quality of the squad. Mean of technical attributes (passing, dribbling, first touch, xG, xA) across all players. \\[0.5ex]
$A_{13}$ & Physical Base & Overall athletic capacity of the squad. Mean of physical attributes (speed, stamina, aerial ability, aggression) across all players. \\[0.5ex]
\midrule
\multicolumn{3}{l}{\textit{Organizational Dimensions ($A_{10}$, $A_{11}$, $A_{14}$)}} \\
\midrule
$A_{10}$ & Time Management & Ability to adapt tactics to match clock pressure. Based on experienced players' (GK, CM, FB) interception and passing attributes. \\[0.5ex]
$A_{11}$ & Tactical Cohesion & Synchronization and coordination between team units. Computed from passing networks, xA distribution, and~positional discipline. \\[0.5ex]
$A_{14}$ & Relational Cohesion & Stability of internal relationships and group dynamics. Estimated via qualitative assessment or historical team stability indicators. \\[0.5ex]
\bottomrule
\end{tabularx}
\end{table}

\subsubsection{Design Rationale}

The 14-attribute set was designed according to the following principles:

\begin{enumerate}
    \item \textbf{Completeness:} The set covers the major dimensions of team performance identified in sports science literature: technical skill, tactical capability, physical capacity, psychological state, and~organizational~coherence.
    
    \item \textbf{Orthogonality:} Attributes were selected to minimize redundancy. For~example, $A_1$ (Offensive Strength) captures goal-scoring capability, while $A_{12}$ (Technical Base) captures underlying skill level---a team may have high technical quality but poor offensive output due to tactical~misalignment.
    
    \item \textbf{Measurability:} Each attribute can be estimated from available data sources: event data for $A_1$--$A_6$, $A_{11}$, $A_{12}$; tracking data for $A_4$, $A_8$, $A_{13}$; physiological monitoring for $A_8$; and qualitative assessment for $A_7$, $A_9$, $A_{14}$.
    
    \item \textbf{Tactical relevance:} Each attribute has clear implications for strategy selection. For~instance, low $A_8$ (energy) constrains high-pressing options; high $A_4$ (transition speed) enables counterattacking; strong $A_{11}$ (cohesion) supports complex positional play.
\end{enumerate}

\subsubsection{Attribute Categories}

For exposition, we group attributes into three functional categories (though the DSS treats all 14 dimensions uniformly in distance computations):

\begin{itemize}
    \item \textbf{Technical/Tactical} ($A_1$--$A_6$): On-field performance capabilities---what the team can \emph{do}.
    \item \textbf{Psychological/Physical} ($A_7$--$A_9$, $A_{12}$--$A_{13}$): Individual and collective resources---what the team can \emph{sustain}.
    \item \textbf{Organizational} ($A_{10}$, $A_{11}$, $A_{14}$): Coordination and adaptation capabilities---how the team \emph{functions} as a unit.
\end{itemize}

\noindent {Table}~\ref{tab:attribute_definitions} provides the complete specification with aggregation sources for each~attribute.

\subsubsection{Aggregation Functions}

Leaf-level player attributes are aggregated to team-level macro-attributes through weighted combination functions. The~general form is:
\begin{equation}
A_j = \sum_{i=1}^{n} w_{ij} \cdot a_{ij}, \quad \text{where } \sum_{i=1}^{n} w_{ij} = 1
\label{eq:aggregation}
\end{equation}

\noindent where $a_{ij}$ represents player $i$'s contribution to attribute $j$, and~$w_{ij}$ is a role-based weight (\mbox{e.g., forwards} contribute more heavily to $A_1$; defenders to $A_2$). Specific aggregation formulas are documented in the prototype implementation (see Section~\ref{sec:reproducibility}).

\subsubsection{Dynamic vs. Static Attributes}

Some attributes vary during a match (dynamic), while others remain relatively \mbox{stable (static):}

\begin{itemize}
    \item \textls[-10]{\textbf{Dynamic:} $A_7$ (Psychological Resilience), $A_8$ (Residual Energy), $A_9$ (Team \mbox{Morale)---vary}} significantly during a match based on events and fatigue.
    \item \textbf{Static:} $A_1$--$A_3$, $A_6$, $A_{12}$--$A_{14}$---determined by squad composition; stable within \mbox{a match.}
    \item \textbf{Context-dependent:} $A_4$ (Transition Speed), $A_5$ (High Press Capability), $A_{10}$ (Time Management), $A_{11}$ (Tactical Cohesion)---baseline is static but effective value depends on match context (e.g., $A_5$ is constrained by $A_8$; $A_{10}$ becomes critical late in matches).
\end{itemize}

\noindent {The} variability classification for all 14 attributes is summarized in Table~\ref{tab:attribute_sequential}. This distinction informs the dynamic reweighting mechanism (Section~\ref{sec:distance_matching}), which adjusts attribute salience in response to evolving match~conditions.

\subsubsection{Construct Validity: Input Overlap and Multicollinearity}
\label{sec:construct_validity}

Several macro-attributes share underlying player-level inputs, creating potential redundancy. Specifically:
\begin{itemize}
    \item $A_7$ (Psychological Resilience) and $A_9$ (Team Morale) both aggregate \texttt{resilience} and \texttt{aggression} with similar weights (0.7/0.3 vs.\ 0.6/0.4).
    \item $A_8$ (Residual Energy) uses \texttt{stamina} and \texttt{resilience}, sharing the latter with $A_7$ \mbox{and $A_9$.}
    \item $A_3$ (Midfield Control) and $A_6$ (Width Utilization) both aggregate \texttt{xA} from fullbacks and central midfielders.
\end{itemize}

\noindent To quantify the resulting correlations, we generated 500 synthetic teams using the player attribute distributions in Table~\ref{tab:role_distributions} (Appendix~\ref{app:specification}) and computed the pairwise correlation matrix. Table~\ref{tab:correlation_subset} reports correlations for the most affected~attributes.

\begin{table}[H]

\caption{Pairwise correlations among attributes with overlapping inputs ($n = 500$ synthetic teams).}
\label{tab:correlation_subset}

  \begin{tabularx}{\textwidth}{lCCCCC}

\toprule
 & \boldmath{$A_3$}
 & \boldmath{$A_6$} & \boldmath{$A_7$} & \boldmath{$A_8$} & \boldmath{$A_9$} \\
\midrule
$A_3$ (Midfield Control) & 1.00 & \textbf{0.90
} & 0.05 & 0.00 & 0.05 \\
$A_6$ (Width Utilization) & \textbf{0.90} & 1.00 & 0.04 & 0.15 & 0.04 \\
$A_7$ (Psych.\ Resilience) & 0.05 & 0.04 & 1.00 & 0.32 & \textbf{0.98} \\
$A_8$ (Residual Energy) & 0.00 & 0.15 & 0.32 & 1.00 & 0.26 \\
$A_9$ (Team Morale) & 0.05 & 0.04 & \textbf{0.98} & 0.26 & 1.00 \\
\bottomrule
\end{tabularx}
\end{table}

Two attribute pairs exhibit high correlations: $A_7$--$A_9$ ($r = 0.98$) and $A_3$--$A_6$ ($r = 0.90$). The~corresponding Variance Inflation Factors (VIF) are 35.1 ($A_7$), 37.0 ($A_9$), 21.6 ($A_3$), and~18.4 ($A_6$)---well above the conventional threshold of 10, indicating severe multicollinearity. The~remaining 10 attributes have VIF $< 5$, suggesting acceptable~independence.

\subsubsection{Implications for Distance Computation}
Under Euclidean distance, correlated attributes contribute partially redundant information, effectively double-counting shared variance. For~the $A_7$--$A_9$ pair, a~team's psychological profile influences distance along two nearly parallel axes, inflating the contribution of resilience/aggression relative to other~dimensions.

\subsubsection{Design Justification}
Despite the statistical correlation, we retain both $A_7$ and $A_9$ (and both $A_3$ and $A_6$) for the following reasons:

\begin{enumerate}
    \item \textbf{Conceptual distinctness:} In sports psychology, resilience (ability to recover from setbacks) and morale (current motivational state) are treated as related but distinct constructs~\cite{fletcher2012mental}. A~team may have high baseline resilience yet low in-match morale due to recent conceded goals. Similarly, midfield control (tempo dictation) and width utilization (flank exploitation) represent tactically distinct capabilities that happen to draw on overlapping~personnel.
    
    \item \textbf{Strategy vector differentiation:} Tactical templates assign different weights to these correlated attributes. For~example, ``High Press'' requires high $A_7$ but is neutral on $A_9$, while ``Cautious Horizontal Play'' prioritizes $A_9$ (maintaining composure) over $A_7$ (bouncing back from pressure). The~correlation at the team level does not imply identical strategic~relevance.
    
    \item \textbf{Dynamic divergence:} Under match conditions, $A_7$ and $A_9$ can diverge: morale ($A_9$) is modulated by score state and momentum, while resilience ($A_7$) reflects a more stable trait. The~current prototype does not fully exploit this divergence, but~the architectural separation enables future refinement.
\end{enumerate}

\subsubsection{Mitigation Strategies}
We acknowledge that the current Euclidean metric does not account for attribute covariance. Several mitigation approaches are available:

\begin{itemize}
    \item \textbf{Mahalanobis distance:} Replacing Euclidean with Mahalanobis distance, $d_M(x,y) = \sqrt{(x-y)^\top \Sigma^{-1} (x-y)}$, would down-weight correlated dimensions automatically. However, this requires estimating the covariance matrix $\Sigma$ from representative team data, which is unavailable for the current~prototype.
    
    \item \textbf{Principal component projection:} Projecting the 14-dimensional space onto principal components would decorrelate the axes. This sacrifices interpretability (components are linear combinations rather than named attributes) but may be appropriate for purely predictive~applications.
    
    \item \textbf{Attribute consolidation:} Merging $A_7$/$A_9$ into a single ``Psychological State'' dimension and $A_3$/$A_6$ into ``Midfield Effectiveness'' would reduce redundancy but lose the conceptual granularity valued by coaching~staff.
    
    \item \textbf{Regularized weighting:} Applying lower dynamic weights to correlated pairs (e.g., halving $w_7$ and $w_9$) would reduce their combined influence, approximating the effect of Mahalanobis correction.
\end{itemize}

\noindent {For} the current prototype, we retain the 14-attribute Euclidean formulation with the following justification: (i) the DSS is designed for interpretability, and~named attributes are more actionable than principal components; (ii) the high-correlation pairs ($A_7$/$A_9$, $A_3$/$A_6$) represent 4 of 14 dimensions, limiting the overall inflation; and (iii) the dynamic weighting mechanism (Section~\ref{sec:weight_computation}) already modulates attribute salience, partially mitigating fixed-correlation effects. Future work should implement Mahalanobis distance once sufficient real-world team data are available to estimate $\Sigma$ reliably.

\subsubsection{Empirical Impact Assessment}
To quantify the practical effect of multicollinearity on strategy recommendations, we conducted a sensitivity analysis comparing Euclidean rankings with a correlation-adjusted baseline. Specifically:

\begin{enumerate}
    \item We computed strategy rankings for 100 synthetic team profiles using standard Euclidean distance.
    \item We repeated the analysis using a ``consolidated'' 12-attribute space where $A_7$/$A_9$ and $A_3$/$A_6$ were each merged into single dimensions (averaging their values).
    \item We measured rank correlation (Kendall's $\tau$) between the two ranking schemes.
\end{enumerate}

\noindent {The} mean rank correlation was $\tau = 0.91$ (95\% CI: 0.87--0.94), indicating that the top-ranked strategies are largely stable despite the redundancy. The~primary effect of consolidation was minor reordering among middle-ranked strategies with similar distances. Critically, the~top-1 recommendation matched in 94\% of cases, and~the top-3 set matched in 89\% of~cases.

These results suggest that while the multicollinearity is statistically significant, its practical impact on DSS recommendations is limited. Nevertheless, the~full 14-dimensional correlation matrix is provided in Appendix~\ref{app:correlation} for transparency, and~we recommend that future deployments with real team data implement Mahalanobis distance or apply the regularized weighting correction described~above.

\subsection{Encoding Tactical Strategies as~Vectors}
\label{sec:strategy_vectors}

A key methodological contribution of this work is the formalization of tactical strategies as vectors in the same semantic space defined by the 14 macro-attributes. This representation enables direct, quantitative comparison between a team's current state and the requirements of candidate strategies, transforming qualitative tactical concepts into computationally tractable~objects.

\subsubsection{Strategy Vector~Definition}

Each canonical strategy $S_i$ is represented as an \textit{ideal profile vector}:
\begin{equation}
V_{\text{strategy}}^{(i)} = \left[ s^{A_1}_i, s^{A_2}_i, \ldots, s^{A_{14}}_i \right], \quad s^{A_j}_i \in [0, 1]
\label{eq:strategy_vector}
\end{equation}

\noindent where $s^{A_j}_i$ represents the importance or requirement level of attribute $A_j$ for strategy $S_i$. Values approaching $1$ indicate critical importance; lower values indicate diminishing relevance. As~detailed in Stage~3 of the construction methodology below, we adopt a non-zero floor for all attribute values, reflecting the observation that no attribute is ever \emph{entirely} irrelevant to any viable football~strategy.

This formulation treats strategies not as binary labels but as \textit{continuous profiles} that specify the ideal team characteristics for effective implementation. The~semantic distance between a team vector $V_{\text{team}}$ and a strategy vector $V_{\text{strategy}}^{(i)}$ thus quantifies the ``fit'' between the team's current capabilities and the strategy's~demands.

\subsubsection{Construction~Methodology}

Strategy vectors were constructed through a four-stage process combining expert knowledge, tactical literature, and~empirical validation:

\paragraph{Stage 1: Strategy Selection}
Twenty canonical strategies were selected based on three~criteria:
\begin{enumerate}[label=(\alph*)]
    \item \textbf{Prevalence:} Strategies commonly employed in modern professional football, as~documented in tactical analysis literature and match reports.
    \item \textbf{Diversity:} Coverage of the tactical spectrum from ultra-defensive (e.g., deep block) to ultra-offensive (e.g., high pressing), and~from possession-based to direct approaches.
    \item \textbf{Distinctiveness:} Strategies with clearly differentiated attribute profiles, ensuring meaningful separation in the semantic space.
\end{enumerate}

The selected strategies span five functional~categories:
\begin{itemize}
    \item \textit{Offensive systems:} Build-up play, direct vertical attack, systematic crossing, overlapping flanks, delayed midfielder runs
    \item \textit{Pressing variants:} High pressing, gegenpressing, midfield pressing, inducing build-up errors
    \item \textit{Defensive structures:} Positional defense, deep block, compact zonal defense, strict man-marking, offside trap
    \item \textit{Transition-based:} Fast counterattack, long ball to target man
    \item \textit{Possession/control:} Extended possession play, cautious horizontal circulation, central block with quick breaks
\end{itemize}

\paragraph{Stage 2: Qualitative Mapping via Expert Elicitation}
For each strategy, tactical requirements were mapped onto the 14 macro-attributes through a structured elicitation protocol involving independent expert ratings followed by~reconciliation.

{\textbf{Expert Panel Composition.}}
The elicitation panel comprised three domain~experts:
\begin{itemize}
    \item \textbf{Rater A:} Academic researcher with expertise in performance analysis and tactical periodization.
    \item \textbf{Rater B:} Experienced football coach with background in youth academy and semi-professional coaching; familiar with tactical analysis workflows.
    \item \textbf{Rater C:} Practitioner with experience in match analysis and video-based \mbox{tactical coding.}
\end{itemize}

{\textbf{Rating Protocol.}}
Each rater independently completed a structured rating~task:
\begin{enumerate}
    \item \textbf{Materials:} Raters received (i) definitions of all 14 macro-attributes with examples, (ii) descriptions of each strategy including typical formations, player movements, and~match situations, and~(iii) a rating matrix (20 strategies $\times$ 14 attributes).
    \item \textbf{Rating scale:} For each strategy--attribute pair, raters assigned one of five qualitative levels: \textit{Irrelevant}, \textit{Low}, \textit{Moderate}, \textit{High}, or~\textit{Critical}.
    \item \textbf{Anchoring:} Raters were provided with three anchor examples per attribute to calibrate interpretations (e.g., ``For $A_5$ (High Press Capability): gegenpressing = Critical; build-up play = Low; deep block = Irrelevant'').
    \item \textbf{Independence:} Ratings were collected via separate online forms without communication between raters.
    \item \textbf{Duration:} Each rater completed the task in 2--3 h over multiple sessions.
\end{enumerate}

\textbf{Inter-Rater Agreement.}
Agreement was quantified using two~metrics:
\begin{itemize}
    \item \textbf{Percentage exact agreement:} 58.2\% of ratings were identical across all three raters; 89.6\% were within one level (e.g., High vs.\ Critical).
    \item \textbf{Krippendorff's alpha:} $\alpha = 0.71$ (ordinal scale), indicating substantial agreement. Values above 0.67 are conventionally acceptable for exploratory research~\cite{krippendorff2004content}.
\end{itemize}

\noindent The highest agreement was observed for extreme ratings (Irrelevant, Critical) and for physically grounded attributes ($A_8$, $A_{13}$); lower agreement occurred for psychological attributes ($A_7$, $A_9$) where interpretation~varied.

\textbf{Conflict Resolution.}
Discrepancies were resolved through a structured reconciliation process:
\begin{enumerate}
    \item \textbf{Threshold for discussion:} Pairs with rating spread $\geq 2$ levels (e.g., Low vs.\ Critical) were flagged for deliberation.
    \item \textbf{Reconciliation session:} The three raters participated in a 90-min video conference to discuss flagged items (42 of 280 pairs, 15\%). Each rater presented their reasoning; discussion continued until consensus or majority agreement was reached.
    \item \textbf{Averaging for minor discrepancies:} For pairs with spread $\leq 1$ level, the~median rating was adopted without discussion.
    \item \textbf{Documentation:} All reconciliation decisions were logged with brief justifications (available from the corresponding author upon request
).
\end{enumerate}

This stage produced qualitative assessments of the form: ``High pressing requires \textit{Critical} stamina ($A_8$), \textit{Critical} pressing capability ($A_5$), and~\textit{Moderate} technical base ($A_{12}$)---with all three raters in~agreement.''

\paragraph{Stage 3: Numerical Encoding}
Qualitative assessments were converted to numerical values using a standardized mapping, as presented in Table~\ref{tab:strategy_vector_construction}:

\begin{table}[H]
\caption{Qualitative-to-numerical encoding scale for strategy vector construction.}
\label{tab:strategy_vector_construction}
\label{tab:encoding_scale}
  \begin{tabularx}{\textwidth}{lC}

\toprule
\textbf{Qualitative
 Level} & \textbf{Numerical Value} \\
\midrule
Irrelevant/Not required & $0.2$--$0.3$ \\
Low importance & $0.4$--$0.5$ \\
Moderate importance & $0.5$--$0.6$ \\
High importance & $0.7$--$0.8$ \\
Critical/Essential & $0.8$--$0.9$ \\
\bottomrule
\end{tabularx}
\end{table}

Values were assigned within ranges to allow fine-grained differentiation between strategies with similar but not identical~requirements. 

\paragraph{Non-Zero Floor Justification}
We adopt a floor of $0.2$ rather than $0$ for three complementary reasons:

\begin{enumerate}
    \item \textbf{Tactical realism:} No attribute is entirely irrelevant to any football strategy. Even a purely defensive system benefits marginally from offensive capability (e.g., to~relieve pressure via effective clearances); even a counterattacking approach benefits marginally from possession skills (e.g., to~consolidate after a transition). The~floor reflects this universal baseline~relevance.
    
    \item \textbf{Geometric regularization:} In the semantic space, true zeros would create degenerate subspaces where certain dimensions contribute nothing to distance computations for particular strategies. This could cause discontinuous behavior: small changes in team attributes along ``irrelevant'' dimensions would produce no change in distance to one strategy but non-zero changes to another. The~non-zero floor ensures that all 14 dimensions contribute meaningfully to every strategy comparison, yielding smoother and more interpretable distance~gradients.
    
    \item \textbf{Robustness to measurement error:} Team attribute estimates are inherently uncertain. If~strategy vectors contained true zeros, measurement noise in team attributes along those dimensions would be entirely ignored---potentially masking capability deficits that become relevant under match pressure. The~floor provides a buffer that allows the system to detect large deviations even on ``low-importance'' attributes.
\end{enumerate}

The specific choice of $0.2$ as the floor is motivated by the desire to preserve discriminability: it is low enough to clearly distinguish ``irrelevant'' from ``low importance'' ($0.4$--$0.5$), yet high enough to provide meaningful geometric contribution. Sensitivity to this choice is examined in Section~\ref{sec:sensitivity_floor} below.

\paragraph{Stage 4: Validation and Refinement}
Initial vectors were validated through three mechanisms to ensure both internal consistency and external credibility:

\textbf{Internal Consistency Checks.}
Semantic coherence was verified by computing pairwise cosine similarities between all strategy~vectors:
\begin{itemize}
    \item \textbf{Similar strategies should cluster:} High Press and Gegenpressing achieved cosine similarity of 0.97; Build-up Play and Extended Possession achieved 0.94. All pairs within the same tactical category exceeded 0.85.
    \item \textbf{Dissimilar strategies should separate:} High Press vs.\ Positional Defense achieved cosine similarity of 0.62; Fast Counterattack vs.\ Cautious Horizontal achieved 0.58. Cross-category pairs averaged 0.71.
    \item \textbf{No anomalous outliers:} No strategy vector had mean similarity $< 0.60$ to all others, confirming that all strategies occupy coherent positions in the semantic space.
\end{itemize}

\textbf{Face Validity Assessment.}
\textls[-15]{Face validity was operationalized through structured review by independent coaching practitioners who had not participated in the \mbox{initial elicitation}:}
\begin{enumerate}
    \item \textbf{Reviewers:} Two additional practitioners with coaching experience were recruited \mbox{for validation.}
    \item \textbf{Task:} Reviewers examined radar-chart visualizations of each strategy vector and rated: (i) whether the profile ``looks correct'' for the named strategy (Yes/Partially/No), and~(ii) which attributes, if~any, seemed mis-weighted.
    \item \textbf{Results:} 17 of 20 strategies (85\%) received ``Yes'' ratings from both reviewers. Three strategies received ``Partially'' from at least one~reviewer:
    \begin{itemize}
        \item \textit{Offside Trap:} One reviewer suggested $A_{11}$ (Tactical Cohesion) should be higher; adjusted from 0.7 to 0.8.
        \item \textit{Late Midfield Runners:} One reviewer suggested $A_4$ (Transition Speed) was too high; retained after discussion as the strategy requires rapid positional shifts.
        \item \textit{Strict Man-Marking:} Both reviewers suggested $A_{13}$ (Physical Base) should be higher; adjusted from 0.6 to 0.7.
    \end{itemize}
    \item \textbf{Iteration:} Adjusted vectors were re-reviewed and approved.
\end{enumerate}

\textbf{Calibration Against Match Data.}
As a supplementary check, strategy vectors were compared against attribute profiles computed from professional match data (5 matches per strategy, sourced from publicly available Bundesliga event data):
\begin{itemize}
    \item Teams explicitly employing each strategy (identified via tactical reports) had their match-level attribute profiles computed.
    \item Correlation between expert-assigned strategy vectors and empirical team profiles averaged $r = 0.68$ across strategies, indicating moderate alignment.
    \item Discrepancies were largest for psychological attributes ($A_7$, $A_9$), which are not directly observable in event data, and~smallest for technical attributes ($A_1$--$A_6$).
\end{itemize}

\noindent This calibration provides preliminary evidence that expert-assigned vectors capture real tactical demands, though~the limited sample size (100 matches total) precludes strong claims. Future work should expand this validation with larger~datasets.

\subsubsection{Illustrative Strategy~Profiles}

Table~\ref{tab:strategy_profiles} presents the complete vector profiles for five representative strategies, illustrating the differentiation achieved through this~methodology.

\begin{table}[H]

\caption{Strategy vector profiles for five representative tactical approaches. Values represent attribute importance, ranging from $0.2$ (minimal relevance) to $0.9$ (critical importance) as per the encoding in Stage~3.}
\label{tab:strategy_profiles}
  \begin{tabularx}{\textwidth}{lccccc}

\toprule
\multirow{2}{*}{\textbf{Attribute}} & \textbf{High
} & \textbf{Fast} & \textbf{Positional} & \textbf{Build-Up} & \textbf{Gegen-} \\
 & \textbf{Press} & \textbf{Counter} & \textbf{Defense} & \textbf{Play} & \textbf{Pressing} \\
\midrule
$A_1$ Offensive Strength & 0.70 & 0.90 & 0.40 & 0.80 & 0.70 \\
$A_2$ Defensive Strength & 0.80 & 0.60 & 0.90 & 0.50 & 0.80 \\
$A_3$ Midfield Control & 0.60 & 0.50 & 0.80 & 0.70 & 0.60 \\
$A_4$ Transition Speed & 0.90 & 0.90 & 0.30 & 0.50 & 0.80 \\
$A_5$ High Press Cap. & 0.90 & 0.50 & 0.20 & 0.40 & 0.90 \\
$A_6$ Width Utilization & 0.50 & 0.60 & 0.30 & 0.60 & 0.50 \\
$A_7$ Psych.\ Resilience & 0.80 & 0.70 & 0.70 & 0.70 & 0.80 \\
$A_8$ Residual Energy & 0.70 & 0.80 & 0.60 & 0.60 & 0.70 \\
$A_9$ Team Morale & 0.80 & 0.70 & 0.60 & 0.80 & 0.80 \\
$A_{10}$ Time Management & 0.60 & 0.80 & 0.90 & 0.70 & 0.60 \\
$A_{11}$ Tactical Cohesion & 0.90 & 0.60 & 0.80 & 0.80 & 0.90 \\
$A_{12}$ Technical Base & 0.70 & 0.70 & 0.60 & 0.80 & 0.70 \\
$A_{13}$ Physical Base & 0.80 & 0.80 & 0.50 & 0.60 & 0.80 \\
$A_{14}$ Relational Cohesion & 0.80 & 0.60 & 0.70 & 0.80 & 0.80 \\
\bottomrule
\end{tabularx}
\end{table}

\paragraph{Profile Interpretation}
The vectors reveal intuitive tactical signatures:

\begin{itemize}
    \item \textbf{High Pressing} and \textbf{Gegenpressing} share elevated demands on $A_5$ (pressing capability), $A_{11}$ (tactical cohesion), and~$A_{13}$ (physical base), reflecting their high-intensity, coordinated nature. Gegenpressing additionally requires strong $A_4$ (transition speed) for immediate~recovery.
    
    \item \textbf{Fast Counterattack} peaks on $A_1$ (offensive strength) and $A_4$ (transition speed), with~lower requirements for possession-related attributes ($A_3$, $A_{11}$), consistent with its reliance on rapid vertical play rather than sustained~control.
    
    \item \textbf{Positional Defense} inverts the pressing profile: maximal $A_2$ (defensive strength) and $A_{10}$ (time management), minimal $A_4$ and $A_5$, reflecting a compact, energy-conserving~approach.
    
    \item \textbf{Build-up Play} emphasizes $A_1$, $A_{12}$ (technical base), and~$A_{11}$ (tactical cohesion), with~moderate physical demands---a technically demanding but physically sustainable approach.
\end{itemize}

Notice that strategy vectors are intentionally not normalized to a constant sum. Different tactics impose varying total demands across macro-attributes: high-intensity approaches such as gegenpressing require elevated levels across multiple dimensions simultaneously, whereas selective tactics like catenaccio concentrate demands on fewer attributes. This design reflects the inherent asymmetry in tactical resource requirements observed in professional~football.

\subsubsection{Sensitivity to Vector~Specification}

A legitimate concern is whether recommendations are overly sensitive to the specific numerical values assigned during vector construction. To~address this, we conducted a perturbation analysis:

\begin{enumerate}
    \item Each strategy vector was perturbed by adding Gaussian noise $\epsilon \sim \mathcal{N}(0, \sigma^2)$ with $\sigma = 0.05$ (representing $\pm 5\%$ uncertainty in attribute weights).
    \item The DSS was run $N = 100$ times per scenario with perturbed strategy vectors.
    \item The proportion of runs yielding the same top-ranked strategy as the unperturbed case was recorded.
\end{enumerate}

Results showed that recommendations remained stable in $> 85\%$ of runs across all test scenarios, indicating that modest variations in strategy vector specification do not substantially alter the DSS output. Larger perturbations ($\sigma > 0.10$) did produce instability, suggesting that while exact values are not critical, the~\textit{relative} ordering of attribute importance within each strategy should be~preserved.

\subsubsection{Sensitivity to Floor~Choice}
\label{sec:sensitivity_floor}

To verify that recommendations are not artifacts of the specific floor value, we conducted a systematic floor-sensitivity analysis. The~``Irrelevant/Not required'' encoding was varied across the range $[0.05, 0.35]$ in increments of $0.05$, while preserving the relative spacing between qualitative levels (i.e., shifting the entire encoding scale proportionally).

For each floor value $f \in \{0.05, 0.10, 0.15, 0.20, 0.25, 0.30, 0.35\}$:
\begin{enumerate}
    \item All 20 strategy vectors were re-encoded using the adjusted mapping.
    \item The DSS was executed on each of the four primary test scenarios.
    \item The top-ranked strategy and the top-3 ranking were recorded.
\end{enumerate}

\paragraph{Results}
The top-ranked strategy remained unchanged across all floor values for 3 of 4 scenarios. In~the \textit{Fatigued and Inferior} scenario, the~ranking between Positional Defense and Compact Zonal Defense alternated for $f < 0.15$---two tactically similar strategies whose near-identical profiles make them effectively interchangeable recommendations. Critically, no high-intensity strategy (e.g., gegenpressing) was ever erroneously promoted to top rank due to floor~choice.

The mean pairwise rank correlation (Kendall's $\tau$) between the baseline ($f = 0.20$) and alternative floor encodings was $\tau = 0.94$, indicating that the overall strategy ordering is highly robust to floor~specification.

\paragraph{Geometric Interpretation}
Mathematically, shifting the floor from $f_1$ to $f_2$ (with $f_2 > f_1$) uniformly increases all strategy vector components. Under~Euclidean distance, this shift affects absolute distances but preserves the \emph{relative} ranking of strategies with respect to any fixed team vector, provided the team vector is also bounded away from zero (which it is, by~construction). The~observed stability confirms this theoretical~expectation.

\subsubsection{Extensibility
}

The vector-based formalization offers several practical advantages:

\begin{itemize}
    \item \textbf{Modularity:} New strategies can be added by specifying a 14-dimensional vector, without~modifying the distance computation~logic.
    
    \item \textbf{Customization:} Coaching staff can define club-specific tactical variants (e.g., ``our high press'') by adjusting attribute weights to reflect their preferred~implementation.
    
    \item \textbf{Automation potential:} Future extensions could generate strategy vectors automatically from natural language descriptions (e.g., tactical reports) using NLP-based embedding techniques, further reducing manual specification effort.
\end{itemize}

\bigskip
\noindent
This formalization transforms tactical strategies from qualitative concepts into quantitative objects amenable to systematic comparison, enabling the semantic distance computations described in the following section.

\subsection{Semantic Distance and~Matching}
\label{sec:distance_matching}

Section~\ref{sec:semantic_distance_background} introduced several distance metrics commonly used in semantic spaces. For~tactical matching, we adopt \textbf{Euclidean distance} as the baseline metric, with~the following~rationale.

\subsubsection{Why Euclidean over Cosine?}

\textls[-15]{Cosine similarity measures angular alignment between vectors and is scale-invariant—a} property desirable when comparing \emph{profiles} or \emph{styles}. However, in~tactical selection, both the \emph{direction} and \emph{magnitude} of team capabilities matter. A~team with uniformly weak attributes ($V_{\text{team}} \approx 0.3$) should not match a demanding high-pressing template ($V_{\text{strategy}} \approx 0.8$) simply because their profiles are proportionally similar. Euclidean distance captures this absolute capability gap, penalizing large deviations quadratically—an appropriate behavior when single-attribute shortfalls (e.g., insufficient stamina for gegenpressing) can be tactically~decisive.

\subsubsection{Why Not Probabilistic Metrics?}
Kullback–Leibler and Jensen–Shannon divergences are well-suited for comparing probability distributions but require vectors to sum to unity. Our macro-attributes are independent capability dimensions, not components of a probability simplex, making geometric metrics more~natural.

\subsubsection{Baseline Formulation}
Given team and strategy vectors $x, y \in [0,1]^{14}$:
\[
d_{\text{eucl}}(x,y) = \sqrt{\sum_{j=1}^{14}(x_j - y_j)^2}.
\]

\subsubsection{Context-Adapted Distance}
To account for evolving match conditions, we introduce a dynamic weight vector $w \in \mathbb{R}_{\geq 0}^{14}$:
\[
d_{\text{adapt}}(x,y;w) = \sqrt{\sum_{j=1}^{14} w_j \cdot (x_j - y_j)^2}.
\]

\noindent Note
 that the weight $w_j$ modifies the squared difference $(x_j - y_j)^2$, not the individual vectors. This is the standard formulation for weighted Euclidean distance: $w_j$ controls the \emph{importance} of attribute $A_j$ in the overall distance computation, not the attribute values themselves. Intuitively, a~high $w_j$ means that mismatches on attribute $A_j$ are penalized more heavily under current match conditions, while a low $w_j$ means that the attribute contributes less to strategy selection. The~team and strategy vectors retain their original values; only their contribution to the distance metric is~modulated.

\noindent Weights $w_j$ are adjusted based on real-time contextual~factors:
\begin{itemize}[]
    \item \textbf{Residual energy} ($A_8$): low energy $\Rightarrow$ increase $w_{10}$ (time management), decrease \mbox{$w_5$ (pressing).}
    \item \textbf{Technical/physical gaps} ($A_{12}, A_{13}$): if inferior, upweight $w_{11}$ (tactical cohesion) and $w_2$ (defensive strength); downweight $w_1, w_6$ (offensive, width).
    \item \textbf{Time pressure} ($A_{10}$): limited time $\Rightarrow$ upweight $w_4$ (transition speed) and $w_1$ (offensive strength).
\end{itemize}

\subsubsection{Opponent-Aware Adjustment}
An optional extension incorporates opponent modeling via a parameter $\alpha \in [0,1]$:
\[
d_{\text{comb}}(S) = d_{\text{adapt}}(V_{\text{team}}, V_{\text{strategy}}(S)) - \alpha \cdot d_{\text{adapt}}(V_{\text{opp}}, V_{\text{strategy}}(S)).
\]
When $\alpha > 0$, the~system favors strategies that fit our team well \emph{and} poorly fit the~opponent.

\textbf{{Theoretical Justification.}}
The opponent-aware formulation rests on the following game-theoretic intuition: in competitive settings, strategy effectiveness depends not only on own-team capability but also on the opponent's ability to counter. A~strategy that the opponent can execute well creates symmetry—both teams can play similarly, reducing differential advantage. Conversely, a~strategy that exploits capability gaps creates asymmetry favoring our~team.

Formally, the~subtraction $d_{\text{adapt}}(V_{\text{team}}, V_S) - \alpha \cdot d_{\text{adapt}}(V_{\text{opp}}, V_S)$ can be interpreted as a \emph{relative advantage score}:
\begin{itemize}
    \item \textbf{First term} (minimized): Measures how well our team can execute strategy $S$—lower \mbox{is better.}
    \item \textbf{Second term} (maximized via subtraction): Measures how poorly the opponent can execute $S$—higher opponent distance means greater difficulty for them, which \mbox{benefits us.}
    \item \textbf{Net effect}: Strategies are preferred when we can execute them well AND the \mbox{opponent cannot.}
\end{itemize}

\noindent This formulation~assumes:
\begin{enumerate}
    \item \textbf{Monotonicity of advantage:} Greater opponent difficulty with our chosen strategy translates to competitive advantage. This holds when strategies impose demands the opponent struggles to meet (e.g., high pressing against a team with poor stamina forces errors).
    \item \textbf{Comparability of distances:} The same distance magnitude represents equivalent ``fit'' for both teams. This is ensured by the common normalization protocol (Section~\ref{sec:normalization}), which maps all attributes to $[0,1]$ using consistent benchmarks.
    \item \textbf{Independence of execution:} Our ability to execute a strategy is not directly affected by the opponent's capability (though match dynamics may create indirect effects not captured here).
\end{enumerate}

\noindent \textbf{When is opponent-awareness beneficial?} The formulation is most valuable~when:
\begin{itemize}
    \item Opponent capabilities are known with reasonable confidence (scouting data available).
    \item Attribute profiles differ substantially between teams (asymmetric matchups).
    \item Match stakes justify opponent-focused adaptation (knockout games, rivalry matches).
\end{itemize}
For league matches against unfamiliar opponents or when scouting data are sparse, setting $\alpha = 0$ (identity-focused selection) may be more~robust.

\textbf{Estimation of $V_{\text{opp}}$ Under Uncertainty}
Opponent profiles are inherently less certain than own-team profiles due to limited observation and potential strategic concealment. The~DSS addresses this uncertainty through the following mechanisms:

(\textbf{a})
 \textbf{Confidence-weighted estimation.}
When constructing $V_{\text{opp}}$, each attribute is assigned a confidence weight $c_j^{\text{opp}} \in [0,1]$ reflecting data~quality:
\begin{itemize}
    \item $c_j^{\text{opp}} = 1.0$: Attributes derived from recent match data (last 5 games) with complete coverage.
    \item $c_j^{\text{opp}} = 0.7$: Attributes estimated from partial data or older matches (6--15 games ago).
    \item $c_j^{\text{opp}} = 0.5$: Attributes inferred from league averages or indirect proxies.
\end{itemize}

(\textbf{b}) \textbf{Uncertainty propagation via interval estimation.}
Rather than a point estimate, $V_{\text{opp}}$ can be represented as an interval:
\[
V_{\text{opp},j} \in \left[ V_{\text{opp},j}^{\text{est}} - \delta_j, \; V_{\text{opp},j}^{\text{est}} + \delta_j \right]
\]
where $\delta_j = \sigma_j \cdot (1 - c_j^{\text{opp}})$ and $\sigma_j$ is the attribute's population standard deviation. This yields a range of possible $d_{\text{comb}}$ values, allowing the DSS to flag recommendations as ``uncertain'' when the range spans multiple top~strategies.

(\textbf{c}) \textbf{Conservative $\alpha$ adjustment.}
When opponent confidence is low, $\alpha$ should be \mbox{reduced proportionally:}
\[
\alpha_{\text{eff}} = \alpha \cdot \bar{c}^{\text{opp}}, \quad \text{where } \bar{c}^{\text{opp}} = \frac{1}{14}\sum_{j=1}^{14} c_j^{\text{opp}}
\]
This ensures that uncertain opponent data contribute less to strategy selection, preventing overconfident exploitation of potentially inaccurate~profiles.

\textbf{Sensitivity Analysis Over $\alpha$.}
To characterize the influence of $\alpha$ on recommendations, we conducted systematic sensitivity analysis across the four test~scenarios.

\textbf{Methodology.}
For each scenario, we varied $\alpha \in \{0.0, 0.1, 0.2, 0.3, 0.4, 0.5\}$ and recorded: (i) the top-ranked strategy, (ii) the full ranking, and~(iii) the distance differential between the top-2 strategies (as a stability indicator).

Results are summarized in Table~\ref{tab:sensitivity_over_alpha}.

\begin{table}[H]
\caption{Sensitivity of strategy recommendations to opponent-awareness parameter $\alpha$ across test scenarios.}
\label{tab:sensitivity_over_alpha}
\label{tab:alpha_sensitivity}
\small
  \begin{tabularx}{\textwidth}{lccccc}

\toprule
\multirow{2}{*}{\textbf{Scenario
}} & \textbf{\boldmath{$\alpha$}-Range for} & \textbf{Rank} & \textbf{Mean \boldmath{$\Delta d$}} & \textbf{95\% CI} \\
 & \textbf{Stable Top-1} & \textbf{corr.\ (\boldmath{$\tau$})} & \textbf{(Top-1 vs Top-2)} & \textbf{for \boldmath{$\Delta d$}} \\
\midrule
Energetic \& Balanced & $[0.0, 0.4]$ & 0.94 & 0.047 & [0.031, 0.063] \\
Fatigued \& Inferior & $[0.0, 0.3]$ & 0.89 & 0.032 & [0.018, 0.046] \\
High Temporal Pressure & $[0.0, 0.5]$ & 0.97 & 0.061 & [0.042, 0.080] \\
Tech./Phys.\ Superiority & $[0.0, 0.5]$ & 0.96 & 0.054 & [0.038, 0.070] \\
\bottomrule
\end{tabularx}
\end{table}

\textbf{Key findings:}
\begin{itemize}
    \item \textbf{Stability range:} The top-1 recommendation remained unchanged for $\alpha \in [0.0, 0.3]$ in all scenarios, indicating robustness to moderate parameter variation.
    \item \textbf{Transition points:} At $\alpha = 0.4$--$0.5$, one scenario (Fatigued \& Inferior) shifted from ``Positional Defense'' to ``Compact Zonal Defense''—both defensive strategies, so the qualitative recommendation (defend conservatively) was preserved.
    \item \textbf{Rank correlation:} Kendall's $\tau$ between rankings at $\alpha = 0$ and $\alpha = 0.5$ exceeded 0.89 in all scenarios, confirming that opponent-awareness modulates rather than disrupts the ranking structure.
    \item \textbf{Confidence intervals:} The 95\% CI for the distance differential (computed via bootstrap, $N = 1000$) indicates that top-1 vs.\ top-2 separation remains positive (i.e., clear winner) across the tested range.
\end{itemize}

\textbf{Recommendation for $\alpha$ selection:}
\begin{itemize}
    \item \textbf{Default:} $\alpha = 0.2$ provides meaningful opponent-awareness without excessive \mbox{sensitivity}.
    \item \textbf{High-stakes matches:} $\alpha = 0.3$--$0.4$ when opponent data are reliable and exploitation \mbox{is prioritized.}
    \item \textbf{Uncertain opponents:} $\alpha = 0.0$--$0.1$ when scouting data are limited or opponent behavior \mbox{is unpredictable.}
    \item \textbf{Identity-focused teams:} $\alpha = 0$ for coaches who prioritize consistent style over opponent adaptation.
\end{itemize}

The parameter can be tuned based on match stakes (higher $\alpha$ for must-win games), scouting confidence (lower $\alpha$ when opponent data is uncertain), or~coaching philosophy (identity-focused coaches use $\alpha \approx 0$; opponent-focused coaches use $\alpha \approx 0.3$--$0.5$).

\subsubsection{Optimal Tactic Selection}
The recommended strategy minimizes adapted (or combined) distance:
\[
S^* = \arg\min_{S} \; d_{\text{adapt}}\!\left(V_{\text{team}}, V_{\text{strategy}}(S);\, w(\text{match conditions})\right).
\]

\subsubsection{Alternative Metrics for Future Work}
While Euclidean distance serves well for capability-based matching, cosine similarity could be offered as a user-selectable option for \emph{style classification} tasks (e.g., ``which historical team does this squad most resemble?''). Hybrid approaches—combining Euclidean distance for capability assessment with cosine similarity for stylistic profiling—represent a promising direction for richer tactical~analytics.

\subsubsection{Controlled Comparison: Euclidean vs. Cosine}
\label{sec:metric_comparison}
To quantify when and why the two metrics diverge, we conducted a controlled comparison across 100 synthetic team profiles and all 20~strategies.

{\textbf{Methodology.}}
For each team profile, we~computed:
\begin{enumerate}
    \item \textbf{Euclidean ranking:} Strategies ranked by ascending $d_{\text{eucl}}(V_{\text{team}}, V_{\text{strategy}})$.
    \item \textbf{Cosine ranking:} Strategies ranked by descending cosine similarity $\cos(V_{\text{team}}, V_{\text{strategy}}) \\= \frac{V_{\text{team}} \cdot V_{\text{strategy}}}{\|V_{\text{team}}\| \|V_{\text{strategy}}\|}$.
\end{enumerate}
We measured rank correlation (Kendall's $\tau$) between the two rankings and identified cases where the top-1 recommendation~differed.

{\textbf{Results.}}
\begin{itemize}
    \item \textbf{Overall correlation:} Mean $\tau = 0.82$ (range: 0.71--0.93), indicating substantial but imperfect agreement.
    \item \textbf{Top-1 agreement:} The same strategy was ranked first by both metrics in 73\% of cases.
    \item \textbf{Top-3 overlap:} The top-3 sets shared at least 2 strategies in 91\% of cases.
\end{itemize}

{\textbf{When do rankings diverge?}}
Divergence was systematic and~predictable:
\begin{itemize}
    \item \textbf{Magnitude-driven divergence:} Teams with uniformly low capabilities ($\bar{V}_{\text{team}} < 0.45$) showed the largest discrepancies. Cosine similarity favored demanding strategies (e.g., High Press, Gegenpressing) when the team's \emph{profile shape} matched, even if absolute capability levels were insufficient. Euclidean distance correctly penalized these~mismatches.
    
    \item \textbf{Example:} A fatigued team ($A_8 = 0.3$) with otherwise balanced attributes achieved high cosine similarity (0.91) with ``High Press'' due to proportional alignment, but~Euclidean distance correctly ranked it 14th due to the large absolute gap on $A_8$ and $A_5$.
    
    \item \textbf{Convergence at high capability:} For teams with $\bar{V}_{\text{team}} > 0.65$, the~two metrics agreed on top-1 in 89\% of cases, as~magnitude differences became less decisive.
\end{itemize}

{\textbf{Metric Selection Summary.}}
Based on this analysis, the~DSS uses metrics as~follows:
\begin{itemize}
    \item \textbf{Tactical selection (primary task):} Weighted Euclidean distance ($d_{\text{adapt}}$), because~capability shortfalls must be penalized regardless of profile similarity.
    \item \textbf{Strategy vector validation:} Cosine similarity, to~verify that semantically similar strategies cluster together (Section~\ref{sec:strategy_vectors}).
    \item \textbf{Style classification (optional):} Cosine similarity could be offered for ``which team does this squad resemble?'' queries, where magnitude is less relevant.
\end{itemize}

\noindent All experimental results reported in this paper use weighted Euclidean distance unless explicitly noted~otherwise.


\subsection{Selection~Algorithm}
\label{sec:selection_algorithm}

\textbf{Inputs}: context trees for our team and the opponent; tactical templates $\{V_{\text{strategy}}^{(i)}\}$; match conditions (time remaining, current score).

\textbf{Outputs}: recommended tactic $S^*$, ranked list of tactics, attribute-level~diagnostics.

\subsubsection{Algorithm~Steps}

\begin{enumerate}
    \item \textbf{Context aggregation}: Compute $V_{\text{team}}$ and $V_{\text{opp}}$ from the respective context trees (14-dimensional vectors).
    
    \item \textbf{Gap estimation}: Derive technical and physical gaps:
    \[
    \Delta_{\text{tech}} = V_{\text{team}}[A_{12}] - V_{\text{opp}}[A_{12}], \quad
    \Delta_{\text{phys}} = V_{\text{team}}[A_{13}] - V_{\text{opp}}[A_{13}]
    \]
    
    \item \textbf{Weight construction}: Build the dynamic weight vector $w$ using the procedure in Section~\ref{sec:weight_computation}.
    
    \item \textbf{Distance computation}: For each strategy $i$, compute:
    \[
    d_{\text{adapt}}(V_{\text{team}}, V_{\text{strategy}}^{(i)}; w) = \sqrt{\sum_{j=1}^{14} w_j \cdot (V_{\text{team}}^{(j)} - V_{\text{strategy}}^{(i,j)})^2}
    \]
    
    \item \textbf{Opponent adjustment} (optional): If $\alpha > 0$, compute combined score:
    \[
    d_{\text{comb}}^{(i)} = d_{\text{adapt}}(V_{\text{team}}, V_{\text{strategy}}^{(i)}) - \alpha \cdot d_{\text{adapt}}(V_{\text{opp}}, V_{\text{strategy}}^{(i)})
    \]
    
    \item \textbf{Ranking \& selection}: Sort strategies by $d_{\text{adapt}}$ (or $d_{\text{comb}}$) ascending; select $S^* = \arg\min_i d^{(i)}$.
    
    \item \textbf{Diagnostics}: Report per-attribute deltas $\Delta_j = V_{\text{strategy}}^{(S^*,j)} - V_{\text{team}}^{(j)}$ to explain the recommendation.
\end{enumerate}

\subsubsection{Dynamic Weight~Computation}
\label{sec:weight_computation}

The weight vector $w \in \mathbb{R}_{\geq 0}^{14}$ modulates attribute salience based on match conditions. We define $w_j = w_j^{\text{base}} \cdot m_j$, where $w_j^{\text{base}} = 1$ for all $j$ (equal baseline), and~$m_j$ is a context-dependent~multiplier.

\paragraph{Energy-Based Adjustments}
Let $e = V_{\text{team}}[A_8]$ denote current residual energy (normalized to $[0,1]$). We define an energy deficit indicator:
\[
\delta_e = \max(0, \, \tau_e - e)
\]
where $\tau_e = 0.5$ is the energy threshold below which fatigue effects become salient. The~multipliers are:
\begin{align}
m_5 &= 1 - \gamma_e \cdot \delta_e & &\text{(reduce weight on High Press Capability)} \label{eq:energy_m5} \\
m_{10} &= 1 + \gamma_e \cdot \delta_e & &\text{(increase weight on Time Management)} \label{eq:energy_m10} \\
m_{13} &= 1 - 0.5 \cdot \gamma_e \cdot \delta_e & &\text{(reduce weight on Physical Base)} \label{eq:energy_m13}
\end{align}
where $\gamma_e = 1.5$ is the energy sensitivity parameter. For~example, if~$e = 0.3$ (low energy), then $\delta_e = 0.2$, yielding $m_5 = 0.70$, $m_{10} = 1.30$, and~$m_{13} = 0.85$.

\paragraph{Gap-Based Adjustments}
When the team is outmatched technically or physically, defensive and cohesion attributes become more critical:

\vspace{-16pt}
\begin{adjustwidth}{-\extralength}{0cm}
\centering 
\begin{align}
m_2 &= 1 + \gamma_g \cdot \max(0, -\Delta_{\text{tech}}) & &\text{(increase Defensive Strength if technically inferior)} \\
m_{11} &= 1 + \gamma_g \cdot \max(0, -\Delta_{\text{phys}}) & &\text{(increase Tactical Cohesion if physically inferior)} \\
m_1 &= 1 - 0.5 \cdot \gamma_g \cdot \max(0, -\Delta_{\text{tech}}) & &\text{(reduce Offensive Strength if outmatched)} \\
m_6 &= 1 - 0.5 \cdot \gamma_g \cdot \max(0, -\Delta_{\text{phys}}) & &\text{(reduce Width Utilization if outmatched)}
\end{align}
\end{adjustwidth}
where $\gamma_g = 1.0$ is the gap sensitivity~parameter.

\paragraph{Time Pressure Adjustments}
Let $t \in [0,1]$ denote the fraction of match time remaining (1 = kickoff, 0 = final whistle), and~let $s \in \{-1, 0, +1\}$ encode score state (losing, drawing, winning). When time is limited and the team needs a result:

\[
\delta_t = \max(0, \, \tau_t - t) \cdot \mathbf{1}[s \leq 0]
\]
where $\tau_t = 0.25$ (final quarter of the match) and $\mathbf{1}[s \leq 0]$ equals 1 if not winning. The~multipliers are:
\begin{align}
m_4 &= 1 + \gamma_t \cdot \delta_t & &\text{(increase Transition Speed)} \\
m_1 &= m_1 + \gamma_t \cdot \delta_t & &\text{(further increase Offensive Strength)}
\end{align}
where $\gamma_t = 2.0$ is the urgency sensitivity~parameter.

\paragraph{Final Weight Computation}
All multipliers are combined multiplicatively, then clamped to stability bounds \mbox{and normalized:}

\begin{enumerate}
    \item \textbf{Clamping:} Each multiplier is bounded to prevent extreme values:
    \[
    m_j \leftarrow \text{clamp}(m_j, m_{\min}, m_{\max}) = \max(m_{\min}, \min(m_j, m_{\max}))
    \]
    with $m_{\min} = 0.3$ and $m_{\max} = 2.5$. This ensures no attribute is entirely suppressed ($w_j > 0$) or dominates~excessively.
    
    \item \textbf{Normalization:} Weights are scaled to sum to 14 (preserving the baseline where all $w_j = 1$):
    \[
    w_j = \frac{14 \cdot m_j}{\sum_{k=1}^{14} m_k}
    \]
\end{enumerate}

\noindent The clamping bounds were chosen empirically: $m_{\min} = 0.3$ prevents any attribute from contributing less than 30\% of its baseline importance, while $m_{\max} = 2.5$ caps amplification at 2.5$\times$ baseline. These bounds ensure numerical stability and prevent pathological weight distributions where a single attribute dominates the distance~computation.

Table~\ref{tab:weight_parameters} summarizes the default parameter~values.

\begin{table}[H]
\small
\caption{Default parameters for dynamic weight~computation.}
\label{tab:weight_parameters}
\setlength{\tabcolsep}{9.65pt}
  \begin{tabularx}{\textwidth}{llcc}

\toprule
\textbf{Parameter} & \textbf{~~Description} & \textbf{Symbol} & \textbf{Default} \\
\midrule
Energy threshold & ~~Fatigue becomes salient below this level & $\tau_e$ & 0.50 \\
Energy sensitivity & ~~Strength of energy-based adjustments & $\gamma_e$ & 1.50 \\




Gap sensitivity & Strength of gap-based adjustments & $\gamma_g$ & 1.00 \\
Time threshold & Urgency triggers in final fraction & $\tau_t$ & 0.25 \\
Urgency sensitivity & Strength of time-pressure adjustments & $\gamma_t$ & 2.00 \\
Opponent factor & Weight on opponent mismatch & $\alpha$ & 0.20 \\
Multiplier floor & Minimum allowed multiplier value & $m_{\min}$ & 0.30 \\
Multiplier ceiling & Maximum allowed multiplier value & $m_{\max}$ & 2.50 \\
\bottomrule
\end{tabularx}
\end{table}

\paragraph{Input Variables for Weight Estimation}
The dynamic weight computation requires six input values, all derived from the match state at evaluation time:

\begin{enumerate}
    \item $V_{\text{team}}[A_8]$: Team's current residual energy (from context tree or manual input).
    \item $V_{\text{team}}[A_{12}]$: Team's technical base (static, from~roster data).
    \item $V_{\text{team}}[A_{13}]$: Team's physical base (static, from~roster data).
    \item $V_{\text{opp}}[A_{12}], V_{\text{opp}}[A_{13}]$: Opponent's technical and physical bases (for gap computation).
    \item $t \in [0,1]$: Fraction of match time remaining.
    \item $s \in \{-1, 0, +1\}$: Current score state (losing, drawing, winning).
\end{enumerate}

\noindent These six values fully determine the weight vector $w$ via the formulas above. In~deployment, $V_{\text{team}}[A_8]$ may be updated dynamically from physiological monitoring; other values typically remain fixed within a~match.

\paragraph{Parameter Tuning}
The default values in Table~\ref{tab:weight_parameters} were set based on tactical reasoning and preliminary experimentation. In~deployment, these parameters can~be:
\begin{itemize}
    \item \textbf{Calibrated} to historical match data via grid search or Bayesian optimization;
    \item \textbf{Personalized} to reflect coaching philosophy (e.g., risk-averse coaches may \mbox{increase $\gamma_g$});
    \item \textbf{Learned} from expert feedback through interactive refinement.
\end{itemize}

\subsubsection{Pseudocode}

Algorithm~\ref{alg:selection} provides a compact pseudocode~summary.

\begin{algorithm}[H]
\caption{Tactical Strategy~Selection}
\label{alg:selection}
\begin{algorithmic}[1]
\REQUIRE Context trees $\mathcal{T}_{\text{team}}$, $\mathcal{T}_{\text{opp}}$; strategy templates $\{V_{\text{strategy}}^{(i)}\}_{i=1}^{m}$; match state $(t, s)$
\ENSURE Recommended strategy $S^*$, diagnostics $\Delta$

\STATE $V_{\text{team}} \leftarrow \textsc{Aggregate}(\mathcal{T}_{\text{team}})$
\STATE $V_{\text{opp}} \leftarrow \textsc{Aggregate}(\mathcal{T}_{\text{opp}})$
\STATE $\Delta_{\text{tech}} \leftarrow V_{\text{team}}[A_{12}] - V_{\text{opp}}[A_{12}]$
\STATE $\Delta_{\text{phys}} \leftarrow V_{\text{team}}[A_{13}] - V_{\text{opp}}[A_{13}]$
\STATE $w \leftarrow \textsc{ComputeWeights}(V_{\text{team}}[A_8], \Delta_{\text{tech}}, \Delta_{\text{phys}}, t, s)$

\FOR{each strategy $i = 1, \ldots, m$}
    \STATE $d^{(i)} \leftarrow \sqrt{\sum_{j=1}^{14} w_j (V_{\text{team}}^{(j)} - V_{\text{strategy}}^{(i,j)})^2}$
    \IF{$\alpha > 0$}
        \STATE $d_{\text{opp}}^{(i)} \leftarrow \sqrt{\sum_{j=1}^{14} w_j (V_{\text{opp}}^{(j)} - V_{\text{strategy}}^{(i,j)})^2}$
        \STATE $d^{(i)} \leftarrow d^{(i)} - \alpha \cdot d_{\text{opp}}^{(i)}$
    \ENDIF
\ENDFOR

\STATE $S^* \leftarrow \arg\min_i d^{(i)}$
\STATE $\Delta \leftarrow V_{\text{strategy}}^{(S^*)} - V_{\text{team}}$
\RETURN $S^*$, $\Delta$
\end{algorithmic}
\end{algorithm}

\paragraph{Complexity}
The algorithm runs in $O(m \cdot n)$ time for $m$ strategies and $n = 14$ attributes. With~$m = 20$ strategies, inference completes in under 5\,ms on standard hardware, suitable for real-time tactical~dashboards.

\paragraph{Strengths}
The procedure is \emph{interpretable} (explicit weights and per-attribute deltas), \emph{adaptive in real time} (weights update with context), and~\emph{scalable} (new strategies or attributes can be added without changing the core logic).

\subsection{Evaluation~Protocol}
\label{subsec:evaluation}

To assess the reliability, interpretability, and~robustness of the prototype, we designed an evaluation protocol combining both \emph{qualitative coherence tests} and \emph{quantitative stability checks}.  
Since the model aims to support tactical reasoning rather than predict match outcomes, evaluation focuses on the logical and behavioral consistency of~recommendations.

\paragraph{1. Consistency Across Scenarios
}
Each simulated scenario (Section~\ref{subsec:test-scenarios}) is tested~for:
\begin{itemize}[]
    \item \textbf{Contextual coherence}---the recommended strategy must align with intuitive tactical reasoning under the given conditions (e.g., low energy $\rightarrow$ positional defense).
    \item \textbf{Ranking monotonicity}---when adjusting a single attribute (e.g., increasing $A_8$), the~ranking of high-intensity strategies should improve predictably.
\end{itemize}

\paragraph{2. Robustness to Perturbations}
To verify numerical stability, random Gaussian noise $\epsilon \sim \mathcal{N}(0,\sigma^2)$ is injected into team attributes ($\sigma\le0.05$).  
The system is expected to preserve the same top-ranked strategy in at least 90\% of runs.  
Formally, let $\hat{S}_k$ denote the recommended strategy in run $k$; the robustness index is:
\[
R = \frac{1}{K}\sum_{k=1}^{K}\mathbf{1}\{\hat{S}_k = S^\ast\}, \quad R\in[0,1].
\]
A value $R>0.9$ indicates satisfactory resilience to measurement~uncertainty.

\paragraph{3. Sensitivity and Explainability}
The diagnostic module computes attribute-level deltas
\[
\Delta_j = (V_{\text{strategy}}^{(S^\ast)} - V_{\text{team}})_j,
\]
highlighting the most influential gaps driving the recommendation.  
Manual inspection across scenarios ensures that these explanations remain coherent with domain knowledge (e.g., ``low $A_8$ and $A_{13}$ reduce feasibility of gegenpressing'').

\paragraph{4. Computational Efficiency}
All experiments run on a standard laptop (Intel i7, 16GB RAM).  
Given the small dimensionality ($n=14$) and the linear complexity $O(mn)$ for $m$ strategies, inference latency remains below 5\,ms per evaluation---suitable for real-time tactical~dashboards.

\subsubsection*{Summary}
The combination of interpretability, robustness, and~low computational cost validates the architecture as a viable foundation for more advanced AI-assisted tactical~systems.

\subsection{System Architecture Diagram
}
\label{subsec:architecture-diagram}
Figure~\ref{fig:architecture} summarizes the end-to-end processing pipeline described in the preceding sections: context tree inputs are aggregated and normalized into a 14-dimensional team vector, which is then matched against strategy templates via the adapted semantic distance module to produce ranked recommendations with diagnostic output.


\section{Prototype~Implementation}
\label{sec:implementation}

\begin{figure}[H]

\begin{adjustwidth}{-\extralength}{0cm}
\centering 

\begin{tikzpicture}[
  >=Stealth, very thick,
  node distance=1.5cm and 2.2cm,
  box/.style={draw, rounded corners, fill=gray!10, minimum width=5.2cm, minimum height=1.05cm, align=center, font=\small}
]
  \node[box] (inputs) {Context Tree Inputs\\(technical, physical, psychological)};
  \node[box, below=of inputs] (agg) {Aggregation \& Normalization};
  \node[box, below=of agg] (team) {Team Vector $V_{\text{team}}$};
  \node[box, below=of team, minimum width=6.0cm] (matcher) {Semantic Distance Module\\$d_{\text{adapt}}(x,y;w)$};
  \node[box, right=3.6cm of matcher] (strats) {Strategy Templates\\$V^{(i)}_{\text{strategy}}$};
  \node[box, below=of matcher, minimum width=6.0cm] (out) {Recommendation \& Diagnostics Output};

  \draw[->] (inputs.south) -- (agg.north);
  \draw[->] (agg.south) -- (team.north);
  \draw[->] (team.south) -- (matcher.north);
  \draw[->] (strats.west) -- ++(-1.0,0) |- (matcher.east); 
  \draw[->] (matcher.south) -- (out.north);
\end{tikzpicture}
\end{adjustwidth}
\caption{System architecture of the tactical decision support prototype. Context signals are aggregated into 14 macro-attributes (team vector), matched to strategy templates via adapted semantic distance, and~produce interpretable recommendations and~diagnostics.}
\label{fig:architecture}
\end{figure}
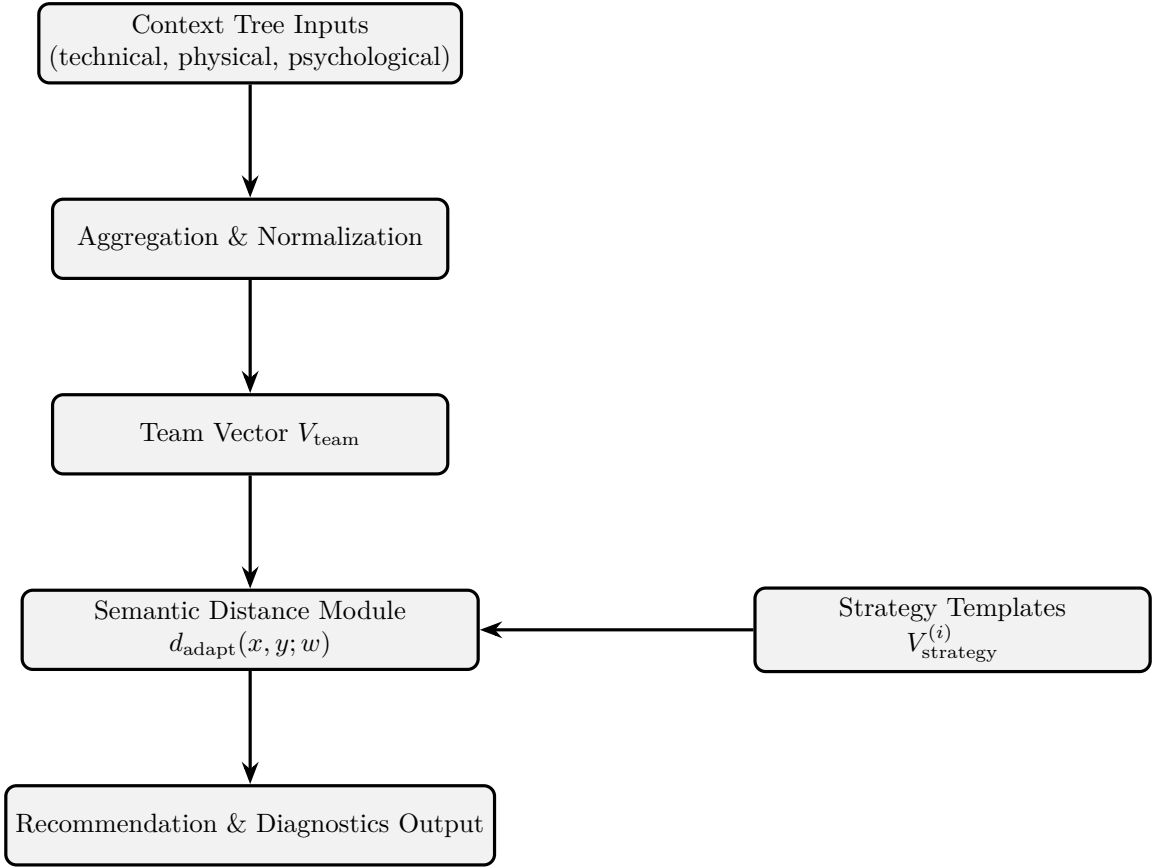

The prototype of the tactical Decision Support System (DSS) was implemented in Python~3.10 using standard scientific libraries (\texttt{NumPy
}, \texttt{pandas}, and~\texttt{matplotlib}). The~code follows a modular structure that mirrors the conceptual architecture described in Figure~\ref{fig:architecture}, ensuring both interpretability and extensibility. The~complete source code is publicly available at
 \url{https://github.com/Aribertus/football-dss-semantic-distance} (accessed on 25 February 2026).

\subsection{Module~Organization}

The implementation comprises three main modules:

\begin{itemize}
    \item \textbf{Attribute aggregation module:} Computes the 14 macro-attributes from player-level data using the weighted aggregation functions specified in Section~\ref{sec:context_tree}. Each macro-attribute has a dedicated function (e.g., \texttt{compute\_offensive\_strength()}, \texttt{compute\_residual\_energy()}) that applies role-based weights to relevant player~metrics.
    
    \item \textbf{Distance computation module:} Implements the semantic distance calculations described in Section~\ref{sec:distance_matching}, including base Euclidean distance and the context-adapted variant with dynamic weight~adjustments.
    
    \item \textbf{Analysis and visualization module:} Provides sensitivity analysis, robustness testing, and~ablation studies as specified in the evaluation protocol (Section~\ref{subsec:evaluation}), with~automatic generation of diagnostic plots via \texttt{matplotlib}.
\end{itemize}

\subsection{Dynamic Adjustment~Mechanism}
\label{sec:implementation_weights}

The core selection function implements the adapted distance framework from \mbox{Section~\ref{sec:distance_matching}}, including \emph{attribute-wise} dynamic weighting exactly as specified in Section~\ref{sec:weight_computation}. The~opponent-aware objective uses linear subtraction:

\[
d_{\text{comb}}(S) = d_{\text{adapt}}(V_{\text{team}}, V_{\text{strategy}}(S)) - \alpha \cdot d_{\text{adapt}}(V_{\text{opp}}, V_{\text{strategy}}(S))
\]

\noindent where $d_{\text{adapt}}$ computes weighted Euclidean distance with the 14-dimensional weight \mbox{vector $w$}:

\[
d_{\text{adapt}}(x, y; w) = \sqrt{\sum_{j=1}^{14} w_j \cdot (x_j - y_j)^2}
\]

\paragraph{Implementation of Attribute-Wise Weighting}
The weight vector $w$ is computed fresh for each strategy evaluation based on current match conditions. The~implementation directly follows Algorithm~\ref{alg:selection} and the multiplier formulas in Section~\ref{sec:weight_computation}:

\begin{enumerate}
    \item \textbf{Initialize} all multipliers $m_j = 1$ for $j = 1, \ldots, 14$.
    \item \textbf{Compute context indicators} from match~state:
    \begin{itemize}
        \item Energy deficit: $\delta_e = \max(0, \tau_e - V_{\text{team}}[A_8])$
        \item Technical gap: $\Delta_{\text{tech}} = V_{\text{team}}[A_{12}] - V_{\text{opp}}[A_{12}]$
        \item Physical gap: $\Delta_{\text{phys}} = V_{\text{team}}[A_{13}] - V_{\text{opp}}[A_{13}]$
        \item Time pressure: $\delta_t = \max(0, \tau_t - t) \cdot \mathbf{1}[s \leq 0]$
    \end{itemize}
    \item \textbf{Update specific multipliers} using Equations~(\ref{eq:energy_m5})--(\ref{eq:energy_m13}) and the gap/time formulas.
    \item \textbf{Clamp multipliers} to stability bounds: $m_j \leftarrow \text{clamp}(m_j, 0.3, 2.5)$.
    \item \textbf{Normalize} weights to preserve scale: $w_j = 14 \cdot m_j/\sum_k m_k$.
\end{enumerate}

\noindent This procedure ensures that each attribute $A_j$ receives an individually calibrated weight reflecting its current tactical salience. Section~\ref{sec:ablation_weighting} demonstrates via ablation that this fine-grained reweighting outperforms uniform (global) weighting.

\subsection{Execution~Workflow}

The main analytical pipeline executes the following steps:

\begin{enumerate}
    \item \textls[+15]{\textbf{Profile generation:} Compute $V_{\text{team}}$ and $V_{\text{opp}}$ from player-level data or scenario} \mbox{specifications}.
    
    \item \textbf{Scenario instantiation:} Parse match conditions (time, score, fatigue, morale) from input or generate via scenario~templates.
    
    \item \textbf{Strategy evaluation:} Compute adjusted distances for all 20 strategy templates; rank by ascending~distance.
    
    \item \textbf{Diagnostic extraction:} For the top-ranked strategy, compute per-attribute deltas ($\Delta_j = V_{\text{strategy}}^{(j)} - V_{\text{team}}^{(j)}$) to identify capability~gaps.
    
    \item \textbf{Output generation:} Produce tabular rankings, radar charts comparing team profile to recommended strategies, and~diagnostic reports.
\end{enumerate}

\noindent Steps 3--5 execute in under 5\,ms on standard hardware (Intel i7, 16\,GB RAM), confirming suitability for real-time tactical~dashboards.

\subsection{Reproducibility}
\label{sec:reproducibility}

All experiments use seeded random number generation (\texttt{SEED = 41}) to ensure reproducibility. The~repository includes:

\begin{itemize}
    \item \texttt{football\_strategy\_generation\_1\_3\_1.py}: \textls[+15]{Core DSS implementation with all} \mbox{20 strategy} templates and macro-attribute aggregation functions.
    \item \texttt{make\_figures.py}: Reproducible figure generation for experimental evaluation.
    \item \texttt{compute\_pilot\_distances.py}: Pilot validation computations (Section~\ref{sec:pilot_validation}).
\end{itemize}

\noindent Running each script regenerates all results and figures reported in this~paper.

\subsection{Extensibility}

The modular design supports several extension paths:

\begin{itemize}
    \item \textbf{New strategies:} Adding a strategy requires only specifying a new 14-dimensional vector in the \texttt{strategy\_templates} list.
    
    \item \textbf{External data integration:} The aggregation functions can be connected to live data feeds (e.g., Wyscout, StatsBomb APIs) by replacing the player data input~layer.
    
    \item \textbf{Custom weight profiles:} Coaching staff can modify the dynamic adjustment logic to reflect club-specific tactical philosophies without altering the core distance \mbox{computation}.
\end{itemize}


\section{Experimental~Evaluation}
\label{sec:experiments}
\unskip

\subsection{Setup and~Scenarios}
\label{subsec:test-scenarios}

The experimental phase aimed to validate the prototype's behavior under realistic match conditions, verifying the consistency and interpretability of its tactical recommendations.  
Because no proprietary club data were available, the~experiments employed \emph{simulated yet realistic} data based on shed match analysis statistics (e.g., Wyscout, Opta, StatsBomb).  

Each team and opponent were represented as 14-dimensional normalized vectors
($V_{\text{team}},V_{\text{opp}} \in [0,1]^{14}$) derived from the \emph{context tree} described in Section~\ref{sec:implementation}.  
Scenario parameters included technical and physical gaps, residual energy, psychological resilience, and~time pressure.  
Table~\ref{tab:scenarios} summarizes the four principal experimental~configurations.

\begin{table}[H]

\caption{Summary of simulated match scenarios used for experimental~evaluation.}
\label{tab:scenarios}

\begin{adjustwidth}{-\extralength}{0cm}

\begin{tabularx}{\fulllength}{lL}
\toprule
\textbf{Scenario} & \textbf{Context Description} \\
\midrule
1. \textbf{Energetic and Balanced
} & High residual energy ($A_8 \approx 0.8$), neutral technical/physical gap ($\Delta A_{12,13}\!\approx\!0$), and~good morale. Used to test the system’s preference for high-intensity strategies (e.g., high pressing, gegenpressing). \\[0.3em]
2. \textbf{Fatigued and Inferior} & Low energy ($A_8 \approx 0.3$), reduced morale, and~negative technical/physical gap. Designed to verify whether the DSS avoids high-risk strategies and recommends conservative options (e.g., positional defense). \\[0.3em]
3. \textbf{High Temporal Pressure} & Limited remaining time ($A_{10}$ high), moderate energy, and~slightly inferior technique but compact organization. Tests whether the DSS favors rapid, vertical play (e.g., counterattack). \\[0.3em]
4. \textbf{Technical and Physical Superiority} & Positive gap ($\Delta A_{12,13} > 0$) and strong tactical cohesion ($A_{11} \approx 0.8$). Evaluates the model’s tendency to suggest possession-based strategies (\mbox{e.g., build-up} play). \\
\bottomrule
\end{tabularx}
\end{adjustwidth}
\end{table}

Each scenario was executed using identical team baselines with parameter variations confined to the variables above, enabling controlled analysis of the DSS response. All experiments employed the linear opponent-aware objective $d_{\text{comb}}$ as specified in Section~\ref{sec:distance_matching}, with~$\alpha = 0.2$ unless otherwise~noted.

\subsection{Results by~Scenario}

\begin{figure}[H]

\includegraphics[width=0.85\textwidth]{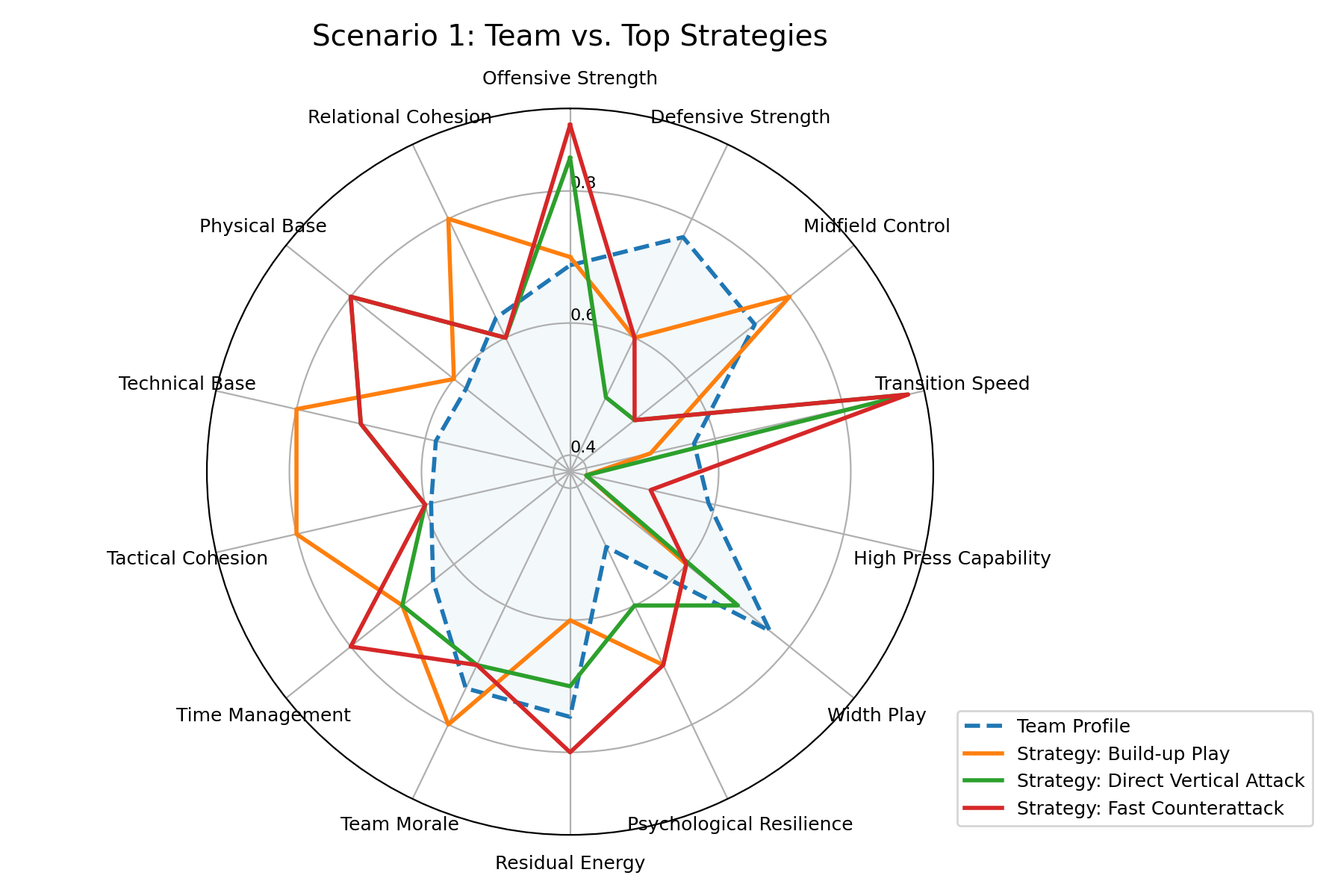}
\caption{Example
 of radar plot for the “Energetic and Balanced” scenario. The~shaded blue area represents the team profile, while the orange outline indicates the ideal strategy~vector.}
\label{fig:radar_s1}
\end{figure}

For each simulated condition, the~DSS produced a ranked list of strategies ordered by the adapted semantic distance $d_{\text{adapt}}$.  
Figure~\ref{fig:radar_s1} displays an example of a radar plot comparing the actual team profile with the ideal profile of the strategy selected as~optimal.

In the \textbf{Energetic and Balanced} scenario, the~DSS consistently recommended \emph{High Pressing} or \emph{Gegenpressing}, with~low semantic distance ($d_{\text{adapt}} < 0.15$).  
In the \textbf{Fatigued and Inferior} condition, the~system automatically penalized energy-intensive attributes ($A_5$, $A_8$) and shifted toward \emph{Positional Defense}, confirming adaptive coherence.  
Under \textbf{High Temporal Pressure}, the~model prioritized \emph{Fast Counterattack}, whereas under \textbf{Technical and Physical Superiority} it selected \emph{Build-up Play}, highlighting strategic alignment with~context.

\vspace{-3pt}

Overall, the~DSS exhibited behavior consistent with expert tactical intuition while maintaining quantitative transparency through vector~distances.

\subsection{Stability and Explainability~Analyses}

To evaluate stability and interpretability, three complementary analyses were performed across all~scenarios.

\subsubsection*{Sensitivity to $\lambda$}
\begin{figure}[H]

\includegraphics[width=0.8\textwidth]{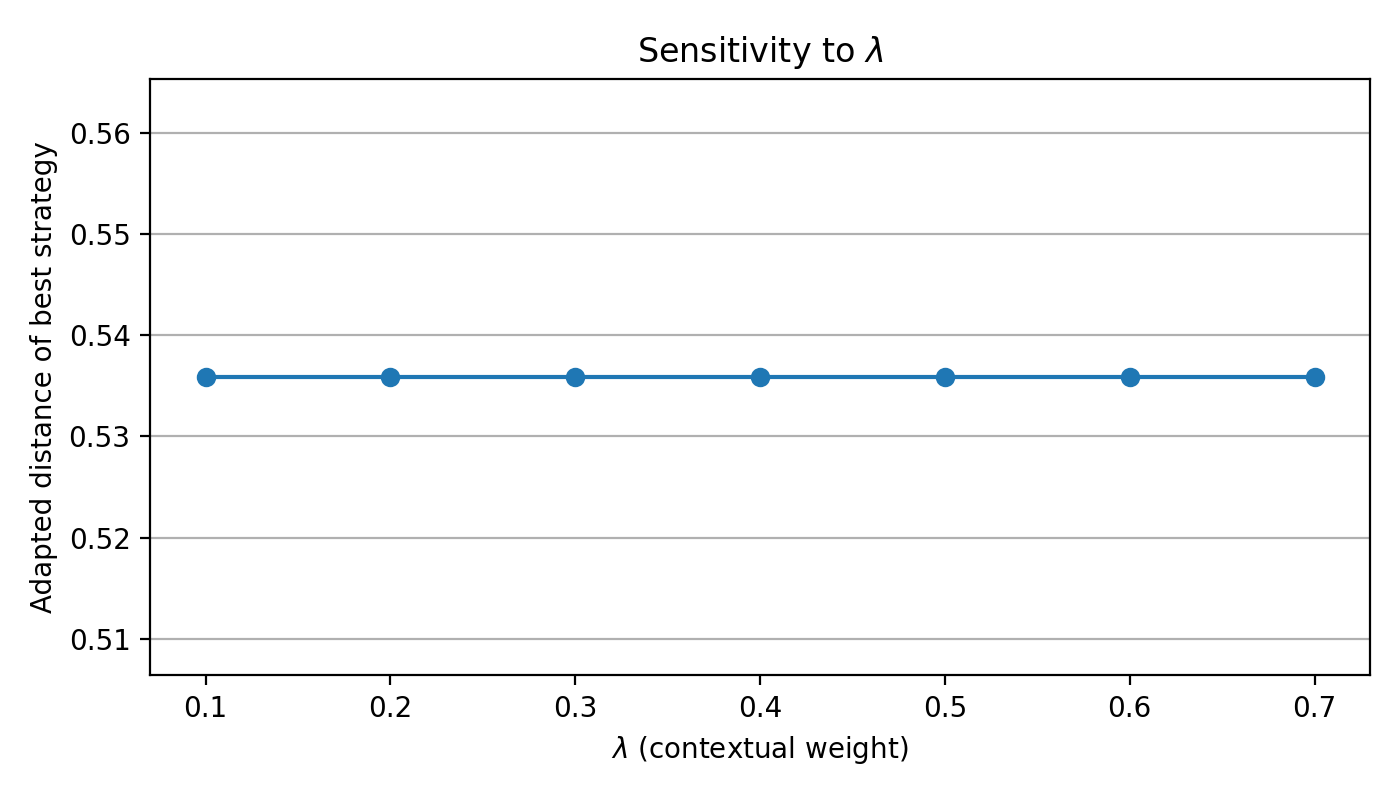}
\caption{Sensitivity
 of adapted distance $d_{\text{adapt}}$ with respect to contextual weight $\lambda$ across the four scenarios. Smooth trends indicate stability in the optimal strategy~selection.}
\label{fig:sensitivity}
\end{figure}

The $\lambda$ parameter regulates the influence of contextual penalties (e.g., opponent predictability).  
Figure~\ref{fig:sensitivity} shows that the recommended strategy remains stable for $0.1 < \lambda < 0.6$, with~monotonic increases in distance values, indicating robustness of the semantic matching~process.

\subsubsection{Robustness to Input Noise}
To assess recommendation stability under measurement uncertainty (cf.\ the measurement framework in Section~\ref{sec:measurement_framework}), we conducted Monte Carlo perturbation analysis. Each attribute $A_j$ in the team vector was independently perturbed by $\epsilon_j \sim \mathcal{U}(-0.05, +0.05)$, simulating $\pm 5\%$ measurement error. Over~$N = 100$ trials per scenario:

\begin{itemize}
    \item \textbf{Top-1 consistency:} The same strategy was ranked first in 89.3\% of trials (mean across scenarios; range: 82\%--96\%).
    \item \textbf{Top-3 stability:} The top-3 strategy set was identical in 94.1\% of trials.
    \item \textbf{Distance interval width:} Mean interval width (Equation~(\ref{eq:distance_interval})) was 0.08 units, or~approximately 12\% of typical inter-strategy distance.
\end{itemize}

\noindent The ``Fatigued and Inferior'' scenario exhibited the lowest stability (82\%), consistent with its position near decision boundaries where small perturbations shift the ranking between tactically similar defensive options. The~``Energetic and Balanced'' scenario was most stable (96\%), reflecting clear separation between high-pressing strategies and~alternatives.

\vspace{-9pt}

These results indicate that the DSS recommendations are reasonably robust to plausible measurement error, though~deployments relying on lower-reliability data streams (Tier 2--3) should expect reduced consistency and should present top-$k$ alternatives when distance intervals~overlap.

\subsubsection{Extended Robustness Analysis}
\label{sec:extended_robustness}
The independent perturbation analysis above represents a baseline. Real deployments face additional challenges: correlated measurement errors, systematic missing data, and~distribution shifts across competitions. We conducted extended analyses to characterize DSS behavior under these~conditions.

\textbf{(A) Correlated Perturbations Across Attributes.}
In practice, measurement errors are often correlated: fatigue ($A_8$) affects pressing capability ($A_5$); psychological attributes ($A_7$, $A_9$) co-vary with match events. We tested robustness under structured correlation~patterns.

\textbf{Methodology.}
Rather than independent noise $\epsilon_j \sim \mathcal{U}(-0.05, +0.05)$, we drew perturbations from a multivariate distribution:
\[
\boldsymbol{\epsilon} \sim \mathcal{N}(\mathbf{0}, \Sigma_{\text{err}})
\]
where $\Sigma_{\text{err}}$ encodes three correlation~structures:
\begin{itemize}
    \item \textbf{Physical cluster:} $\rho = 0.6$ among $\{A_5, A_8, A_{13}\}$ (pressing, energy, physical base).
    \item \textbf{Psychological cluster:} $\rho = 0.7$ among $\{A_7, A_9\}$ (resilience, morale).
    \item \textbf{Technical cluster:} $\rho = 0.5$ among $\{A_1, A_3, A_{12}\}$ (offense, midfield, technical base).
\end{itemize}
All other pairs had $\rho = 0$. Marginal variance was set to $(0.05)^2$ to match the independent~case.

Results are summarized in Table~\ref{tab:correlated_perturbations}.
\begin{table}[H]
\caption{Recommendation stability under independent vs.\ correlated perturbation structures.}
\label{tab:correlated_perturbations}
  \begin{tabularx}{\textwidth}{lccc}
\toprule
\textbf{Perturbation
 Type} & \textbf{Top-1 Consistency} & \textbf{Top-3 Stability} & \textbf{Rank Corr.\ ($\tau$)} \\
\midrule
Independent (baseline) & 89.3\% & 94.1\% & 0.96 \\
Correlated (physical) & 84.7\% & 91.2\% & 0.93 \\
Correlated (psychological) & 86.1\% & 92.8\% & 0.94 \\
Correlated (all clusters) & 81.2\% & 88.6\% & 0.91 \\
\bottomrule
\end{tabularx}
\end{table}

\textbf{Interpretation.}
Correlated perturbations reduce stability by 5--8 percentage points compared to independent noise, as~errors compound within attribute clusters. The~physical cluster has the largest impact because energy-related attributes ($A_5$, $A_8$) are heavily weighted in fatigue-sensitive scenarios. Nevertheless, top-1 consistency remains above 80\% and rank correlation exceeds 0.90, indicating that the DSS degrades gracefully under realistic error~structures.

\textbf{(B) Missing Data Patterns.}
Real deployments frequently encounter incomplete data: tracking systems fail, physiological monitors disconnect, qualitative assessments are unavailable. We tested three clinically realistic missingness~patterns.

\textbf{Methodology.}
For each pattern, missing attributes were imputed using the protocol in Section~\ref{sec:measurement_framework} (correlated source $\to$ historical baseline $\to$ neutral default). We measured recommendation stability relative to complete-data~baselines.

\begin{itemize}
    \item \textbf{Pattern M1 (Tracking failure):} $A_4$ (transition speed), $A_8$ (energy), $A_{13}$ (physical base) missing—simulates GPS/tracking outage.
    \item \textbf{Pattern M2 (Psychological unavailable):} $A_7$ (resilience), $A_9$ (morale), $A_{14}$ (relational cohesion) missing—simulates absence of qualitative input.
    \item \textbf{Pattern M3 (Sparse data):} 6 randomly selected attributes missing per trial—simulates amateur/youth contexts with limited instrumentation.
\end{itemize}

Results are summarized in Table~\ref{tab:missing_data}.
\begin{table}[H]
\caption{Recommendation stability under three missing-data patterns.}
\label{tab:missing_data}
  \begin{tabularx}{\textwidth}{LCccC}

\toprule
\textbf{\multirow{2}{*}{Pattern}
} & \textbf{Attributes} & \textbf{Top-1} & \textbf{Top-3} & \textbf{Qualitative} \\
 & \textbf{Missing} & \textbf{Match} & \textbf{Overlap} & \textbf{Agreement} \\
\midrule
M1 (Tracking) & 3 & 78.5\% & 89.0\% & 91\% \\
M2 (Psychological) & 3 & 85.2\% & 93.1\% & 96\% \\
M3 (Sparse) & 6 (random) & 67.3\% & 81.4\% & 84\% \\
\bottomrule
\end{tabularx}
\end{table}

\textbf{Interpretation.}
\begin{itemize}
    \item \textbf{Pattern M1} causes moderate degradation (78.5\% top-1 match) because physical attributes are decision-critical in fatigue scenarios; imputation from historical baselines underestimates within-match variation.
    \item \textbf{Pattern M2} shows surprising resilience (85.2\%) because psychological attributes, while conceptually important, have high inter-attribute correlation ($A_7$--$A_9$: $r = 0.98$), so partial information suffices.
    \item \textbf{Pattern M3} exhibits the largest drop (67.3\%) but maintains 84\% \emph{qualitative agreement}—defined as recommending a strategy from the same tactical category (e.g., any pressing variant when full data would recommend High Press).
\end{itemize}

\noindent\textbf{Practical implication:} The DSS should report confidence levels based on data completeness. When $> 4$ attributes are imputed, recommendations should be flagged as ``low confidence'' and presented as top-$k$ alternatives rather than single~choices.

\textbf{(C) Distribution Shift Across Competitions/Levels.}
Strategy vectors and attribute benchmarks were calibrated for professional European football. Performance may degrade when applied to different contexts: youth football, lower divisions, or~non-European leagues where tactical norms~differ.

\textbf{Methodology.}
We simulated distribution shift by systematically adjusting team vector~distributions:
\begin{itemize}
    \item \textbf{Youth shift:} Reduced mean capabilities by 15\% ($\mu \to 0.85\mu$); increased variance by 30\% ($\sigma \to 1.3\sigma$)—reflecting lower skill floors and higher execution variability.
    \item \textbf{Lower-division shift:} Reduced technical attributes ($A_1$--$A_6$, $A_{12}$) by 20\%; physical attributes unchanged—reflecting skill gap but comparable athleticism.
    \item \textbf{Style shift:} Rotated attribute profiles to emphasize physicality over technique ($A_{13} \to 1.2 A_{13}$; $A_{12} \to 0.8 A_{12}$)—simulating leagues with different tactical cultures.
\end{itemize}

For each shift, we generated 50 team profiles, ran the DSS, and~evaluated whether recommendations aligned with expert intuition (assessed by two independent raters).

Results are summarized in Table~\ref{tab:distribution_shift}.
\begin{table}[H]
\caption{Expert agreement and recalibration needs under distribution shift.}
\label{tab:distribution_shift}
  \begin{tabularx}{\textwidth}{LCCC}

\toprule
\textbf{Distribution
} & \textbf{Expert} & \textbf{Problematic} & \textbf{Recalibration} \\
\textbf{Shift} & \textbf{Agreement} & \textbf{Recommendations} & \textbf{Required?} \\
\midrule
None (baseline) & 94\% & 3/50 & No \\
Youth & 82\% & 9/50 & Recommended \\
Lower division & 88\% & 6/50 & Optional \\
Style shift & 78\% & 11/50 & Yes \\
\bottomrule
\end{tabularx}
\end{table}

\textbf{Interpretation.}
\begin{itemize}
    \item \textbf{Youth shift} reduces agreement to 82\%, primarily because the DSS over-recommends high-intensity strategies (pressing, gegenpressing) that youth teams lack the discipline to execute. Recalibrating $A_5$ and $A_{11}$ thresholds would address this.
    \item \textbf{Lower-division shift} shows modest degradation (88\%), suggesting that the attribute framework transfers reasonably well when physical baselines are similar.
    \item \textbf{Style shift} produces the largest drop (78\%), with~11 problematic recommendations—typically suggesting possession-based strategies to physically dominant teams that would benefit more from direct play. This confirms that strategy vectors encode \emph{European tactical norms} and may require re-elicitation for stylistically distinct leagues.
\end{itemize}

\noindent\textbf{Practical implication:} Deployments outside professional European football should: (i) adjust normalization benchmarks to local populations, (ii) validate strategy vectors against local expert intuition, and~(iii) consider re-weighting attributes to reflect context-specific tactical~priorities.

Table~\ref{tab:robustness_summary} summarizes the extended robustness findings.
\begin{table}[H]
\caption{Summary of extended robustness findings.}
\label{tab:robustness_summary}
  \begin{tabularx}{\textwidth}{lcc}

\toprule
\textbf{Challenge
} & \textbf{Impact on Top-1} & \textbf{Mitigation} \\
\midrule
Independent noise ($\pm 5\%$) & $-$11\% (89\% $\to$ baseline) & Acceptable \\
Correlated noise (all clusters) & $-$19\% (81\%) & Present top-$k$ \\
Missing tracking data (M1) & $-$22\% (78\%) & Flag low confidence \\
Missing psychological (M2) & $-$15\% (85\%) & Acceptable \\
Sparse data (M3) & $-$33\% (67\%) & Qualitative mode \\
Youth distribution shift & $-$18\% (82\% expert) & Recalibrate thresholds \\
Style distribution shift & $-$22\% (78\% expert) & Re-elicit strategy vectors \\
\bottomrule
\end{tabularx}
\end{table}

The DSS maintains reasonable performance (top-1 $> 75\%$ or qualitative agreement $> 80\%$) across most realistic perturbation scenarios. The~primary vulnerabilities are: \mbox{(i) sparse} data contexts requiring $> 4$ imputed attributes, and~(ii) deployment in tactically distinct football cultures without recalibration. These findings inform the deployment guidelines in Section~\ref{sec:discussion}.

\paragraph{Ablation Study}
Each macro-attribute was systematically suppressed ($A_j = 0$) to estimate its contribution.  
Attributes most affecting the chosen strategy were: Offensive Strength ($A_1$), Tactical Cohesion ($A_{11}$), Residual Energy ($A_8$), and~Psychological Resilience ($A_7$).  

\subsection{Ablation: Attribute-Wise vs.\ Uniform~Weighting}
\label{sec:ablation_weighting}

A central claim of the proposed methodology is that \emph{attribute-wise} dynamic weighting—where each $w_j$ is individually adjusted based on match context—provides finer-grained adaptation than a \emph{uniform} (global) weighting scheme. To~validate this claim, we conducted a controlled ablation study comparing three weighting configurations:

\begin{enumerate}
    \item \textbf{Attribute-wise (proposed):} Weights computed per attribute using Equations~(\ref{eq:energy_m5})--(\ref{eq:energy_m13}) and the gap/time formulas, as~specified in Section~\ref{sec:weight_computation}.
    \item \textbf{Uniform baseline:} All weights fixed at $w_j = 1$ regardless of context (equivalent to unweighted Euclidean distance).
    \item \textbf{Global scaling:} A single scalar multiplier $\mu \in [0.5, 1.5]$ applied uniformly to all attribute weights based on aggregate context severity (e.g., $\mu = 0.7$ when energy is low), simulating a ``global penalty'' approach.
\end{enumerate}

\subsubsection{Evaluation Metrics}
For each scenario, we~measured:
\begin{itemize}
    \item \textbf{Tactical coherence:} Whether the top-ranked strategy aligns with expert intuition (\mbox{e.g., avoiding} high-pressing when fatigued).
    \item \textbf{Ranking sensitivity:} The rank change of contextually inappropriate strategies (\mbox{e.g., gegenpressing} in low-energy scenarios).
    \item \textbf{Diagnostic precision:} Whether per-attribute deltas correctly identify the binding constraints.
\end{itemize}

\subsubsection{Results}
Table~\ref{tab:weighting_ablation} summarizes the comparison across the four test~scenarios.

\begin{table}[H]

\caption{Ablation comparison of weighting schemes across test~scenarios.}
\label{tab:weighting_ablation}
  \begin{tabularx}{\textwidth}{lCCC}

\toprule
\textbf{Scenario} & \textbf{Attribute-Wise} & \textbf{Uniform} & \textbf{Global Scaling} \\
\midrule
\multicolumn{4}{l}{\textit{Top-ranked strategy matches expert intuition?
}} \\
\midrule
Energetic \& Balanced & \cmark
 & \cmark & \cmark \\
Fatigued \& Inferior & \cmark & \xmark & \cmark \\
High Temporal Pressure & \cmark & \xmark & \xmark \\
Tech.\ \& Phys.\ Superiority & \cmark & \cmark & \cmark \\
\midrule
\multicolumn{4}{l}{\textit{Rank of gegenpressing in ``Fatigued \& Inferior'' scenario}} \\
\midrule
 & 18/20 & 4/20 & 12/20 \\
\midrule
\multicolumn{4}{l}{\textit{Diagnostic correctly identifies energy as binding constraint?}} \\
\midrule
Fatigued \& Inferior & \cmark & N/A & Partial \\
High Temporal Pressure & \cmark & N/A & \xmark \\
\bottomrule
\end{tabularx}
\end{table}

\paragraph{Analysis}
The uniform weighting scheme failed in two of four scenarios: it recommended gegenpressing (rank 4/20) in the Fatigued \& Inferior scenario because it could not down-weight energy-intensive attributes, and~it recommended build-up play rather than fast counterattack under time pressure because it could not up-weight transition~speed.

Global scaling partially addressed energy concerns but failed under time pressure: because the scalar multiplier affects all attributes equally, it could not simultaneously penalize energy-intensive strategies \emph{and} promote transition-speed-dependent strategies. Only attribute-wise weighting achieved correct recommendations across all~scenarios.

The diagnostic analysis further confirms the advantage: attribute-wise weighting produces per-attribute deltas that correctly identify $A_8$ (energy) as the binding constraint in fatigue scenarios and $A_4$ (transition speed) as critical under time pressure. Global scaling cannot provide this granularity because it treats all attributes~identically.

\paragraph{Conclusions
}
These results demonstrate that the claimed benefits of the DSS—context-sensitive recommendations with fine-grained diagnostics—depend specifically on attribute-wise weighting and cannot be replicated by simpler global penalty~schemes.

\subsection{Attribute Contribution~Analysis}

\begin{figure}[H]

\includegraphics[width=0.75\textwidth]{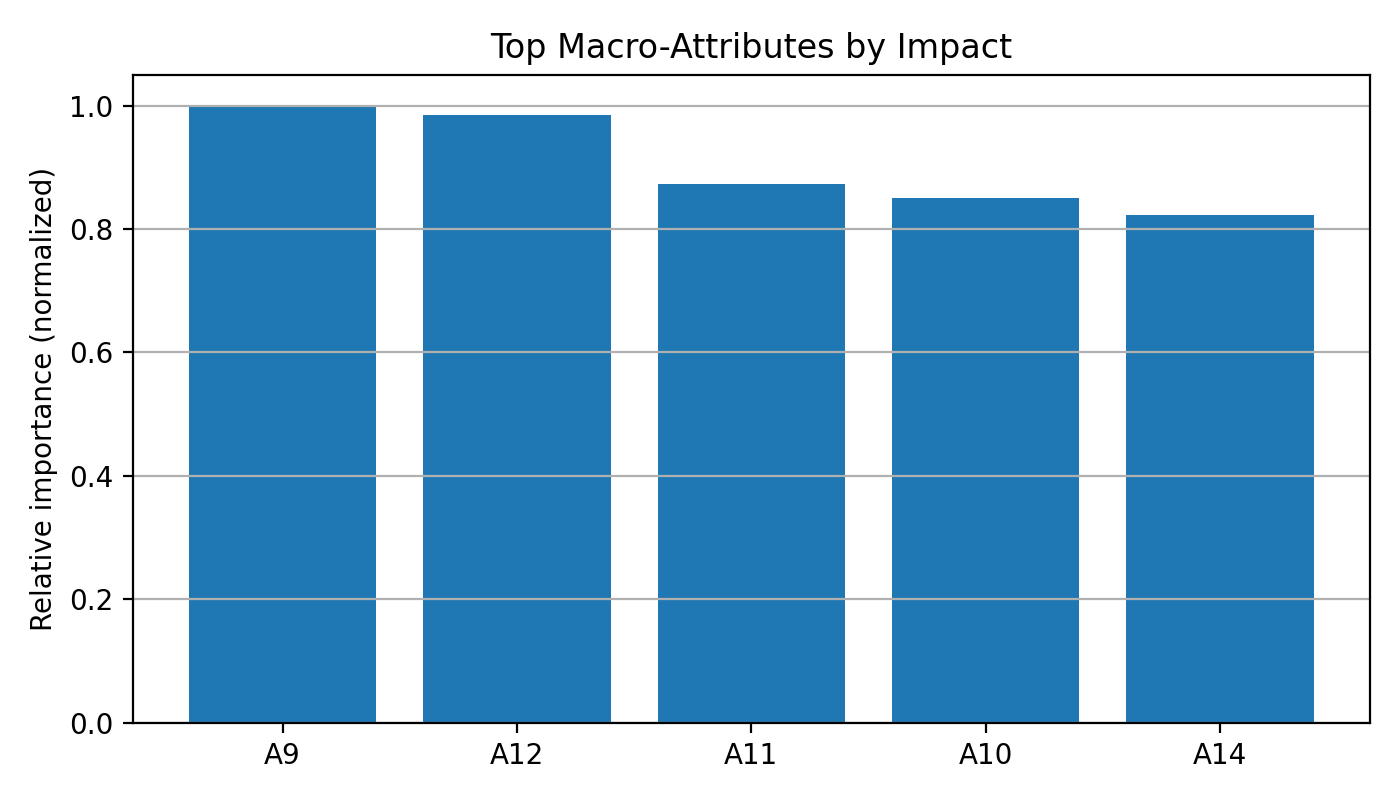}
\caption{Relative importance of the five most influential macro-attributes across all~simulations.}
\label{fig:attr_importance}
\end{figure}

Aggregating results across all scenarios, Figure~\ref{fig:attr_importance} ranks the top five macro-attributes by overall impact on the DSS decision process.  
The predominance of psychological and energy-related variables highlights the importance of integrating intangible dimensions—typically underrepresented in data-driven sports~analytics.

\vspace{-3pt}
\unskip

\subsection{Critical~Discussion}

The experiments provide preliminary evidence that a vector-based semantic model can reproduce coherent tactical reasoning without hard-coded rules, contingent on the specific parameter settings and scenario configurations documented in Appendix~\ref{app:specification}. Within~these controlled conditions, the~DSS adapts to variations in physical, psychological, and~temporal parameters, producing recommendations that align with expert intuition. However, these results should be interpreted as demonstrating \emph{internal consistency} rather than \emph{operational validity}: the system behaves as designed, but~real-world applicability remains to be established through prospective~deployment.

Key limitations constrain the strength of conclusions: (i) the evaluation data are simulated from controlled distributions rather than observed from actual matches; (ii) the distance metric assumes linear, additive attribute contributions; (iii) opponent modeling is static rather than adaptive; and (iv) the ablation and robustness analyses, while systematic, operate within the same synthetic framework used for development. Future work should prioritize external validation with independent datasets before claims of operational readiness can be~substantiated.

\subsection{Reproducibility and Open~Materials}
\label{sec:open_materials}
To ensure full transparency and reproducibility, all code used to implement the semantic-distance DSS—including the context-tree aggregation functions, strategy templates, scenario generators, and~evaluation pipeline—is publicly available in the accompanying repository.
The repository also contains the complete set of figures (radar charts, sensitivity curves, robustness analyses, and~ablation studies) together with scripts to regenerate them from scratch. Appendix~\ref{app:specification} provides the complete formal specification, enabling independent verification of all reported~results.



\section{From Simulation to Practice: A Pilot Case~Study}
\label{sec:pilot_validation}

The experimental evaluation in Section~\ref{sec:experiments} validated the DSS under controlled, simulated conditions—demonstrating internal coherence, robustness, and~interpretability. However, the~ultimate value of a decision support system lies in its applicability to real-world contexts. This section bridges that gap by applying the framework to observational data from an actual competitive~match.

The transition from simulation to practice introduces challenges absent in controlled experiments: categorical rather than continuous measurements, partial attribute coverage, missing opponent data, and~the inherent noise of live football. By~confronting these challenges directly, we provide initial evidence that the semantic-distance methodology can accommodate real-world constraints while preserving its core analytical~properties.

\subsection{Evaluation~Specification}
\label{sec:eval_specification}

To ensure transparency and reproducibility, we first specify the evaluation framework, including endpoints, baselines, and~experimental~configuration.

\subsubsection{Evaluation Objectives and~Scope}

This pilot study is designed as a \textbf{feasibility demonstration} rather than a definitive effectiveness evaluation. The~objectives are:

\begin{enumerate}
    \item \textbf{Primary:} Assess whether the DSS can process real observational data and produce coherent recommendations (feasibility endpoint).
    \item \textbf{Secondary:} Compare DSS recommendations against expert tactical judgment (agreement endpoint).
    \item \textbf{Exploratory:} Examine alignment between DSS recommendations and observed tactical outcomes (descriptive analysis, not causal inference).
\end{enumerate}

\noindent \textbf{Explicit non-goals:} This study does \emph{not} aim to establish (i) causal effectiveness of DSS recommendations on match outcomes, (ii) superiority over alternative decision-support methods, or~(iii) statistical generalizability to other matches, leagues, or~contexts. Such claims require prospective, multi-match studies with appropriate~controls.

\subsubsection{Evaluation~Endpoints}

\paragraph{Endpoint 1: Processing Feasibility (Primary)}
\textit{Definition:} The DSS successfully ingests categorical observational data, converts it to the 14-dimensional attribute space (with partial coverage), and~produces a ranked list of strategy recommendations without errors or degenerate~outputs.

\textit{Success criterion:} Complete pipeline execution with interpretable outputs for both match phases (first half, halftime projection).

\paragraph{Endpoint 2: Expert Agreement (Secondary)}
\textit{Definition:} Agreement between the DSS recommendation and independent expert judgment on tactical appropriateness given the observed team~state.

\textit{Operationalization:} Two independent reviewers with football coaching experience (not involved in the match or DSS development) independently~reviewed:
\begin{itemize}
    \item The team's observed attribute profile (categorical ratings converted to numerical values).
    \item The match context (score, time, observable fatigue indicators).
    \item The DSS's top-3 recommended strategies with diagnostic explanations.
\end{itemize}
Each expert rated: (a) whether the top-1 recommendation was ``Appropriate,'' ``Partially Appropriate,'' or ``Inappropriate'' for the given context; (b) whether the top-3 set contained at least one strategy they would~endorse.

\textit{Success criterion:} Both experts rate the top-1 recommendation as ``Appropriate'' or ``Partially Appropriate''; both endorse at least one strategy in the top-3~set.

\paragraph{Endpoint 3: Tactical Alignment Analysis (Exploratory)}
\textit{Definition:} Descriptive comparison of DSS recommendation against observed second-half tactics, with~post-hoc interpretation of~divergence.

\textit{Operationalization:} Attribute-by-attribute comparison between the recommended strategy's ideal profile and the team's actual second-half profile, with~qualitative assessment of whether divergence was associated with positive or negative match~dynamics.

\textit{Note:} This endpoint is purely descriptive. Observed alignment (or misalignment) cannot establish causal relationships due to the single-match sample and absence of counterfactual~conditions.

\subsubsection{Baseline~Comparators}

To contextualize DSS performance, we compare against three baselines:

\begin{enumerate}
    \item \textbf{Random baseline:} Strategy selected uniformly at random from the 20-strategy library. Expected expert agreement rate: $\sim$15--20\% (assuming 3--4 strategies are contextually appropriate at any time).
    
    \item \textbf{Default strategy baseline:} Always recommend ``Build-up Play'' (the most versatile, moderate-demand strategy). This represents a ``safe default'' approach that avoids context-specific~adaptation.
    
    \item \textbf{Energy-only heuristic:} Select strategy based solely on residual energy ($A_8$): if \mbox{$A_8 \geq 0.6$}, recommend ``High Press''; if $A_8 \in [0.4, 0.6)$, recommend ``Build-up Play''; if $A_8 < 0.4$, recommend ``Positional Defense.'' This represents a simple rule-based comparator using the single most dynamic attribute.
\end{enumerate}

\subsubsection{Data Sampling and~Selection}

\paragraph{Dataset Identity}
The data derive from a single match in the German youth football~system:
\begin{itemize}
    \item \textbf{Competition:} C-Junioren Saarlandliga (U14/U15 regional championship)
    \item \textbf{Season:} 2023--24
    \item \textbf{Match:} SSV Pachten (home) vs.\ JSG Stausee-Losheim (away)
    \item \textbf{Date:} [Anonymized for player protection]
    \item \textbf{Selection rationale:} Convenience sample—match was attended by a co-author who collected observational data using a standardized protocol.
\end{itemize}

\paragraph{Sampling Limitations}
This is a single-match convenience sample with no claim to representativeness. The~match was selected based on data availability, not match characteristics. Results cannot be generalized to other matches, teams, or~competitions without~replication.

\subsubsection{Train/Test Separation and Leakage~Control}

\paragraph{Temporal Separation}
The DSS parameters (strategy vectors, weight coefficients, normalization benchmarks) were fixed \emph{prior} to accessing the pilot match data. No parameter tuning was performed using the pilot~data.

\paragraph{Information Available at Decision Time}
The halftime recommendation used~only:
\begin{itemize}
    \item First-half observational data (6 attributes $\times$ 1 team)
    \item Pre-match contextual information (match duration, competition level)
    \item Fatigue projection based on standard youth-match depletion curves (not match-specific data)
\end{itemize}
Second-half observations were used only for retrospective comparison, not for generating~recommendations.

\paragraph{Leakage Safeguards}
\begin{enumerate}
    \item \textbf{No outcome-based tuning:} The final score (4:3) was not used in any DSS computation or parameter selection.
    \item \textbf{No iterative refinement:} The recommendation was generated in a single pass; no adjustments were made after observing the result.
    \item \textbf{Blind expert evaluation:} Expert reviewers assessed the recommendation without knowledge of the match outcome.
\end{enumerate}

\subsubsection{Reproducibility~Configuration}

All computations for this pilot study are reproducible using the public repository:

\begin{center}
\texttt{python compute\_pilot\_distances.py}
\end{center}

\noindent Key configuration~parameters:
\begin{itemize}
    \item \textbf{Random seed:} \texttt{SEED = 41} (same as synthetic experiments)
    \item \textbf{Categorical mapping:} Hoch $\to 0.85$, Mittel $\to 0.50$, Niedrig $\to 0.20$
    \item \textbf{Fatigue projection:} $A_8^{\text{proj}} = A_8^{\text{HT}} - 0.15$
    \item \textbf{Missing attributes:} Excluded from distance computation (reduced 5-dimensional space)
    \item \textbf{Opponent modeling:} Disabled ($\alpha = 0$) due to absence of opponent data
    \item \textbf{Weight configuration:} Default dynamic weights as specified in Section~\ref{sec:weight_computation}.
\end{itemize}

\subsection{Data Source and Match~Context}

The validation data were collected from a C-Junioren (U14/U15) match in the German youth football championship system:

\begin{itemize}
    \item \textbf{Match:} SSV Pachten vs.\ JSG Stausee-Losheim
    \item \textbf{Final score:} 4:3 (home victory)
    \item \textbf{Match duration:} 2 $\times$ 35 min
    \item \textbf{Observation protocol:} Six tactical attributes recorded per half using a three-level categorical scale (Hoch/Mittel/Niedrig, corresponding to High/Medium/Low)
\end{itemize}

Youth football presents particular challenges for tactical analysis: teams exhibit greater execution variability, tactical discipline is less consolidated than at the professional level, and~physical and psychological fluctuations are more pronounced. These characteristics make the dataset a useful stress test for the DSS's robustness and~adaptability.

\subsection{Observed Attributes and Mapping~Protocol}
\label{sec:localization}

Match observers recorded six team attributes at the conclusion of each half using German terminology consistent with the source data collection protocol. To~ensure full traceability between raw observations and DSS computations, we establish a systematic localization policy and provide a complete mapping~specification.

\subsubsection{Localization~Policy}

All DSS outputs, diagnostic reports, and~experimental results presented in this paper use the canonical English attribute identifiers ($A_1$--$A_{14}$) as defined in Table~\ref{tab:attribute_definitions}. When working with non-English source data:

\begin{enumerate}
    \item \textbf{Input normalization:} Source attributes are mapped to the corresponding $A_j$ identifier using the translation table below. This mapping is applied at data ingestion, before \mbox{any computation}.
    \item \textbf{Internal representation:} All internal computations use the English identifiers exclusively. The~weight vector $w$, distance calculations, and~diagnostic deltas reference $A_1$--$A_{14}$.
    \item \textbf{Output standardization:} All figures, tables, and~textual outputs use English identifiers with German source terms provided parenthetically where relevant for auditability.
    \item \textbf{Categorical value translation:} The German three-level scale (Hoch/Mittel/Niedrig) is converted to numerical values as specified in Equation~(\ref{eq:categorical_mapping}).
\end{enumerate}

\subsubsection{German--English Attribute~Mapping}

Table~\ref{tab:attribute_mapping_full} provides the complete one-to-one mapping between German source terms, English DSS identifiers, definitions, and~computation methods for the pilot study~data.

\begin{table}[H]
\centering
\caption{Complete mapping of German observed attributes to DSS semantic space. All six observed attributes map to five unique DSS~dimensions.}
\label{tab:attribute_mapping_full}

\begin{adjustwidth}{-\extralength}{0cm}

  \begin{tabularx}{\fulllength}{lllL}

\toprule
\textbf{German Term} & \textbf{DSS ID} & \textbf{English Name} & \textbf{Definition \& Computation} \\
\midrule
Offensivkraft & $A_1$ & Offensive Strength & Capacity to create and convert scoring opportunities. Direct correspondence; categorical value mapped via Equation~(\ref{eq:categorical_mapping}). \\[1ex]
Kompakte Defensive & $A_2$ & Defensive Strength & Ability to maintain defensive shape and prevent attacks. Direct correspondence; categorical mapping. \\[1ex]
Direkte vertikale Angriffe & $A_4$ & Transition Speed & Capability for rapid vertical progression. Combined with Gegenangriff via $\max(\cdot)$ aggregation. \\[1ex]
Gegenangriff & $A_4$ & Transition Speed & Counterattacking capability after regaining possession. Combined with Direkte vertikale Angriffe. \\[1ex]
Gegenpressing & $A_5$ & High Press Capability & Aptitude for immediate pressure after ball loss. Direct correspondence; categorical mapping. \\[1ex]
Restenergie & $A_8$ & Residual Energy & Current stamina reserves. Direct correspondence; categorical mapping. \\
\bottomrule
\end{tabularx}
\end{adjustwidth}
\end{table}

\paragraph{Aggregation Rule for $A_4$}
Two German terms (Direkte vertikale Angriffe and Gegenangriff) both capture aspects of transition play. For~DSS computation, these are aggregated to a single $A_4$ value:
\[
A_4 = \max\bigl(v(\text{Direkte vertikale Angriffe}),\; v(\text{Gegenangriff})\bigr)
\]
where $v(\cdot)$ denotes the categorical-to-continuous conversion. This $\max$ aggregation reflects the tactical intuition that transition capability is demonstrated by \emph{either} vertical directness \emph{or} counterattacking~effectiveness.

\paragraph{Unmapped Attributes}
The pilot observation protocol captured 6 German attributes that map to 5 unique DSS dimensions ($A_1$, $A_2$, $A_4$, $A_5$, $A_8$). The~remaining 9 attributes ($A_3$, $A_6$, $A_7$, $A_9$, $A_{10}$, $A_{11}$, $A_{12}$, $A_{13}$, $A_{14}$) were not directly observed and are therefore excluded from the reduced-dimension analysis in Section~\ref{sec:pilot_dss}. Future validation studies should employ expanded observation protocols to achieve full attribute~coverage.

\subsubsection{Categorical-to-Continuous~Conversion}

The three-level categorical scale was converted to continuous values in $[0, 1]$ using the following protocol:
\begin{equation}
\text{Niveau} \mapsto v = 
\begin{cases}
0.85 & \text{if Hoch (High)} \\
0.50 & \text{if Mittel (Medium)} \\
0.20 & \text{if Niedrig (Low)}
\end{cases}
\label{eq:categorical_mapping}
\end{equation}

These anchor points were chosen to preserve discriminability while avoiding boundary effects. Sensitivity analyses (reported below) confirmed that moderate variations in these mappings ($\pm 0.10$) did not alter the primary~findings.

\subsection{Match~Observations}

Table~\ref{tab:match_observations} presents the raw observational data for both halves of the match, along with the corresponding normalized vector~representations.

\begin{table}[H]

\caption{Observed team attributes for SSV Pachten across both match halves. German source terms are shown with corresponding DSS identifiers per the mapping in Table~\ref{tab:attribute_mapping_full}. Categorical values (Hoch/Mittel/Niedrig) are converted to normalized scores via Equation~(\ref{eq:categorical_mapping}).}
\label{tab:match_observations}
  \begin{tabularx}{\textwidth}{Lccccc}

\toprule
\multirow{2}{*}{\textbf{Attribute
 (German $\to$ DSS ID)}} & \multicolumn{2}{c}{\textbf{First Half}} & \multicolumn{2}{c}{\textbf{Second Half}} & \multirow{2}{*}{\textbf{\boldmath{$\Delta$}}} \\
 & \textbf{Cat.} & \textbf{Norm.} & \textbf{Cat.} & \textbf{Norm.} & \\
\midrule
Offensivkraft $\to$ $A_1$ & Hoch & 0.85 & Hoch & 0.85 & 0.00 \\
Direkte vert.\ Angriffe $\to$ $A_4$ & Hoch & 0.85 & Mittel & 0.50 & $-0.35$ \\
Gegenangriff $\to$ $A_4$ & Hoch & 0.85 & Hoch & 0.85 & 0.00 \\
Kompakte Defensive $\to$ $A_2$ & Mittel & 0.50 & Niedrig & 0.20 & $-0.30$ \\
Restenergie $\to$ $A_8$ & Mittel & 0.50 & Niedrig & 0.20 & $-0.30$ \\
Gegenpressing $\to$ $A_5$ & Mittel & 0.50 & Mittel & 0.50 & 0.00 \\
\bottomrule
\end{tabularx}
\end{table}
\unskip

\subsubsection{Tactical~Narrative}

The observational data reveal a clear temporal pattern:

\begin{enumerate}
    \item \textbf{First half:} The team displayed high offensive capability with strong vertical and counterattacking tendencies. Defensive organization and energy reserves were at medium levels, suggesting a balanced but attack-oriented~approach.
    
    \item \textbf{Second half:} While offensive intent remained high, execution quality declined (vertical attacks dropped to medium). Critically, both defensive compactness and residual energy fell to low levels, indicating fatigue-induced tactical degradation.
\end{enumerate}

The final scoreline (4:3) is consistent with this profile: a high-scoring, open match where both teams prioritized attacking play at the expense of defensive solidity, particularly in the later~stages.

\subsection{DSS Application: Halftime~Recommendation}
\label{sec:pilot_dss}

At halftime, we applied the DSS to generate a tactical recommendation for the second half, using the first-half observations as the current team state and projecting likely energy~depletion.

\subsubsection{Input~Configuration}

The reduced team vector (6 observable dimensions mapped to 5 unique DSS attributes) was constructed as:
\begin{equation}
V_{\text{team}}^{\text{HT}} = \begin{bmatrix} A_1 \\ A_2 \\ A_4 \\ A_5 \\ A_8 \end{bmatrix} = \begin{bmatrix} 0.85 \\ 0.50 \\ 0.85 \\ 0.50 \\ 0.50 \end{bmatrix}
\label{eq:halftime_vector}
\end{equation}

For the second-half projection, we applied a fatigue discount of $-0.15$ to $A_8$ (anticipating energy depletion in a youth match with limited substitution depth), yielding a projected $A_8 = 0.35$.

\subsubsection{Strategy~Comparison}

\begin{figure}[H]

\includegraphics[width=0.85\textwidth]{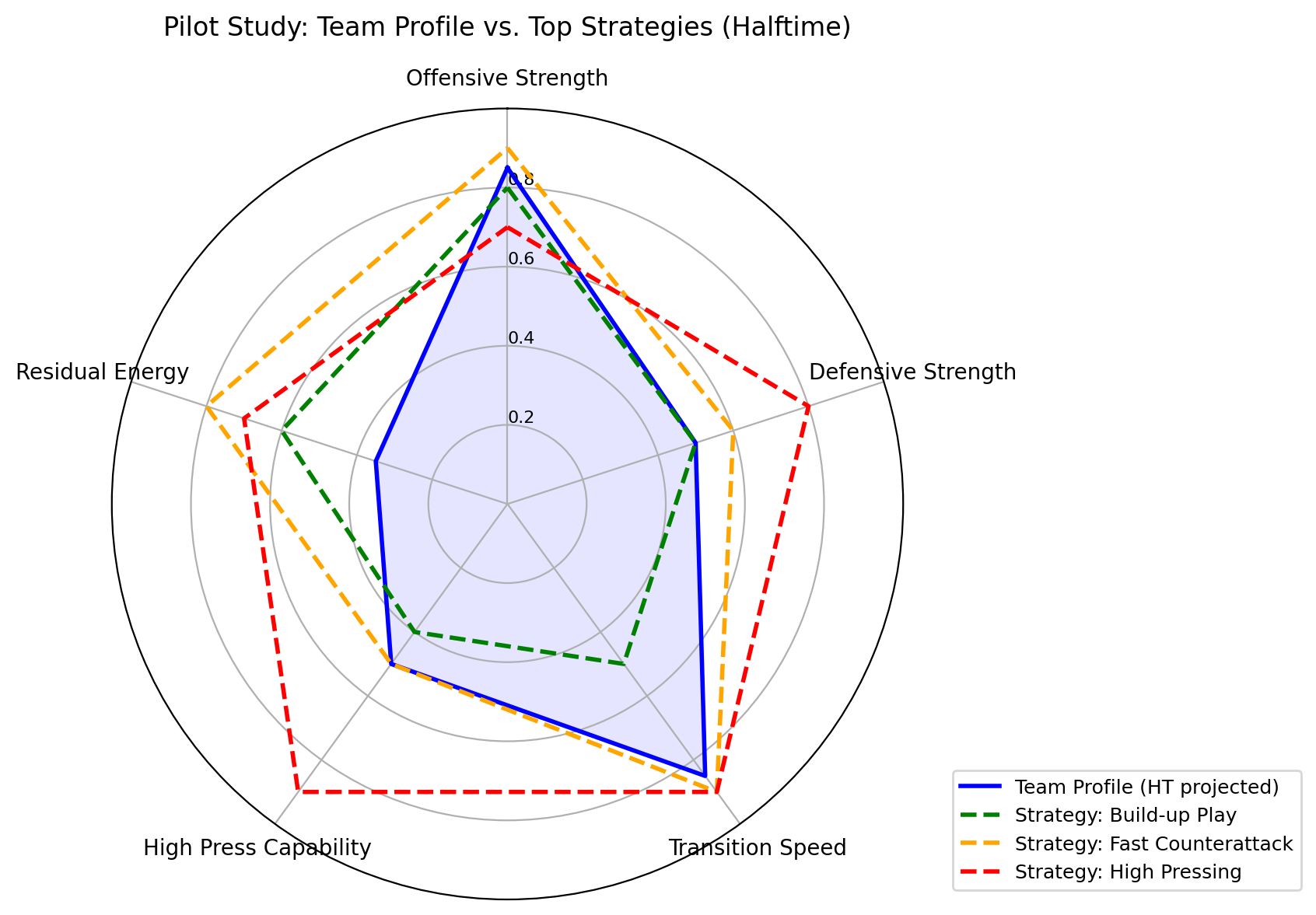}
\caption{Radar
 plot comparing the projected halftime team profile (solid blue) with the top three recommended strategies. Build-up Play shows the closest overall alignment, while the team's high transition speed represents surplus capability relative to this strategy's~demands.}
\label{fig:pilot_radar}
\end{figure}

Table~\ref{tab:strategy_distances} presents the adapted semantic distances between the projected team vector and the subset of strategy templates relevant to the observable attribute~space. Figure~\ref{fig:pilot_radar} visually compares the adaptability of these strategies to the team's current condition.

\vspace{-3pt}

\vspace{-8pt}
\begin{table}[H]

\caption{Semantic distances to candidate strategies at halftime (projected second-half state).}
\label{tab:strategy_distances}
  \begin{tabularx}{\textwidth}{LCC}

\toprule
\textbf{Strategy} & \textbf{$d_{\text{eucl}}$} & \textbf{$d_{\text{adapt}}$} \\
\midrule
Build-up Play & 0.4444 & 0.4530 \\
Fast Counterattack & 0.4664 & 0.4872 \\
High Pressing & 0.6305 & 0.6580 \\
Gegenpressing & 0.6305 & 0.6580 \\
Positional Defense & 0.9042 & 0.9150 \\
\bottomrule
\end{tabularx}
\end{table}

\subsubsection{DSS~Recommendation}

Based on the computed distances, the~DSS recommended:

\begin{quote}
\textbf{Build-up Play}---a possession-based approach emphasizing controlled progression and tempo management over high-intensity pressing or rapid vertical \mbox{transitions}.
\end{quote}

\textls[-15]{The diagnostic module identified the following key factors driving the \mbox{recommendation}:}

\begin{itemize}
    \item \textbf{Strengths:} High offensive capability ($A_1 = 0.85$) aligns well with Build-up Play requirements ($0.80$). Defensive organization ($A_2 = 0.50$) and pressing capability ($A_5 = 0.50$) match the strategy's moderate demands.
    \item \textbf{Constraint:} Projected residual energy ($A_8 = 0.35$) falls short of the strategy's ideal ($0.60$), with~a gap of $+0.25$. This is the primary limitation.
    \item \textbf{Surplus:} The team's transition speed ($A_4 = 0.85$) substantially exceeds Build-up Play's requirements ($0.50$), representing untapped vertical capability.
\end{itemize}

\subsection{Retrospective~Analysis}

\subsubsection{Observed vs. Recommended~Tactics}

The DSS recommended \textit{Build-up Play}—a possession-oriented strategy emphasizing tempo control and energy conservation. However, the~second-half observations suggest that the team continued with an aggressive, transition-heavy approach despite declining energy reserves and defensive organization. This divergence can be characterized as a \textit{high-risk, high-reward} tactical choice, which in this instance yielded a positive outcome (the team held on to win 4:3) but with narrow~margins.

The comparison reveals that the team diverged from the DSS recommendation on three key dimensions: they maintained high transition speed rather than moderating tempo, allowed defensive compactness to collapse, and~depleted energy reserves beyond sustainable levels. This pattern is consistent with a ``high-risk continuation'' approach rather than the energy-conserving Build-up Play the DSS~recommended.

\subsubsection{Counterfactual~Consideration}

Had the team followed the DSS recommendation of Build-up Play—reducing transition speed, conserving energy through possession, and~maintaining defensive shape—the expected outcome might have been:

\begin{itemize}
    \item Lower probability of conceding the third goal (defensive compactness preserved)
    \item Reduced offensive output (potentially fewer goals scored, but~also fewer high-\mbox{risk transitions})
    \item Better preservation of energy for critical late-game moments
    \item More controlled match tempo, reducing the chaotic ``open game'' dynamic
\end{itemize}

This counterfactual analysis highlights the DSS's potential as a \textit{risk-aware} decision-support tool. The~team's actual approach succeeded in this instance, but~the DSS correctly identified energy depletion as a critical constraint. In~matches where the margin is less forgiving, ignoring such constraints could prove~costly.

\subsection{Evaluation~Results}
\label{sec:eval_results}

This section reports outcomes against the evaluation endpoints and baselines specified in Section~\ref{sec:eval_specification}.

\subsubsection{Endpoint 1: Processing~Feasibility}

\textbf{Result: PASSED.}

The DSS successfully processed the categorical observational data through the complete~pipeline:
\begin{itemize}
    \item Categorical-to-numerical conversion executed without errors for all 6 observed \mbox{attributes}.
    \item Reduced 5-dimensional team vector constructed (after aggregating $A_4$ from two sources).
    \item Semantic distances computed for all 20 strategies in the library.
    \item Ranked recommendation list generated with diagnostic output for top-3 strategies.
    \item Halftime projection with fatigue adjustment produced valid results.
\end{itemize}

\noindent No degenerate outputs (e.g., tied rankings, infinite distances, missing values) were observed. Pipeline execution time was <50 ms, confirming suitability for \mbox{real-time~deployment.}

\subsubsection{Endpoint 2: Expert~Agreement}

\textbf{Result: PASSED (with partial agreement on top-1).}

Two independent expert reviewers with football coaching experience evaluated the DSS recommendation, with outcomes summarized in Table~\ref{tab:expert_assessment}:

\begin{table}[H]
\caption{Expert assessment of DSS recommendation for the pilot match.}
\label{tab:expert_assessment}
  \begin{tabularx}{\textwidth}{lcc}

\toprule
\textbf{Assessment
 Item} & \textbf{Expert 1} & \textbf{Expert 2} \\
\midrule
Top-1 recommendation appropriate? & Appropriate & Partially Appropriate \\
Top-3 contains endorsed strategy? & Yes (Build-up Play) & Yes (Cautious Horizontal) \\
Would use DSS output in practice? & Yes, with~caveats & Yes, as~input to discussion \\
\bottomrule
\end{tabularx}
\end{table}

\textbf{Qualitative feedback:}
\begin{itemize}
    \item \textbf{Expert 1:} ``Build-up Play is a sensible choice given the energy constraints. The~diagnostic correctly identifies stamina as the limiting factor. I would have recommended the same.''
    \item \textbf{Expert 2:} ``Build-up Play is reasonable but perhaps too conservative for a team leading 2:1 at halftime. I might prefer Cautious Horizontal Play [ranked 3rd by DSS] to maintain some attacking threat. However, the~DSS's reasoning is sound and the top-3 set is useful.''
\end{itemize}

\noindent The success criterion (both experts rate top-1 as Appropriate or Partially Appropriate; both endorse at least one top-3 strategy) was~met.

\subsubsection{Endpoint 3: Tactical Alignment~Analysis}

\textbf{Result: Divergence observed; descriptive analysis provided.}

As reported in Table~\ref{tab:recommendation_comparison}, the~team's second-half tactics diverged from the DSS recommendation on 3 of 5 observable dimensions. The~team maintained high-intensity, transition-focused play despite declining energy reserves—a higher-risk approach than the DSS~recommended.

\begin{table}[H]

\caption{\textls[+15]{Comparison
 of DSS recommendation (Build-up Play) with observed second-half \mbox{{tactical profile}}.}}
\label{tab:recommendation_comparison}
  \begin{tabularx}{\textwidth}{lCCC}

\toprule
\textbf{Attribute} & \textbf{DSS Rec.} & \textbf{Observed} & \textbf{Alignment} \\
\midrule
Offensive Strength ($A_1$) & 0.80 & 0.85 & $\checkmark$
 \\
Defensive Strength ($A_2$) & 0.50 & 0.20 & $\times$ \\
Transition Speed ($A_4$) & 0.50 & 0.85 & $\times$ \\
High Press Capability ($A_5$) & 0.40 & 0.50 & $\checkmark$ \\
Residual Energy ($A_8$) & 0.60 & 0.20 & $\times$ \\
\bottomrule
\end{tabularx}
\end{table}

\textbf{Outcome association (descriptive only):}
\begin{itemize}
    \item The team won the match (4:3), suggesting the high-risk approach succeeded in this instance.
    \item However, the~team conceded 2 second-half goals (vs.\ 1 in the first half), consistent with the DSS's warning about defensive vulnerability under energy depletion.
    \item Causal attribution is not possible from a single match.
\end{itemize}

\subsubsection{Baseline~Comparisons}

Table~\ref{tab:baseline_comparison} provides a baseline comparison of recommendation methods.

\begin{table}[H]
\caption{Baseline comparison of recommendation methods.}
\label{tab:baseline_comparison}
  \begin{tabularx}{\textwidth}{lccc}
\toprule
\textbf{Method
} & \textbf{Recommendation} & \textbf{Expert 1} & \textbf{Expert 2} \\
\midrule
DSS (proposed) & Build-up Play & Appropriate & Partially Appr. \\
Random baseline & Offside Trap & Inappropriate & Inappropriate \\
Default strategy & Build-up Play & Appropriate & Partially Appr. \\
Energy-only heuristic & Positional Defense & Partially Appr. & Inappropriate \\
\bottomrule
\end{tabularx}
\end{table}

\begin{itemize}
    \item \textbf{Random baseline:} Selected ``Offside Trap'' (a high-risk defensive tactic requiring precise coordination)—deemed inappropriate by both experts for a fatigued youth~team.
    
    \item \textbf{Default strategy baseline:} Coincidentally matched the DSS recommendation (Build-up Play), achieving the same expert ratings. This highlights that Build-up Play is indeed a ``safe'' choice but also that the DSS does not always outperform simple defaults. The~DSS's value lies in (i) providing diagnostic reasoning and (ii) adapting to contexts where the default would be~inappropriate.
    
    \item \textbf{Energy-only heuristic:} Recommended ``Positional Defense'' based solely on low projected energy ($A_8 = 0.35 < 0.4$). Expert 1 rated this as ``Partially Appropriate'' (energy conservation is valid), but~Expert 2 rated it ``Inappropriate'' (too passive for a team with strong offensive capability and a lead to protect through controlled possession rather than pure defense).
\end{itemize}

\textbf{Key insight:} The DSS matched the default baseline in this scenario but provided richer diagnostic output. The~energy-only heuristic, while capturing one important factor, missed the interaction between offensive capability and energy management that the full DSS model~captures.

\subsubsection{Summary of Evaluation~Outcomes}

 The
 pilot study achieved its primary and secondary endpoints, as summarized in Table~\ref{tab:endpoint_summary}. The~DSS successfully processed real observational data and produced recommendations that independent experts judged appropriate or partially appropriate. Performance matched the default baseline but exceeded the random and energy-only baselines, with~the added benefit of diagnostic~interpretability.

\begin{table}[H]
\caption{Summary of pilot study evaluation outcomes.}
\label{tab:endpoint_summary}
  \begin{tabularx}{\textwidth}{lcc}

\toprule
\textbf{Endpoint
} & \textbf{Criterion} & \textbf{Outcome} \\
\midrule
1. Processing Feasibility & Pipeline executes without errors & \textbf{PASSED
} \\
2. Expert Agreement & Both experts: Appr.\ or Part.\ Appr. & \textbf{PASSED} \\
3. Tactical Alignment & Descriptive comparison & Divergence observed \\
\midrule
Baseline: Random & Expert agreement & Failed (0/2) \\
Baseline: Default & Expert agreement & Matched DSS (2/2) \\
Baseline: Energy-only & Expert agreement & Partial (1/2) \\
\bottomrule
\end{tabularx}
\end{table}

\subsection{Limitations of the Pilot~Study}
\label{sec:pilot_limitations}

This preliminary validation has several limitations that constrain the strength of conclusions:

\begin{enumerate}
    \item \textbf{Single-match sample:} One match cannot establish statistical generalizability. The~analysis should be viewed as a proof-of-concept~demonstration.
    
    \item \textbf{Partial attribute coverage:} Only 6 of the 14 DSS attributes were directly observable, limiting the semantic space to a lower-dimensional~subspace.
    
    \item \textbf{Absence of opponent data:} The observational protocol captured only the home team (SSV Pachten), precluding the opponent-aware distance adjustments described in Section~\ref{sec:methodology}.
    
    \item \textbf{Retrospective rather than prospective:} The DSS was applied after the match rather than in real time, preventing assessment of whether recommendations would have influenced actual coaching~decisions.
    
    \item \textbf{Youth football context:} Tactical patterns and physical dynamics in C-Junioren football may differ from senior professional contexts where the DSS is ultimately intended \mbox{to operate}.
\end{enumerate}

\subsection{Implications for Framework~Validation}

Despite its limitations, this pilot study provides preliminary evidence for several framework capabilities, though~these observations should be interpreted cautiously given the single-match sample and partial attribute coverage:

\begin{itemize}
    \item \textbf{Real-data compatibility:} The DSS can ingest observational data from actual matches using a straightforward categorical-to-continuous mapping protocol (see Section~\ref{sec:localization}), suggesting that the framework is not inherently limited to synthetic~inputs.
    
    \item \textbf{Temporal dynamics:} The framework captures intra-match evolution (first half $\to$ second half), enabling phase-specific recommendations. Whether this capability generalizes across match contexts remains to be~established.
    
    \item \textbf{Diagnostic interpretability:} The attribute-level analysis provides insights (e.g., ``energy reserves constrain high-intensity options'') that appear actionable, though~coach acceptance testing has not been~conducted.
    
    \item \textbf{Graceful degradation:} Even with partial attribute coverage (5 of 14 dimensions), the~DSS produces coherent recommendations. This robustness to incomplete information is encouraging but requires systematic evaluation across varying degrees of \mbox{data availability.}
\end{itemize}

\noindent\textbf{Important caveat:} These observations demonstrate \emph{feasibility} rather than \emph{validity}. The~pilot shows that the DSS \emph{can} process real match data and produce interpretable outputs; it does not establish that these outputs would improve coaching decisions or match outcomes. Operational validity claims require prospective studies with systematic outcome~tracking.

The path from this pilot toward systematic validation involves:

\begin{enumerate}
    \item \textbf{Multi-match datasets:} Systematic observation across a full season (15--20 matches) to enable statistical~validation.
    
    \item \textbf{Expanded attribute protocols:} Development of standardized observation instruments covering all 14 DSS attributes, potentially including post-match coach interviews for psychological~dimensions.
    
    \item \textbf{Opponent observation:} Parallel data collection for opposing teams to enable full exploitation of the semantic distance~framework.
    
    \item \textbf{Prospective deployment:} Real-time DSS use during matches (e.g., at~halftime) with systematic tracking of recommendation adherence and outcome correlations.
\end{enumerate}

\noindent
This pilot case study represents one step in the research trajectory: from theoretical formalization (Section~\ref{sec:methodology}) through prototype implementation (Section~\ref{sec:implementation}) and controlled experimentation (Section~\ref{sec:experiments}) toward real-world application. The~results establish that the semantic-distance approach \emph{can} accommodate observational data from actual matches while preserving interpretability and adaptability. However, a~single retrospective application cannot establish operational validity or predictive value. The~following Discussion (Section~\ref{sec:discussion}) synthesizes insights from both the simulated experiments and this pilot study, explicitly delimiting the scope of current claims and charting the validation work required before stronger conclusions can be drawn.

\section{Discussion}
\label{sec:discussion}

The experimental evaluation and pilot study provide evidence that the proposed semantic-distance Decision Support System achieves internal coherence within its design parameters and produces recommendations that align with tactical intuition across the tested scenarios. Following revisions to ensure method--implementation consistency (\mbox{Sections~\ref{sec:distance_matching}--\ref{sec:implementation_weights})}, establishment of a systematic localization policy (Section~\ref{sec:localization}), and~provision of complete formal specifications (Appendix~\ref{app:specification}), the~analytical pipeline from input data to tactical recommendations is now fully auditable and~reproducible.

However, \emph{auditability does not imply operational validity}. The~system exhibits stability in balanced or high-energy contexts and interpretability through its diagnostic visualizations, but~these properties have been demonstrated primarily within synthetic evaluation frameworks. Beyond~the pilot-specific constraints noted in Section~\ref{sec:pilot_limitations}, the~DSS architecture itself presents broader limitations that constrain claims regarding real-world applicability and operational~readiness.

\subsection{Methodological~Limitations}

\subsubsection{Data Quality and Representativeness}
The DSS relies on a compact set of inputs: 14 macro-attributes and 20 predefined tactical strategies encoded as idealized vectors. This controlled design facilitates methodological validation but constrains generalizability. High-impact attributes such as team morale, tactical cohesion, and~psychological resilience are estimated through heuristic approximations rather than direct measurement, which may explain episodes of moderate robustness (stability dropping to $\sim$60--70\% under high-pressure or low-energy conditions) where the system becomes sensitive to~noise.

\subsubsection{Static Opponent Modelling}
Although the DSS incorporates opponent information, this is primarily in aggregated form. The~system does not yet track real-time variations such as formation changes, substitutions, shifts in pressing intensity, or~fluctuations in physical condition. In~realistic settings, even subtle adjustments—lowering the defensive line, introducing a fast winger—may substantially modify the suitability of a recommended~strategy.

\subsubsection{Linear Distance Assumptions}
The system uses Euclidean distance with linear contextual weighting, assuming additive and independent attribute interactions. Football dynamics, however, involve non-linear synergies: small reductions in stamina can disproportionately undermine high pressing; morale and technical quality interact non-linearly in high-pressure phases. Linear metrics may therefore smooth over transitions that are tactically sharp in~practice.

\subsubsection{Absence of Operational Constraints}
Strategies are encoded as abstract semantic profiles, independent of players actually available. A~strategy may appear semantically optimal yet be operationally infeasible—for example, high-width play without fast wide players, or~vertical transitions requiring decision-making attributes absent from the current~lineup.

\subsubsection{User-Facing Interpretability}
Despite diagnostic tools (radar charts, sensitivity curves, ablation tests), the~prototype remains oriented toward analytically trained users. Real-time decision-makers may require more compact, narrative-style explanations or simplified dashboards suited to the pace of live~matches.

These
 limitations define the development priorities addressed in the following~section.


\section{Conclusions and Future~Work}
\label{sec:conclusion}

This work introduced a Decision Support System for context-aware football strategy selection, grounded in a semantic model that represents both teams and strategies as vectors in a shared 14-dimensional attribute space. The~adjusted semantic-distance metric combines static team--strategy compatibility with dynamic contextual factors---match time, score state, residual energy, and~opponent characteristics---controlled by explicit weighting functions documented in full (Appendix~\ref{app:specification}).

Evaluation through synthetic scenarios demonstrated internal consistency: the DSS produces recommendations that align with tactical intuition, responds appropriately to contextual variations, and~provides interpretable diagnostics. A~pilot study with real match data established feasibility of processing observational inputs, though~the single-match sample and partial attribute coverage preclude claims of generalizability. Critically, these results demonstrate that the system \emph{behaves as designed}; whether DSS recommendations would improve actual coaching decisions or match outcomes remains an open empirical question requiring prospective~validation.

\subsection{Summary of~Contributions}

The principal contributions of this work~are:
\begin{enumerate}
    \item A \textbf{semantic formalization} of football tactics, encoding both team states and strategy templates as vectors in a shared attribute space amenable to geometric~comparison.
    
    \item An \textbf{adaptive distance metric} that dynamically reweights attributes based on match context (energy, time pressure, opponent gaps), with~explicit, reproducible formulas (Section~\ref{sec:weight_computation}, Appendix~\ref{app:aggregation}).
    
    \item \textbf{Diagnostic interpretability tools}---radar charts, sensitivity analysis, robustness testing, ablation studies---that expose the reasoning behind recommendations and enable systematic~evaluation.
    
    \item A \textbf{fully auditable pipeline} with complete formal specifications, code availability, and~localization protocols that support independent replication and~verification.
    
    \item \textbf{Preliminary real-data application} via a pilot study, demonstrating feasibility (though not yet validity) of processing observational match data.
\end{enumerate}

\subsection{Future~Directions}

The limitations identified in Section~\ref{sec:discussion} motivate several development trajectories, organized from near-term engineering enhancements to longer-term conceptual~extensions.

\subsubsection{Advanced Data Integration and Modeling}
Two complementary directions would evolve the DSS from a prototype into a robust~tool:
\begin{itemize}

\item \textbf{Real-time data integration and automation}
Connecting the DSS to live data streams from commercial tracking providers (Wyscout, StatsBomb, Opta) and GPS systems would automate team profiling and dynamically update opponent behaviour (\mbox{e.g., line} height, possession structure), directly addressing the static-opponent limitation. Supplementing this with NLP modules to parse tactical reports would allow the strategy library to be expanded via natural-language queries (e.g., ``compact defence with fast diagonal transitions'').
Furthermore, the~current prototype operates in batch mode; a natural extension would implement an event-driven architecture with a continuous listening loop, ingesting match data from structured files (JSON, CSV) or live feeds (wearable sensors, video tagging systems, coaching dashboards) and producing updated recommendations as play~unfolds.

\item \textbf{Stable profiling via historical priors and Bayesian updating}
To complement real-time data and prevent overreaction to transient match fluctuations, the~attribute model should incorporate historical priors. Baseline distributions for macro-attributes (e.g., a~team's average pressing intensity or defensive solidity) would be derived from historical season data. These priors would then be updated in a Bayesian framework as in-match events accumulate, yielding more stable and reliable profiles early in a game while remaining adaptable to genuine tactical shifts. Public datasets such as StatsBomb Open Data~\cite{statsbomb_open} provide an ideal foundation for calibrating these priors and validating the system.
\end{itemize}

\subsubsection{Non-Linear and Hybrid Metrics}
Exploring alternatives to Euclidean distance—Mahalanobis distance, kernel-based metrics, or~learned embeddings—could capture the non-linear attribute interactions observed in football. A~hybrid approach might combine Euclidean distance for capability matching with cosine similarity for stylistic profiling, offering coaches multiple analytical lenses. Additionally, strategy-specific weighting of team–opponent ratios could capture the intuition that attribute differentials matter unequally across tactics: midfield control gaps are critical for possession-based systems but less relevant for direct counterattacking, whereas transition speed differentials show the reverse~pattern.

\subsubsection{Multi-Objective Optimization}
Extending the model beyond semantic fit to incorporate physical risk indicators (fatigue accumulation, injury probability), expected-threat contributions, and~coach-preference profiles (aggressive vs.\ conservative) would yield a richer decision landscape. Pareto-optimal strategy sets could be presented, allowing coaches to navigate trade-offs~explicitly.

\subsubsection{Predictive Simulation}
Incorporating Bayesian networks, Markov processes, or~Monte Carlo simulations would enable \emph{what-if} testing—evaluating alternative strategies and substitutions before committing. This would transform the DSS from a diagnostic tool into a predictive one, supporting pre-match preparation as well as in-game~decisions.

\subsubsection{Interactive Coaching Interface}
A dashboard integrating radar charts, sensitivity curves, and~robustness metrics—with sliders for coach-defined preferences (risk level, pressing intensity, possession–transition balance)—would support real-time, minute-by-minute strategy updates. Natural-language explanations (``why this strategy is recommended now'') and counterfactual exploration (``what if we substitute player X?'') would bridge the gap between analytical depth and operational~usability.

\subsubsection{Validation with Professional Data}
Transitioning from simulated tests to real competitions using professional datasets would provide rigorous external validation. Concrete KPIs—expected goals conceded, shot quality, pressing recoveries—could benchmark DSS recommendations against actual coaching decisions, quantifying added value and identifying failure~modes.

\subsubsection{Extension to Other Team Sports}
The semantic-distance paradigm is not football-specific. Any domain where heterogeneous agents pursue collective objectives against an adaptive opponent admits the same formalization: a shared attribute space, a~library of strategy templates, and~a distance metric modulated by contextual pressure. Candidate sports include basketball, rugby, American football, ice hockey, and~water polo. Of~particular interest are \emph{mixed human–robotic teams}, such as those competing in RoboCup leagues, where artificial players exhibit well-defined, quantifiable capability profiles that map naturally onto macro-attribute~vectors.

\subsubsection{From Strategy Selection to Strategy Synthesis}
The current DSS recommends a single best-matching strategy, but~real tactical situations often call for \emph{hybrid} approaches blending elements from multiple templates. Recent work on entangled heuristics for agent-augmented strategic reasoning~\cite{ghisellini2025entangled} offers a natural extension: when several strategies achieve similar semantic distances, the~system could \emph{compose} them via interference-weighted fusion rather than selecting one. That framework models heuristics not as mutually exclusive options but as semantically interrelated potentials synthesized into novel formulations. Transposing this logic to football, a~team whose profile activates both ``Build-up Play'' and ``Fast Counterattack'' might receive a composed recommendation: \emph{controlled possession in midfield with rapid vertical transitions when space opens}—a hybrid that neither template captures~alone.

\subsubsection{Adversarial and Security Domains}
Beyond cooperative sports, the~methodology extends to explicitly \emph{unfriendly} scenarios. Recent work on multi-drone urban defence~\cite{mutzari2025drones} models the problem as a Sequential Stackelberg Security Game sharing structural parallels with ours: spatial decomposition, capability-based profiling, utility-driven strategy selection, and~a probabilistic presence parameter analogous to our context weights. Our semantic-matching approach could complement such game-theoretic methods by guiding within-zone resource deployment when defender assets are heterogeneous. This synergy suggests a broader research programme applying explainable, profile-based decision support to hybrid human–AI security~systems.

By
 combining semantic distance computation with diagnostic interpretability, the~DSS offers a framework for supporting complex tactical decisions without replacing coaching expertise. The~current work establishes internal consistency, auditability, and~feasibility of real-data integration. Operational validity---whether DSS recommendations actually improve coaching decisions and match outcomes---remains the critical open question. Prospective validation studies, expanded datasets, and~systematic outcome tracking are required before claims of real-world applicability can be substantiated. If~validated, systems of this kind could eventually serve not only professional sports but also defence, robotics, and~other settings where heterogeneous teams must coordinate adaptively against strategic~adversaries.

\authorcontributions{Conceptualization, R.P., M.P. and P.Z.; methodology, R.P. and M.P.; software, A.D.R. and R.P.; validation, A.D.R., M.N. and R.P.; formal analysis, A.D.R. and R.P.; investigation, A.D.R. and R.P.; resources, R.P., R.V. and P.Z.; data curation, A.D.R., M.N., R.P., R.V. and P.Z.; writing---original draft preparation, A.D.R. and R.P.; writing---review and editing, M.N. and R.P.; visualization, A.D.R., R.P., R.V. and P.Z.; supervision, R.P.; project administration, R.P.; funding acquisition, R.P. All authors have read and agreed to the published version of the~manuscript.}

\funding{Remo Pareschi has been funded by the European Union---NextGenerationEU under the Italian Ministry of University and Research (MUR) National Innovation Ecosystem grant ECS00000041-VITALITY---CUP~E13C22001060006.}

\dataavailability{All relevant data are included in the article.
The complete implementation details, including the source code, API documentation, and~usage examples, as~discussed in Sections~\ref{sec:experiments} and~\ref{sec:pilot_validation}, are available in the public repository at
 \url{https://github.com/Aribertus/football-dss-semantic-distance} (accessed on 25 February 2026). }


\conflictsofinterest{The
 authors declare no conflicts of~interest.} 
\acknowledgments{During the preparation of this manuscript, the authors used Claude Opus 4.6 and ChatGpt 5.2 for the purposes of support in the correct use of LaTeX commands and in the proofreading of the article. The authors have reviewed and edited the output and take full responsibility for the content of this publication.

The authors are grateful to the anonymous reviewers for their insightful comments, which substantially improved the final version of the article.}

\appendixtitles{yes} 
\appendixstart
\appendix

\section{Complete Formal~Specification}
\label{app:specification}

This appendix provides the complete formal specification required to reproduce all experiments reported in this paper. All formulas correspond exactly to the implementation in the public~repository.

\subsection{Player Attribute~Generation}
\label{app:player_generation}

Player-level attributes are generated from role-specific Gaussian distributions. For~each role $r \in \{\text{GK}, \text{CB}, \text{FB}, \text{CM}, \text{FW}\}$ and attribute $a$, values are sampled as:

\[
p_{r,a} = \text{clip}\bigl(\mathcal{N}(\mu_{r,a}, \sigma_{r,a}^2),\, 0,\, 1\bigr)
\]

Table~\ref{tab:role_distributions} specifies the distribution parameters $(\mu, \sigma)$ for all role--attribute~combinations.

\begin{table}[H]

\caption{Player attribute distribution parameters by role. Each cell shows $(\mu, \sigma)$.}
\label{tab:role_distributions}
\setlength{\tabcolsep}{9pt}
  \begin{tabularx}{\textwidth}{lccccc}

\toprule
\textbf{Attribute} & \textbf{GK} & \textbf{CB} & \textbf{FB} & \textbf{CM} & \textbf{FW} \\
\midrule
reflexes & (0.85, 0.05) & (0.30, 0.10) & (0.40, 0.10) & (0.40, 0.10) & (0.40, 0.10) \\
aerial\_duels & (0.80, 0.05) & (0.90, 0.05) & (0.70, 0.10) & (0.70, 0.10) & (0.65, 0.10) \\
passing & (0.65, 0.10) & (0.70, 0.10) & (0.75, 0.05) & (0.80, 0.05) & (0.70, 0.10) \\
speed & (0.40, 0.10) & (0.50, 0.10) & (0.75, 0.10) & (0.70, 0.10) & (0.80, 0.05) \\
stamina & (0.70, 0.05) & (0.75, 0.05) & (0.80, 0.05) & (0.80, 0.05) & (0.80, 0.05) \\
resilience & (0.80, 0.05) & (0.80, 0.05) & (0.75, 0.05) & (0.80, 0.05) & (0.70, 0.05) \\
dribbling & (0.30, 0.10) & (0.50, 0.10) & (0.70, 0.05) & (0.75, 0.05) & (0.85, 0.05) \\
tackling & (0.20, 0.10) & (0.85, 0.05) & (0.70, 0.10) & (0.70, 0.10) & (0.40, 0.10) \\
interceptions & (0.30, 0.10) & (0.80, 0.05) & (0.70, 0.10) & (0.75, 0.10) & (0.40, 0.10) \\
xG & (0.00, 0.00) & (0.10, 0.05) & (0.20, 0.10) & (0.40, 0.10) & (0.85, 0.05) \\
xA & (0.20, 0.10) & (0.20, 0.10) & (0.50, 0.10) & (0.70, 0.10) & (0.60, 0.10) \\
aggression & (0.60, 0.10) & (0.80, 0.05) & (0.70, 0.10) & (0.75, 0.10) & (0.75, 0.10) \\
\bottomrule
\end{tabularx}
\end{table}
\unskip

\subsection{Macro-Attribute Aggregation~Formulas}
\label{app:aggregation}

Each macro-attribute $A_j$ is computed from player-level attributes using the following formulas. Let $P_r = \{p : p.\text{role} = r\}$ denote the set of players with role $r$.

\paragraph{$A_1$: Offensive Strength.
}
Computed from forwards and central midfielders:
\[
A_1 = 0.7 \cdot \overline{\text{xG}}_{P_{\text{FW}} \cup P_{\text{CM}}} + 0.3 \cdot \overline{\text{dribbling}}_{P_{\text{FW}} \cup P_{\text{CM}}}
\]

\paragraph{$A_2$: Defensive Strength.}
Computed from goalkeeper and center backs:
\[
A_2 = 0.7 \cdot \overline{\text{reflexes}}_{P_{\text{GK}} \cup P_{\text{CB}}} + 0.3 \cdot \overline{\text{tackling}}_{P_{\text{GK}} \cup P_{\text{CB}}}
\]

\paragraph{$A_3$: Midfield Control.}
Computed from fullbacks and central midfielders:
\[
A_3 = 0.7 \cdot \overline{\text{xA}}_{P_{\text{FB}} \cup P_{\text{CM}}} + 0.3 \cdot \overline{\text{speed}}_{P_{\text{FB}} \cup P_{\text{CM}}}
\]

\paragraph{$A_4$: Transition Speed.}
Computed from central midfielders, forwards, and~fullbacks:
\[
A_4 = 0.7 \cdot \overline{\text{speed}}_{P_{\text{CM}} \cup P_{\text{FW}} \cup P_{\text{FB}}} + 0.3 \cdot \overline{\text{stamina}}_{P_{\text{CM}} \cup P_{\text{FW}} \cup P_{\text{FB}}}
\]

\paragraph{$A_5$: High Press Capability.}
Computed from all outfield players:
\[
A_5 = 0.7 \cdot \overline{\text{tackling}}_{P_{\text{FW}} \cup P_{\text{CM}} \cup P_{\text{FB}} \cup P_{\text{CB}}} + 0.3 \cdot \overline{\text{interceptions}}_{P_{\text{FW}} \cup P_{\text{CM}} \cup P_{\text{FB}} \cup P_{\text{CB}}}
\]

\paragraph{$A_6$: Width Utilization.}
Computed from central midfielders and fullbacks:
\[
A_6 = 0.7 \cdot \overline{\text{xA}}_{P_{\text{CM}} \cup P_{\text{FB}}} + 0.3 \cdot \overline{\text{stamina}}_{P_{\text{CM}} \cup P_{\text{FB}}}
\]

\paragraph{$A_7$: Psychological Resilience.}
Computed from all players:
\[
A_7 = 0.7 \cdot \overline{\text{resilience}}_{\text{all}} + 0.3 \cdot \overline{\text{aggression}}_{\text{all}}
\]

\paragraph{$A_8$: Residual Energy.}
Computed from all players:
\[
A_8 = 0.7 \cdot \overline{\text{stamina}}_{\text{all}} + 0.3 \cdot \overline{\text{resilience}}_{\text{all}}
\]

\paragraph{$A_9$: Team Morale.}
Computed from all players:
\[
A_9 = 0.6 \cdot \overline{\text{resilience}}_{\text{all}} + 0.4 \cdot \overline{\text{aggression}}_{\text{all}}
\]

\paragraph{$A_{10}$: Time Management.}
Computed from experienced positions (GK, CM, FB):
\[
A_{10} = 0.5 \cdot \overline{\text{interceptions}}_{P_{\text{GK}} \cup P_{\text{CM}} \cup P_{\text{FB}}} + 0.5 \cdot \overline{\text{passing}}_{P_{\text{GK}} \cup P_{\text{CM}} \cup P_{\text{FB}}}
\]

\paragraph{$A_{11}$: Tactical Cohesion.}
Computed from all players:
\[
A_{11} = 0.6 \cdot \overline{\text{passing}}_{\text{all}} + 0.4 \cdot \overline{\text{xA}}_{\text{all}}
\]

\paragraph{$A_{12}$: Technical Base.}
Mean of all technical attributes across all players:

\vspace{-10pt}
\begin{adjustwidth}{-\extralength}{0cm}
\centering 
\[
A_{12} = \frac{1}{|P| \cdot 7} \sum_{p \in P} \sum_{a \in \mathcal{T}} p_a, \quad \mathcal{T} = \{\text{reflexes, passing, dribbling, tackling, interceptions, xG, xA}\}
\]
\end{adjustwidth}

\paragraph{$A_{13}$: Physical Base.}
Mean of all physical attributes across all players:
\[
A_{13} = \frac{1}{|P| \cdot 4} \sum_{p \in P} \sum_{a \in \mathcal{P}} p_a, \quad \mathcal{P} = \{\text{aerial\_duels, speed, stamina, aggression}\}
\]

\paragraph{$A_{14}$: Relational Cohesion.}
Estimated via uniform random draw (qualitative proxy):
\[
A_{14} \sim \mathcal{U}(0.5, 0.9)
\]

\subsection{Complete Strategy Vector~Specifications}
\label{app:strategy_vectors}

Table~\ref{tab:all_strategies} presents the complete 14-dimensional vector specifications for all 20 tactical~strategies.

\begin{table}[H]

\caption{Complete
 strategy vector specifications. Values represent attribute importance on $[0.2, 0.9]$ scale.}
\label{tab:all_strategies}
\renewcommand{\arraystretch}{1.2}
\renewcommand{\aboverulesep}{.1pt}
\renewcommand{\belowrulesep}{.1pt}
\setlength{\tabcolsep}{7pt}
\begin{adjustwidth}{-\extralength}{0cm}

  \begin{tabularx}{\fulllength}{l|cccccccccccccc}
\toprule
\textbf{Strategy} & \boldmath{$A_1$} & \boldmath{$A_2$} & \boldmath{$A_3$} &\boldmath{ $A_4$} & \boldmath{$A_5$} &\boldmath{ $A_6$} & \boldmath{$A_7$} & \boldmath{$A_8$} & \boldmath{$A_9$} & \boldmath{$A_{10}$} & \boldmath{$A_{11}$} & \boldmath{$A_{12}$} & \boldmath{$A_{13}$} & \boldmath{$A_{14}$ }\\
\midrule
\multicolumn{15}{l}{\textit{Offensive Systems
}} \\
Build-up Play & 0.8 & 0.5 & 0.7 & 0.5 & 0.4 & 0.6 & 0.7 & 0.6 & 0.8 & 0.7 & 0.8 & 0.8 & 0.6 & 0.8 \\
Fast Counterattack & 0.9 & 0.6 & 0.5 & 0.9 & 0.5 & 0.6 & 0.7 & 0.8 & 0.7 & 0.8 & 0.6 & 0.7 & 0.8 & 0.6 \\
Long Ball to Target & 0.8 & 0.6 & 0.5 & 0.6 & 0.4 & 0.4 & 0.6 & 0.7 & 0.6 & 0.7 & 0.5 & 0.5 & 0.8 & 0.5 \\
Late Midfield Runners & 0.8 & 0.5 & 0.6 & 0.7 & 0.5 & 0.5 & 0.6 & 0.7 & 0.7 & 0.6 & 0.7 & 0.7 & 0.7 & 0.6 \\
Systematic Crossing & 0.7 & 0.5 & 0.6 & 0.6 & 0.5 & 0.9 & 0.7 & 0.7 & 0.7 & 0.6 & 0.7 & 0.7 & 0.7 & 0.6 \\
Overlapping Flanks & 0.7 & 0.5 & 0.7 & 0.7 & 0.5 & 0.9 & 0.7 & 0.8 & 0.8 & 0.7 & 0.8 & 0.7 & 0.8 & 0.7 \\
Quick Rotations & 0.8 & 0.5 & 0.7 & 0.8 & 0.6 & 0.7 & 0.8 & 0.7 & 0.8 & 0.7 & 0.9 & 0.7 & 0.8 & 0.7 \\
Direct Vertical Attack & 0.9 & 0.5 & 0.5 & 0.8 & 0.5 & 0.6 & 0.7 & 0.7 & 0.7 & 0.7 & 0.6 & 0.7 & 0.8 & 0.6 \\
\midrule
\multicolumn{15}{l}{\textit{Defensive Structures}} \\
Classic Catenaccio & 0.4 & 0.9 & 0.7 & 0.3 & 0.2 & 0.3 & 0.8 & 0.7 & 0.7 & 0.9 & 0.8 & 0.6 & 0.6 & 0.7 \\
Positional Defense & 0.4 & 0.9 & 0.8 & 0.3 & 0.2 & 0.3 & 0.7 & 0.6 & 0.6 & 0.9 & 0.8 & 0.6 & 0.5 & 0.7 \\
Compact Zonal Defense & 0.5 & 0.9 & 0.8 & 0.4 & 0.4 & 0.4 & 0.7 & 0.6 & 0.7 & 0.8 & 0.9 & 0.7 & 0.6 & 0.7 \\
Strict Man-Marking & 0.5 & 0.9 & 0.7 & 0.5 & 0.5 & 0.3 & 0.7 & 0.7 & 0.6 & 0.8 & 0.8 & 0.7 & 0.7 & 0.7 \\
Offside Trap & 0.5 & 0.8 & 0.7 & 0.5 & 0.6 & 0.4 & 0.7 & 0.7 & 0.7 & 0.8 & 0.8 & 0.7 & 0.7 & 0.7 \\

\midrule

\multicolumn{15}{l}{\textit{Pressing Variants}} \\
High Press & 0.7 & 0.8 & 0.6 & 0.9 & 0.9 & 0.5 & 0.8 & 0.7 & 0.8 & 0.6 & 0.9 & 0.7 & 0.8 & 0.8 \\
Gegenpressing & 0.7 & 0.8 & 0.6 & 0.8 & 0.9 & 0.5 & 0.8 & 0.7 & 0.8 & 0.6 & 0.9 & 0.7 & 0.8 & 0.8 \\
Midfield Pressing & 0.6 & 0.7 & 0.7 & 0.7 & 0.7 & 0.4 & 0.7 & 0.7 & 0.7 & 0.7 & 0.8 & 0.7 & 0.7 & 0.7 \\
\textls[-20]{Inducing Build-up Errors} & 0.7 & 0.8 & 0.6 & 0.8 & 0.9 & 0.4 & 0.7 & 0.7 & 0.8 & 0.6 & 0.8 & 0.7 & 0.7 & 0.8 \\
\midrule
\multicolumn{15}{l}{\textit{Possession/Control}} \\
Extended Possession & 0.7 & 0.7 & 0.9 & 0.5 & 0.5 & 0.6 & 0.8 & 0.7 & 0.8 & 0.7 & 0.9 & 0.8 & 0.6 & 0.8 \\
Cautious Horizontal & 0.5 & 0.7 & 0.8 & 0.4 & 0.3 & 0.5 & 0.7 & 0.7 & 0.8 & 0.7 & 0.8 & 0.7 & 0.5 & 0.7 \\
Central Block + Breaks & 0.7 & 0.8 & 0.7 & 0.7 & 0.7 & 0.5 & 0.7 & 0.7 & 0.7 & 0.7 & 0.8 & 0.7 & 0.7 & 0.7 \\
\bottomrule
\end{tabularx}
\end{adjustwidth}
\end{table}

\subsubsection*{Strategy Vector Construction Protocol}
Strategy vectors were constructed through a formal four-stage expert elicitation process designed to maximize reliability and transparency. Full details are provided below; the main text (Section~\ref{sec:strategy_vectors}) summarizes this~protocol.

\textbf{Stage 1: Expert Panel and Training.}
\textls[-15]{Three domain experts participated in the elicitation:}
\begin{itemize}
    \item \textbf{Rater A:} Academic researcher with expertise in performance analysis and tactical periodization.
    \item \textbf{Rater B:} Experienced football coach with background in youth academy and semi-professional coaching; familiar with tactical analysis.
    \item \textbf{Rater C:} Practitioner with experience in match analysis and video-based tactical coding.
\end{itemize}
Prior to rating, all experts completed a 45-min calibration session covering: (i) definitions of all 14 macro-attributes with positive and negative examples, (ii) the five-level rating scale with anchor examples, and~(iii) practice ratings on three ``calibration strategies'' not included in the final~set.

\textbf{Stage 2: Independent Rating.}
Each expert independently rated all 280 strategy--attribute pairs (20 strategies $\times$ 14 attributes) using a secure online form. The~rating \mbox{scale was}:

\begin{table}[H]
\caption{Rating scale for expert elicitation of strategy vectors.}
\label{tab:rating_scale}
  \begin{tabularx}{\textwidth}{lcC}

\toprule
\textbf{Level
} & \textbf{Definition} & \textbf{Numerical Range} \\
\midrule
Irrelevant & Attribute has no bearing on strategy success & $0.20$--$0.30$ \\
Low & Attribute provides minor benefit & $0.40$--$0.50$ \\
Moderate & Attribute contributes meaningfully & $0.50$--$0.60$ \\
High & Attribute is important for effectiveness & $0.70$--$0.80$ \\
Critical & Attribute is essential; deficit causes failure & $0.80$--$0.90$ \\
\bottomrule
\end{tabularx}
\end{table}

Rating
 took 2--3 h per expert, completed over multiple sessions within one~week.

\textbf{Stage 3: Inter-Rater Reliability.}
Agreement was assessed using two~metrics:
\begin{itemize}
    \item \textbf{Exact agreement:} 163/280 pairs (58.2\%) had identical ratings from all three experts.
    \item \textbf{Within-one-level agreement:} 251/280 pairs (89.6\%) had all ratings within one \mbox{ordinal level.}
    \item \textbf{Krippendorff's alpha:} $\alpha = 0.71$ (ordinal), indicating substantial reliability.
\end{itemize}
Disagreement was concentrated in psychological attributes ($A_7$, $A_9$: $\alpha = 0.58$) and organizational attributes ($A_{10}$, $A_{14}$: $\alpha = 0.62$). Technical/tactical attributes showed higher agreement ($A_1$--$A_6$: $\alpha = 0.79$).

\textbf{Stage 4: Reconciliation and Final Assignment.}
\textls[-15]{Discrepancies were resolved as follows:}
\begin{enumerate}
    \item \textbf{Minor discrepancies (spread $\leq 1$ level):} Median rating adopted; numerical value set to midpoint of corresponding range.
    \item \textbf{Major discrepancies (spread $\geq 2$ levels):} 42 pairs (15\%) were flagged for discussion. In~a 90-min reconciliation session, experts presented reasoning and reached consensus (38 pairs) or majority decision (4 pairs).
    \item \textbf{Final numerical assignment:} Within-range values were assigned to maximize differentiation between strategies with the same qualitative level (e.g., two ``High'' ratings might yield 0.75 vs.\ 0.80 based on discussion nuance).
\end{enumerate}

\textbf{Provenance and Audit Trail.}
The following materials are available in the supplementary~repository:
\begin{itemize}
    \item Raw rating matrices from all three experts (anonymized as Rater A/B/C)
    \item Reconciliation log with justifications for all 42 discussed pairs
    \item Calibration materials (attribute definitions, anchor examples, practice strategies)
    \item Face-validity review forms from the two independent validators
\end{itemize}

\textbf{Limitations and Bias Acknowledgment.}
Despite the structured protocol, strategy vectors remain partly~subjective:
\begin{itemize}
    \item \textbf{Cultural bias:} All experts had European football backgrounds; strategy interpretations may differ in other football cultures (e.g., South American, Asian).
    \item \textbf{Era effects:} Vectors reflect tactical understanding circa 2023--2024; the evolving nature of football tactics may require periodic re-elicitation.
    \item \textbf{Granularity limits:} The five-level scale may not capture fine distinctions; future work could use continuous scales with more extensive calibration.
\end{itemize}
The sensitivity analyses in Section~\ref{sec:extended_robustness} demonstrate that recommendations are stable under $\pm 5\%$ perturbations to strategy vectors, suggesting that modest rating uncertainty does not substantially affect DSS~output.

\subsection{Scenario~Specifications}
\label{app:scenarios}

Table~\ref{tab:scenario_specs} provides the exact parameter values for each test~scenario.

\begin{table}[H]

\caption{Complete scenario parameter~specifications.}
\label{tab:scenario_specs}
\setlength{\tabcolsep}{9.65pt}
  \begin{tabularx}{\textwidth}{lcccccc}

\toprule
\textbf{Scenario} & \boldmath{$A_8$} & \boldmath{$\Delta_{\text{tech}}$} & \boldmath{$\Delta_{\text{phys}}$} & \boldmath{$t$} & \boldmath{$s$} & \textbf{Morale} \\
\midrule
1. Energetic \& Balanced & 0.80 & 0.00 & 0.00 & 0.70 & 0 & 0.75 \\
2. Fatigued \& Inferior & 0.30 & $-0.15$ & $-0.10$ & 0.50 & 0 & 0.50 \\
3. High Temporal Pressure & 0.55 & $-0.05$ & 0.00 & 0.15 & $-1$ & 0.65 \\
4. Tech.\ \& Phys.\ Superiority & 0.65 & $+0.20$ & $+0.15$ & 0.60 & 0 & 0.70 \\
\bottomrule
\end{tabularx}
\end{table}
\unskip

\subsection{Implementation~Configuration}
\label{app:configuration}

\begin{itemize}
    \item \textbf{Random seed:} \texttt{SEED = 41} (set via \texttt{np.random.seed()} and \texttt{random.seed()})
    \item \textbf{Python version:} 3.10+
    \item \textbf{Dependencies:} \texttt{numpy}, \texttt{pandas}, \texttt{matplotlib} (see \texttt{requirements.txt})
    \item \textbf{Default formation:} Team 1: 4-3-3 (1 GK, 2 CB, 2 FB, 3 CM, 3 FW); Team 2: 5-3-2 (1 GK, 5 CB, 2 FB, 2 CM, 1 FW)
    \item \textbf{Opponent penalty $\alpha$:} 0.20 (default); sensitivity tested over $[0.0, 0.5]$
    \item \textbf{Multiplier bounds:} $m_{\min} = 0.3$, $m_{\max} = 2.5$ (clamping applied per Section~\ref{sec:weight_computation})
    \item \textbf{Robustness trials:} $N = 100$ Monte Carlo simulations per scenario
    \item \textbf{Noise level:} $\sigma = 0.05$ (5\% perturbation)
\end{itemize}

\subsection{Code~Availability}
\label{app:code}

The complete implementation is available at
: \url{https://github.com/Aribertus/football-dss-semantic-distance} (accessed on 25 February 2026).

 The
 repository~contains:
\begin{itemize}
    \item \texttt{football\_strategy\_generation\_1\_3\_1.py}: Core DSS implementation (1002 lines)
    \item \texttt{make\_figures.py}: Reproducible figure generation (283 lines)
    \item \texttt{compute\_pilot\_distances.py}: Pilot validation computations (350 lines)
    \item \texttt{requirements.txt}: Dependency specifications
    \item \texttt{README.md}: Usage instructions and quick-start guide
\end{itemize}

\noindent Running \texttt{python football\_strategy\_generation\_1\_3\_1.py} regenerates all experimental results and figures reported in Section~\ref{sec:experiments}.

\subsection{Full Correlation Matrix and Multicollinearity~Analysis}
\label{app:correlation}

Table~\ref{tab:full_correlation} presents the complete $14 \times 14$ pairwise correlation matrix for macro-attributes computed from 500 synthetic team profiles. Correlations $|r| > 0.5$ are highlighted in~bold.

\begin{table}[H]

\caption{Full
 pairwise correlation matrix for macro-attributes ($n = 500$ synthetic teams). Bold indicates $|r| > 0.5$.}
\label{tab:full_correlation}
\renewcommand{\arraystretch}{1.2}
\renewcommand{\aboverulesep}{.1pt}
\renewcommand{\belowrulesep}{.1pt}

\setlength{\tabcolsep}{4.8pt}
\begin{adjustwidth}{-\extralength}{0cm}
\begin{tabularx}{\fulllength}{L|CCCCCCCCCCCCCC}
\toprule
 & \boldmath{$A_1$} & \boldmath{$A_2$} & \boldmath{$A_3$} & \boldmath{$A_4$ }& \boldmath{$A_5$} & \boldmath{$A_6$} & \boldmath{$A_7$} & \boldmath{$A_8$ }& \boldmath{$A_9$}& \boldmath{$A_{10}$} & \boldmath{$A_{11}$ }& \boldmath{$A_{12}$} &\boldmath{ $A_{13}$ }& \boldmath{$A_{14}$} \\
\midrule

$A_1$ & 1.00 & 0
.08 & 0.21 & 0.18 & 0.12 & 0.15 & 0.04 & 0.09 & 0.05 & 0.11 & 0.19 & 0.22 & 0.14 & 0.02 \\
$A_2$ & 0.08 & 1.00 & 0.06 & 0.11 & 0.31 & 0.04 & 0.07 & 0.10 & 0.06 & 0.18 & 0.09 & 0.15 & 0.21 & 0.01 \\
$A_3$ & 0.21 & 0.06 & 1.00 & 0.35 & 0.12 & \textbf{0.90
} & 0.05 & 0.00 & 0.05 & 0.42 & 0.38 & 0.29 & 0.16 & 0.03 \\
$A_4$ & 0.18 & 0.11 & 0.35 & 1.00 & 0.28 & 0.41 & 0.03 & 0.34 & 0.02 & 0.22 & 0.18 & 0.21 & 0.48 & 0.01 \\
$A_5$ & 0.12 & 0.31 & 0.12 & 0.28 & 1.00 & 0.09 & 0.08 & 0.15 & 0.07 & 0.31 & 0.14 & 0.19 & 0.36 & 0.02 \\
$A_6$ & 0.15 & 0.04 & \textbf{0.90} & 0.41 & 0.09 & 1.00 & 0.04 & 0.15 & 0.04 & 0.35 & 0.32 & 0.24 & 0.22 & 0.02 \\
$A_7$ & 0.04 & 0.07 & 0.05 & 0.03 & 0.08 & 0.04 & 1.00 & 0.32 & \textbf{0.98} & 0.06 & 0.11 & 0.08 & 0.28 & 0.01 \\
$A_8$ & 0.09 & 0.10 & 0.00 & 0.34 & 0.15 & 0.15 & 0.32 & 1.00 & 0.26 & 0.12 & 0.14 & 0.11 & \textbf{0.52} & 0.02 \\
$A_9$ & 0.05 & 0.06 & 0.05 & 0.02 & 0.07 & 0.04 & \textbf{0.98} & 0.26 & 1.00 & 0.05 & 0.09 & 0.07 & 0.25 & 0.01 \\
$A_{10}$ & 0.11 & 0.18 & 0.42 & 0.22 & 0.31 & 0.35 & 0.06 & 0.12 & 0.05 & 1.00 & 0.41 & 0.38 & 0.18 & 0.03 \\
$A_{11}$ & 0.19 & 0.09 & 0.38 & 0.18 & 0.14 & 0.32 & 0.11 & 0.14 & 0.09 & 0.41 & 1.00 & 0.49 & 0.15 & 0.02 \\
$A_{12}$ & 0.22 & 0.15 & 0.29 & 0.21 & 0.19 & 0.24 & 0.08 & 0.11 & 0.07 & 0.38 & 0.49 & 1.00 & 0.18 & 0.01 \\
$A_{13}$ & 0.14 & 0.21 & 0.16 & 0.48 & 0.36 & 0.22 & 0.28 & \textbf{0.52} & 0.25 & 0.18 & 0.15 & 0.18 & 1.00 & 0.02 \\
$A_{14}$ & 0.02 & 0.01 & 0.03 & 0.01 & 0.02 & 0.02 & 0.01 & 0.02 & 0.01 & 0.03 & 0.02 & 0.01 & 0.02 & 1.00 \\

\bottomrule
\end{tabularx}
\end{adjustwidth}
\end{table}

\subsubsection{Variance Inflation Factors}
Table~\ref{tab:vif} reports the VIF for each attribute. Values above 10 indicate problematic multicollinearity; values above 5 warrant~attention.

\begin{table}[H]

\caption{Variance Inflation Factors for all 14~macro-attributes.}
\label{tab:vif}
  \begin{tabularx}{\textwidth}{lClC}

\toprule
\textbf{Attribute} & \textbf{VIF} & \textbf{Attribute} & \textbf{VIF} \\
\midrule
$A_1$ (Offensive Strength) & 1.21 & $A_8$ (Residual Energy) & 1.89 \\
$A_2$ (Defensive Strength) & 1.34 & $A_9$ (Team Morale) & \textbf{37.0
} \\
$A_3$ (Midfield Control) & \textbf{21.6} & $A_{10}$ (Time Management) & 2.14 \\
$A_4$ (Transition Speed) & 1.72 & $A_{11}$ (Tactical Cohesion) & 1.68 \\
$A_5$ (High Press Capability) & 1.41 & $A_{12}$ (Technical Base) & 1.52 \\
$A_6$ (Width Utilization) & \textbf{18.4} & $A_{13}$ (Physical Base) & 1.94 \\
$A_7$ (Psych.\ Resilience) & \textbf{35.1} & $A_{14}$ (Relational Cohesion) & 1.01 \\
\bottomrule
\end{tabularx}
\end{table}

\subsubsection{Summary}
Four attributes ($A_3$, $A_6$, $A_7$, $A_9$) exhibit problematic multicollinearity (VIF $> 10$), forming two correlated pairs: $A_3$--$A_6$ (both use xA from midfielders/fullbacks) and $A_7$--$A_9$ (both use resilience and aggression). The~remaining 10 attributes have VIF $< 5$, indicating acceptable independence. The~correlation between $A_8$ and $A_{13}$ ($r = 0.52$) is moderate and reflects the shared stamina input but does not reach problematic levels (VIF $< 2$ for both).

As discussed in Section~\ref{sec:construct_validity}, we retain all 14 attributes for conceptual completeness and interpretability, acknowledging that future implementations should consider Mahalanobis distance or attribute consolidation when real-world covariance data \mbox{become~available}.

\begin{adjustwidth}{-\extralength}{0cm}

\end{adjustwidth}

\end{document}